\documentclass{article}

\usepackage[nonatbib,final]{neurips_data_2024}

\usepackage[utf8]{inputenc} % allow utf-8 input
\usepackage[T1]{fontenc}    % use 8-bit T1 fonts
\usepackage{hyperref}       % hyperlinks
\usepackage{url}            % simple URL typesetting
\usepackage{booktabs}       % professional-quality tables
\usepackage{amsfonts}       % blackboard math symbols
\usepackage{nicefrac}       % compact symbols for 1/2, etc.
\usepackage{microtype}      % microtypography
\usepackage{xcolor}         % colors
\usepackage{amssymb}
\usepackage{amsmath}
\usepackage{pifont}
\newcommand{\cmark}{\ding{51}}%
\newcommand{\xmark}{\ding{55}}%
\usepackage{graphicx}
\usepackage{makecell}
\usepackage{subcaption}
\usepackage{tabularx}
\usepackage{bbm}
\usepackage{listings}
\usepackage{xcolor}
\usepackage{threeparttable}
\usepackage{wrapfig}
\usepackage{multirow}

\definecolor{keyword}{rgb}{0.26, 0.44, 0.76}
\definecolor{comment}{rgb}{0.5, 0.5, 0.5}
\definecolor{string}{rgb}{0.56, 0.93, 0.56}
\definecolor{backcolour}{rgb}{0.95, 0.95, 0.92}

\lstdefinestyle{pseudocode}{
    language=Python,
    basicstyle=\ttfamily\footnotesize,
    keywordstyle=\color{keyword}\bfseries,
    commentstyle=\color{comment}\itshape,
    stringstyle=\color{string},
    backgroundcolor=\color{backcolour},
    showstringspaces=false,
    breaklines=true,
    frame=single,
    captionpos=b,
    morekeywords={input, output, while, for, if, else, return}  % Add common pseudocode keywords
}
\title{ChaosBench: A Multi-Channel, Physics-Based Benchmark for Subseasonal-to-Seasonal Climate Prediction}

\author{
  {\bf Juan Nathaniel}\textsuperscript{1,*}, {\bf Yongquan Qu}\textsuperscript{1}, {\bf Tung Nguyen}\textsuperscript{2}, {\bf Sungduk Yu}\textsuperscript{3,5}, {\bf Julius Busecke}\textsuperscript{1,4}, \\
  {\bf Aditya Grover}\textsuperscript{2},
  {\bf Pierre Gentine}\textsuperscript{1}
  \\
  \textsuperscript{1}Columbia University, \textsuperscript{2}UCLA, \textsuperscript{3}UCI, \textsuperscript{4}LDEO, \textsuperscript{5} Intel Labs
}

\begin{document}

\maketitle
\renewcommand{\thefootnote}{\fnsymbol{footnote}}
\footnotetext[1]{Corresponding author: jn2808@columbia.edu}

\begin{abstract}
Accurate prediction of climate in the subseasonal-to-seasonal scale is crucial for disaster preparedness and robust decision making amidst climate change. Yet, forecasting beyond the weather timescale is challenging because it deals with problems other than initial condition, including boundary interaction, butterfly effect, and our inherent lack of physical understanding. At present, existing benchmarks tend to have shorter forecasting range of up-to 15 days, do not include a wide range of operational baselines, and lack physics-based constraints for explainability. Thus, we propose ChaosBench, a challenging benchmark to extend the predictability range of data-driven weather emulators to S2S timescale. First, ChaosBench is comprised of variables beyond the typical surface-atmospheric ERA5 to also include ocean, ice, and land reanalysis products that span over 45 years to allow for full Earth system emulation that respects boundary conditions. We also propose physics-based, in addition to deterministic and probabilistic metrics, to ensure a physically-consistent ensemble that accounts for butterfly effect. Furthermore, we evaluate on a diverse set of physics-based forecasts from four national weather agencies as baselines to our data-driven counterpart such as ViT/ClimaX, PanguWeather, GraphCast, and FourCastNetV2. Overall, we find methods originally developed for weather-scale applications fail on S2S task: their performance simply collapse to an unskilled climatology. Nonetheless, we outline and demonstrate several strategies that can extend the predictability range of existing weather emulators, including the use of ensembles, robust control of error propagation, and the use of physics-informed models. Our benchmark, datasets, and instructions are available at \href{https://leap-stc.github.io/ChaosBench}{https://leap-stc.github.io/ChaosBench}.
\end{abstract}

\section{Introduction}
Although critical for economic planning, disaster preparedness, and policy-making, subseasonal-to-seasonal (S2S) prediction is lagging behind the more established field of short/medium-range weather, or long-range climate predictions. For instance, many natural hazards tend to manifest in the S2S scale, including the slow-onset of droughts that lead to wildfire \cite{pendergrass2020flash,buch2023smlfire1}, heavy precipitations that lead to flooding \cite{shamekh2023implicit}, and persistent weather anomalies that lead to extremes \cite{zeppetello2022physics}. So far, current approaches to weather and climate prediction are heavily reliant on physics-based models in the form of Numerical Weather Prediction (NWP). Many NWPs are based on the discretization of governing equations that describe thermodynamics, fluid flows, \emph{etc}. However, these models are expensive to run especially in high-resolution setting. For example, there are massive computational overheads to perform numerical integration at fine spatiotemporal resolutions that are operationally useful \cite{schneider2023harnessing}. Furthermore, their relative inaccessibility to non-experts is a major roadblock to the broader community. As a result, there is a growing interest to apply data-driven models to emulate NWPs, as they tend to have faster inference speed, are less resource-hungry, and more accessible \cite{bi2023accurate,lam2022graphcast,mukkavilli2023ai,pathak2022fourcastnet,qu2023can,qu2024joint,yu2023climsim}. Nevertheless, many data-driven benchmarks have so far been focused on the short (1-5 days), medium (5-15 days), and long (years-decades) forecasting ranges. In this work, we include S2S as a more challenging task that requires different emulation strategies: being in between two extremes, it is doubly sensitive to (1) \emph{initial conditions} (IC) as in the case for short/medium-range weather, and (2) \emph{boundary conditions} (BC) as in the case for long-range climate \cite{prive2013role, wu2005estimating, lorenz1963deterministic,cresswell2024deep}. 

\begin{figure}
    \captionsetup{type=figure}
    \includegraphics[width=\textwidth]{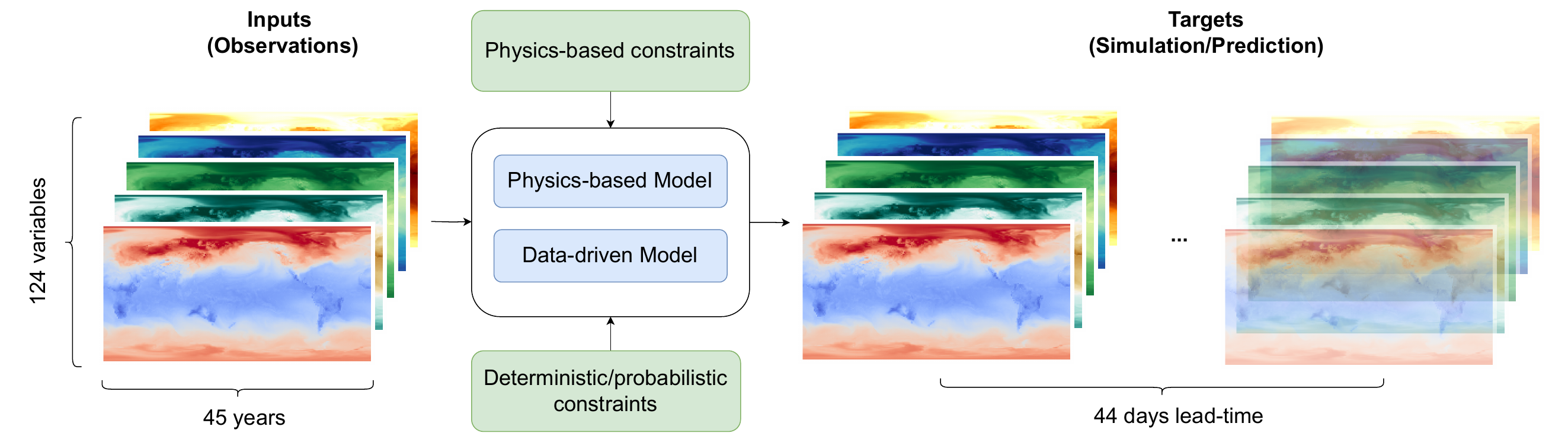}
    \captionof{figure}{We propose ChaosBench, a large-scale, fully-coupled, physics-based benchmark for subseasonal-to-seasonal (S2S) climate prediction. It is framed as a high-dimensional sequential regression task that consists of 45+ years, multi-system observations for validating physics-based and data-driven models, and training the latter. Physics-based forecasts are generated from four national weather agencies with 44-day lead-time and serve as baselines to data-driven forecasts. Our benchmark is one of the first to incorporate physics-based metrics to ensure physically-consistent and explainable models. The blurred image at $\Delta t=44$ represents a challenge of long-term forecasting.}
    \label{fig:chaosbench}
\end{figure}

We propose ChaosBench to bridge these gaps (Figure \ref{fig:chaosbench}). It is comprised of variables beyond the typical surface-atmospheric ERA5 to also include ocean, ice, and land reanalysis products that span over 45 years to allow for full Earth system emulation that respects boundary processes. We also provide 44-day ahead physics-based control (deterministic) and perturbed (ensemble) forecasts from four national weather agencies over the last 8 years as baselines. In addition, we introduce physics-based and incorporate probabilistic, in addition to deterministic metrics, for a more physically-consistent ensemble that accounts for butterfly effect. As far as we know, ChaosBench is one of the first to systematically evaluate several state-of-the-art data-driven models including ViT/ClimaX \cite{nguyen2023climax}, PanguWeather \cite{bi2022pangu}, GraphCast \cite{lam2022graphcast}, and FourCastNetV2 \cite{pathak2022fourcastnet} on S2S predictability.

% \begin{table*}[h]
% \centering
% \begin{tabular}{l|c|c|cccccccccc}
% \toprule
% & & & \multicolumn{10}{c}{\textbf{Vertical Pressure Levels (hPa)}}\\
% \textbf{Variables} & \textbf{Notation} & \textbf{Unit} & 1000 & 925 & 850 & 700 & 500 & 300 & 200 & 100 & 50 & 10 \\ \hline
% Geopotential height & $z$ & $gpm$ & \cmark & \cmark & \cmark & \cmark & \cmark & \cmark & \cmark & \cmark & \cmark & \cmark \\
% Specific humidity &$q$ & $kg\ kg^{-1}$ & \cmark & \cmark & \cmark & \cmark & \cmark & \cmark & \cmark & - & - & - \\
% Temperature & $t$ & $K$ & \cmark & \cmark & \cmark & \cmark & \cmark & \cmark & \cmark & \cmark & \cmark & \cmark \\
% U component of wind & $u$ & $m\ s^{-1}$ & \cmark & \cmark & \cmark & \cmark & \cmark & \cmark & \cmark & \cmark & \cmark & \cmark \\
% V component of wind & $v$ & $m\ s^{-1}$ & \cmark & \cmark & \cmark & \cmark & \cmark & \cmark & \cmark & \cmark & \cmark & \cmark \\
% Vertical velocity & $w$ & $Pa\ s^{-1}$ & - & - & - & - & \cmark & - & - & - & - & - \\
% \bottomrule
% \end{tabular}
% \caption{List of variables and their corresponding pressure levels (hPa) available for physics-based model (48 variable-channels). The input observations consist of all variable/level combinations (60 variable-channels).}
% \label{tab:variable}
% \end{table*}

In this work, we demonstrate that existing physics-based and data-driven models are indistinguishable from unskilled climatology as the forecasting range approaches the S2S timescale. The high spectral divergence observed in many state-of-the-art models suggests the lost of predictive accuracy of multi-scale structures. This leads to significant blurring and a tendency towards smoother predictions. For one, such averaging is of little use when one attempts to identify extreme events requiring high-fidelity forecasts on the S2S scale (e.g., regional droughts, hurricanes, \emph{etc}). Also, performing comparably worse than climatology renders them \emph{operationally unusable}. This highlights the urgent need for a robust and unified data-driven S2S intercomparison project.

\section{Related Work}
\label{sec:related}

In recent years, several benchmarks have been introduced to push the field of data-driven weather and climate prediction \cite{kashinath2021climatenet,racah2017extremeweather,nathaniel2023metaflux,rasp2020weatherbench,hwang2019improving,mouatadid2023subseasonalclimateusa,vitart2022outcomes,watson2022climatebench,kaltenborn2023climateset}. We analyze the limitations of existing works, and propose how ChaosBench fills in these gaps (see Table \ref{tab:other_dataset}, more justifications in Appendix \ref{si:related_work}).

\textbf{Gap in forecast lead-time}. Many existing benchmarks are built for short/medium-range weather (up to 15 days) \cite{rasp2020weatherbench, kashinath2021climatenet,racah2017extremeweather}, and long-term climate (annual to decadal scale) \cite{watson2022climatebench}. As discussed earlier, these problems tend to be easier due to the lack of combined sensitivities to IC and BC \cite{prive2013role, wu2005estimating}.

\textbf{Limited spatiotemporal extent}. Many S2S benchmarks tend to focus on regional forecasts, such as the US \cite{hwang2019improving, mouatadid2023subseasonalclimateusa}. In addition, the temporal extent of observation with common interval is more varied, with some less than 20 years \cite{kashinath2021climatenet, vitart2022outcomes}. ChaosBench has the most extensive overlapping temporal coverage yet, extending to 45+ years of inputs covering multiple reanalysis products beyond ERA5. 

\textbf{Limited diversity of baseline models}. Having a large set of physics-based forecasts as baselines is key to reducing bias and diversifying the target goal-posts. Previous benchmarks are mostly focused on increasing the number of data-driven models for baselines \cite{rasp2020weatherbench, hwang2019improving}. In contrast, ChaosBench also places weights on expanding the diversity of physics-based models, including those operated by leading national weather agencies in the US, Europe, UK, and Asia.

\textbf{Lack of physics-based constraints}. So far, limited number of benchmarks have explicitly incorporated physical principles to improve or constrain forecasts. ChaosBench introduces physics-based metrics that can be used for comparison (\emph{scalar}) and integrated into ML pipeline (\emph{differentiable}).

\begin{table*}[t]
    \centering
    \caption{Comparison with other benchmark datasets: ChaosBench (ours) is evaluated on the largest set of global variables, benchmarked against large number of operational NWPs (four national agencies in the US, Europe, UK, and Asia), and incorporates both physics-based and probabilistic metrics for a more physically-consistent S2S ensemble forecast.}
    \begin{tabular}{c|c|c|c|c|c|c}
         \toprule
         Datasets & \rotatebox{90}{\makecell{\# input \\variables}} & \rotatebox{90}{\makecell{\# target \\variables}} & \rotatebox{90}{\makecell{forecast lead \\(days)}} & \rotatebox{90}{\makecell{physics-based \\ metrics}} & \rotatebox{90}{\makecell{probabilistic \\ metrics}}& \rotatebox{90}{\makecell{spatial extent}} \\
         \midrule
         % ClimateNet \cite{kashinath2021climatenet} & 6 & 2 & 1 & -  & \cmark & \xmark  & \xmark & global & 1995-2015\\
         % ExtremeWeather \cite{racah2017extremeweather} & 16 & 4 & 1 & -  & \cmark & \xmark  & \xmark & global & 1979-2005\\
         WeatherBench \cite{rasp2020weatherbench} & 110 & 110 & 15 & \cmark & \cmark  & global \\
         SubseasonalRodeo \cite{hwang2019improving} & $<$30 & 2 & 44 & \xmark & \xmark & western US \\
         SubseasonalClimateUSA \cite{mouatadid2023subseasonalclimateusa} & $<$30 & 2 & 44 & \xmark & \cmark & contiguous US \\
         CliMetLab \cite{vitart2022outcomes} & $<$30 & 2 & 44 & \xmark & \cmark & global \\
         \midrule
         \textbf{ChaosBench (ours)} & 124 & 124 & 44 & \cmark & \cmark & global\\
         \bottomrule
    \end{tabular}
    
    \label{tab:other_dataset}
\end{table*}

\section{ChaosBench}
\label{sec:dataset}

\subsection{Observations}
\label{observations}

We discuss the components of ChaosBench, including the global reanalysis products of surface-atmosphere (ERA5), sea-ice (ORAS5), and terrestrial (LRA5), as well as simulations from physics-based models. The spatiotemporal resolutions of the former are matched with the latter's daily forecasts at $1.5^\circ$ to allow for consistent evaluation and integration e.g., hybrid physics-based emulator. However, we provide a one-liner script to process higher e.g., $0.25^\circ$ resolution input in Section \ref{si-sec:multiresolution}.

\textbf{ERA5} Reanalysis provides a comprehensive record of the global atmosphere combining physics and observations for correction \cite{hersbach2020era5}. We processed their hourly data from 1979 to present and selected measurements at the 00UTC step. The variables include temperature ($t$), specific humidity ($q$), geopotential height ($z$), and 3D wind speed ($u,v,w$) at 10 pressure levels: $1000, 925, 850, 700, 500, 300, 200, 100, 50, 10$ hpa, totalling 60 variables (full list in \ref{si:reanalysis_era5}). 

\textbf{ORAS5} or the Ocean Reanalysis System 5 provides an extensive record of sea-ice variables that incorporate multiple depth levels \cite{zuo2019ecmwf}. Since the public data is available on a monthly basis, we replicate them for daily compatibility with temporal extent from 1979 to present, for a total of 21 variables, including \texttt{sst} and \texttt{ssh} (full list in \ref{si:reanalysis_oras5}). 

\textbf{LRA5} or ERA5-Land Reanalysis provides a detailed record of variables governing global terrestrial processes with specific corrections tailored for land surface applications such as flood forecasting \cite{munoz2021era5} or carbon fluxes \cite{nathaniel2023metaflux,nathaniel2023above}. We processed hourly data from 1979 to present and selected measurements at the 00UTC step, for a total of 43 variables, including \texttt{t2m}, \texttt{u10}, \texttt{v10}, and \texttt{tp} (full list in \ref{si:reanalysis_lra5}).

\subsection{Simulations}
\label{simulations}

\begin{wrapfigure}{r}{0.45\textwidth}
    \centering
    \includegraphics[width=0.45\textwidth]{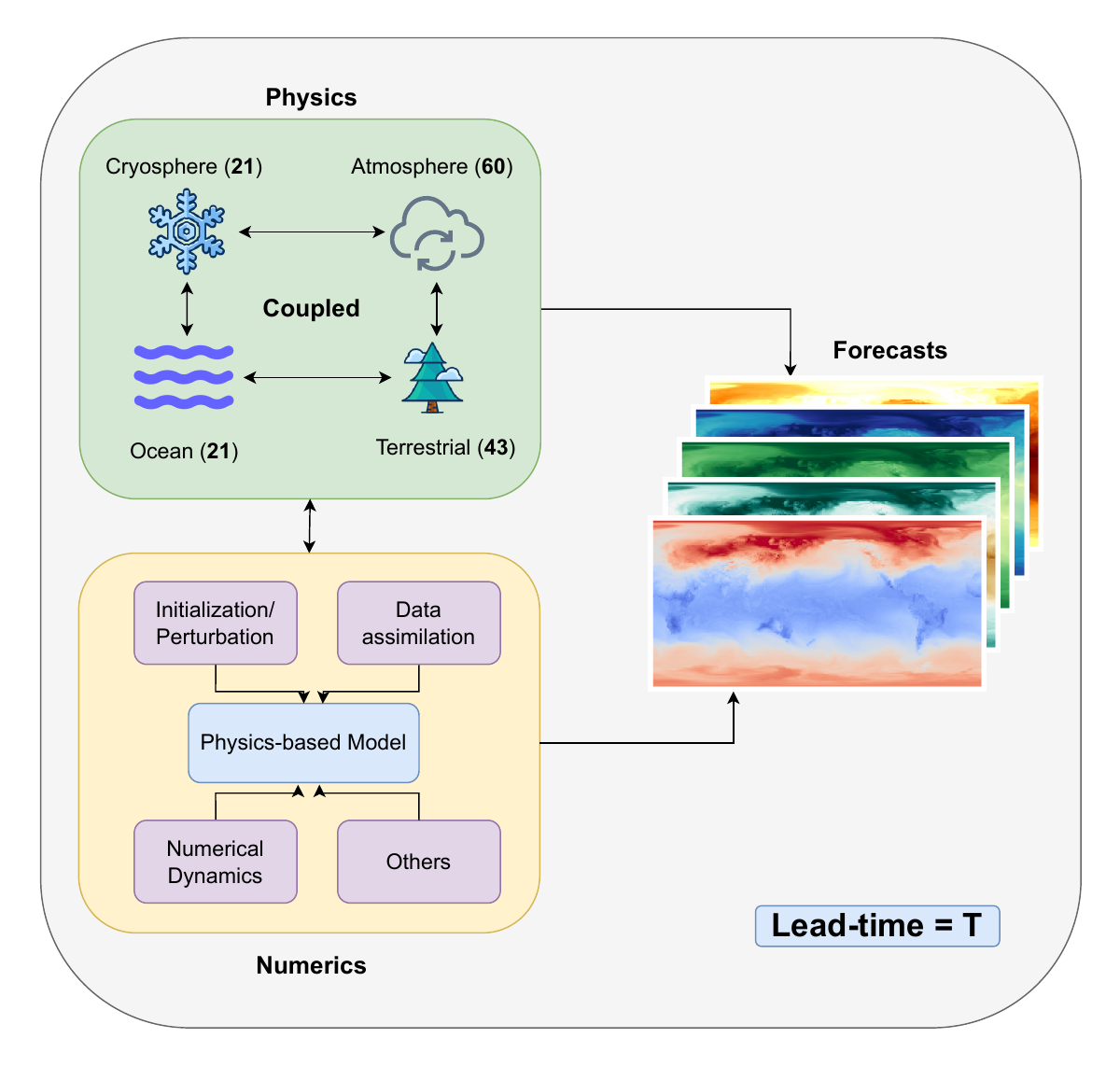}
    \caption{Physics-based simulations that couple different parts of the Earth system along with their operational choices such as data assimilation. The brackets are the number of variables provided in ChaosBench.}
    \label{fig:chaosbench_physics}
\end{wrapfigure}

We briefly describe the forecast generation process from physics-based models (Figure \ref{fig:chaosbench_physics}), including details on forecast frequency and the number of ensemble members. More details are provided in Appendix \ref{si:physics_model}. The list of available variables for physics-based forecast are similar to ERA5 but missing \{q10,q50,q100\} and w $\notin$ \{w500\} for a total of 48 variables. In all, we process control (deterministic) and perturbed (ensemble) forecasts from 2016 to present \cite{vitart2017subseasonal}.

\textbf{UKMO}. The UK Meteorological Office uses the Global Seasonal Forecast System Version 6 (GloSea6) model \cite{williams2015met} to generate daily 3+1 ensemble/control forecasts for 60-day lead time. 
    
\textbf{NCEP}. The National Centers for Environmental Prediction uses the Climate Forecast System 2 (CFSv2) model \cite{saha2014ncep} to generate daily 15+1 ensemble/control forecast for 45-day lead time. 

\textbf{CMA}. The China Meteorological Administration uses the Beijing Climate Center (BCC) fully-coupled BCC-CSM2-HR model \cite{wu2019beijing} to generate 3+1 ensemble/control forecasts at 3-day interval for 60-day lead time. 

\textbf{ECMWF}. The European Centre for Medium-Range Weather Forecasts uses the operational Integrated Forecasting System (IFS) that includes advanced data assimilation strategies and global numerical model of the Earth system \cite{documentationpart}. In particular, we use the CY41R1 version of the IFS to generate 50+1 ensemble/control forecasts twice weekly for 46-day lead time. 

\subsection{Auxiliary}
\label{auixiliary}
In addition to baseline forecasts from physics-based and data-driven models, we provide additional auxiliary data and baselines. This includes \textbf{climatology}, the long-term weather-state statistics, and \textbf{persistence}, which uses initial observation for subsequent rollouts.

\begin{figure*}[t!]
    \centering
    \begin{subfigure}{0.55\textwidth}
        \includegraphics[width=\textwidth]{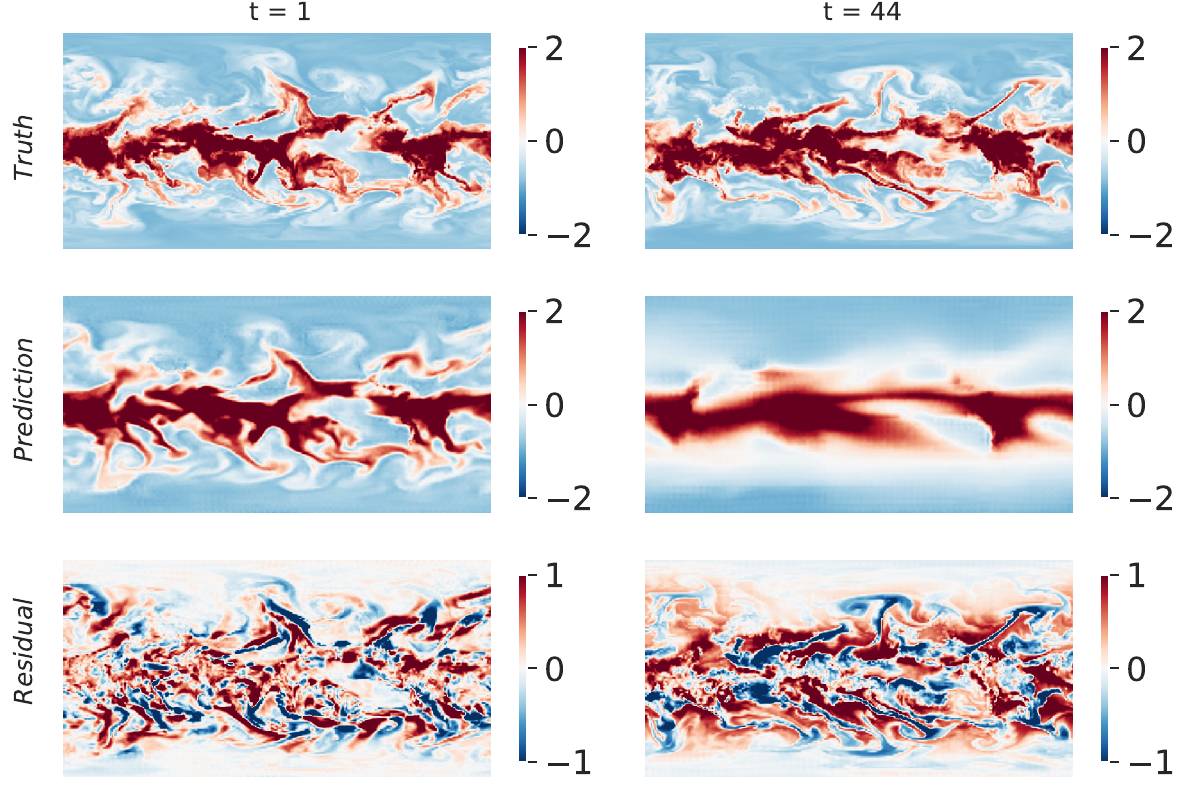}
        \caption{Normalized humidity@700-hpa label, forecast, and residual at the first ($t=1$) and final ($t=44$) step with ClimaX}
        \label{fig:signal_loss}
    \end{subfigure}
    \hfill
    \begin{subfigure}{0.4\textwidth}
        \includegraphics[width=\textwidth]{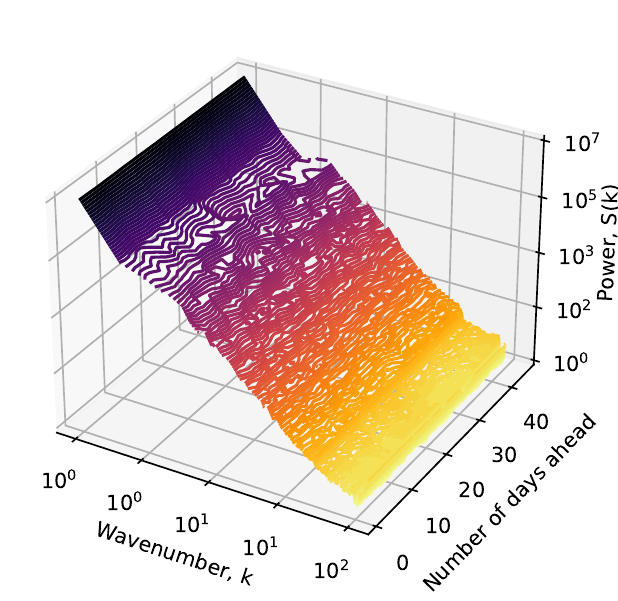}
        \caption{Power spectrum $S(k)$ \emph{vs.} wavenumber $k$ plot as a function of prediction step of normalized humidity@700-hpa with ClimaX}
        \label{fig:spectra_3d}
    \end{subfigure}
    
\caption{Motivating problem: as we perform longer rollouts, the (a) residual error becomes larger and prediction becomes blurry. This behavior is captured in the Fourier frequency domain where the (b) power spectra $S(k)$ at low wavenumber $k$ (i.e., low frequency signal) remains consistent at long rollouts, but not for higher $k$ (i.e., high frequency signal). This phenomenon explains why long-term forecasts excel at capturing \emph{large-scale pattern} but not \emph{fine-grained details} i.e., smooth.}
\label{fig:challenge}
\end{figure*}

\section{Benchmark Metrics}
We provide an assortment of metrics, which we divide into deterministic, probabilistic, and several proposed physics-based criteria, for increased explainability. For each metric, unless otherwise noted, we apply a weighting scheme at each latitude $\theta_i$ as defined by Equation \ref{eq:weight}.

\begin{equation}
\label{eq:weight}
    w(\theta_i) = \frac{cos(\theta_i)}{\frac{1}{|\boldsymbol{\theta}|}\sum_{a=1}^{|\boldsymbol{\theta}|}cos(\theta_a)}
\end{equation}

where $\boldsymbol{\theta}$ is the set of all latitudes in our data, and $|\boldsymbol{\theta}|$ is its cardinality. We denote the input at time $t$ as $\mathbf{X}_t \in \mathbb{R}^{h \times w \times p }$, where $h, w, p$ represent the height (i.e., latitude), width (i.e., longitude), and parameter (e.g., temperature) with its associated vertical level (e.g., 1000-hpa or surface). In addition, we denote $\{\mathbf{Y}_t, \mathbf{\hat{Y}}_t\} \in \mathbb{R}^{h \times w \times p }$ as the ground-truth label and prediction respectively. Finally, we denote each element of latitude and longitude as $\theta_i \in \boldsymbol{\theta}$ and $\gamma_j \in \boldsymbol{\gamma}$.

\subsection{Deterministic Metrics}
We provide popular deterministic metrics in the machine learning and climate science literature alike, including RMSE, Bias, ACC, and MS-SSIM. 

\noindent\textbf{Root Mean Squared Error (RMSE)} is useful to penalize outliers, which are especially critical for weather and climate applications such as extreme event prediction (Equation \ref{eq:rmse}). 

\noindent\textbf{Bias} assists us to identify misspecification and systematic errors present in the model (Equation \ref{eq:bias}).

\noindent\textbf{Anomaly Correlation Coefficient (ACC)} measures the correlation between predicted and observed anomalies. This metric is especially useful in weather and climate applications, where deviations from the norm (e.g., temperature anomalies) often reveal interesting insights (Equation \ref{eq:acc}).

\noindent\textbf{Multi-Scale Structural Similarity (MS-SSIM)} \cite{wang2003multiscale} compares structural similarity between forecast and ground-truth label across scales (refer to Appendix \ref{si:ms-ssim} for more details). This is especially useful in weather systems because they occur at multiple scales, from large systems like cyclones, to smaller features like localized rain thunderstorms.

\subsection{Physics Metrics}
As illustrated in Figure \ref{fig:challenge}, we find that in general, data-driven forecasts tend to become blurry (Figure \ref{fig:signal_loss}) due to power divergence in the spectral domain (Figure \ref{fig:spectra_3d} + \ref{si-fig:specdiv_climax}). This motivates us to propose two physics-based metrics that measure the deviation or difference between the power spectra of prediction ${\hat{S}}(k)$ and target $S(k)$, where $k \in \mathbf{K}$, and $\mathbf{K}$ is the set of all scalar wavenumbers from 2D Fourier transform. Focusing on high-frequency components, we introduce $\mathbf{K}_q = \{k \in \mathbf{K} \mid k \geq Q(q)\}$, where $Q$ is the quantile function of $\mathbf{K}$ and $q \in [0,1]$. We set $q=0$ or $q=0.9$ for training and evaluation respectively. We denote $S_q = \{S(k) \mid k \in \mathbf{K}_q\}$ as the corresponding power spectra on $\mathbf{K}_q$, and we normalize the distribution to $S^\prime(k)$ such that it sums up to 1. Similarly we use $\hat{S}^\prime(k)$ to denote the normalized power for predictions.

\begin{figure*}[h!]
    \centering
    \begin{subfigure}{\textwidth}
        \includegraphics[width=\textwidth]{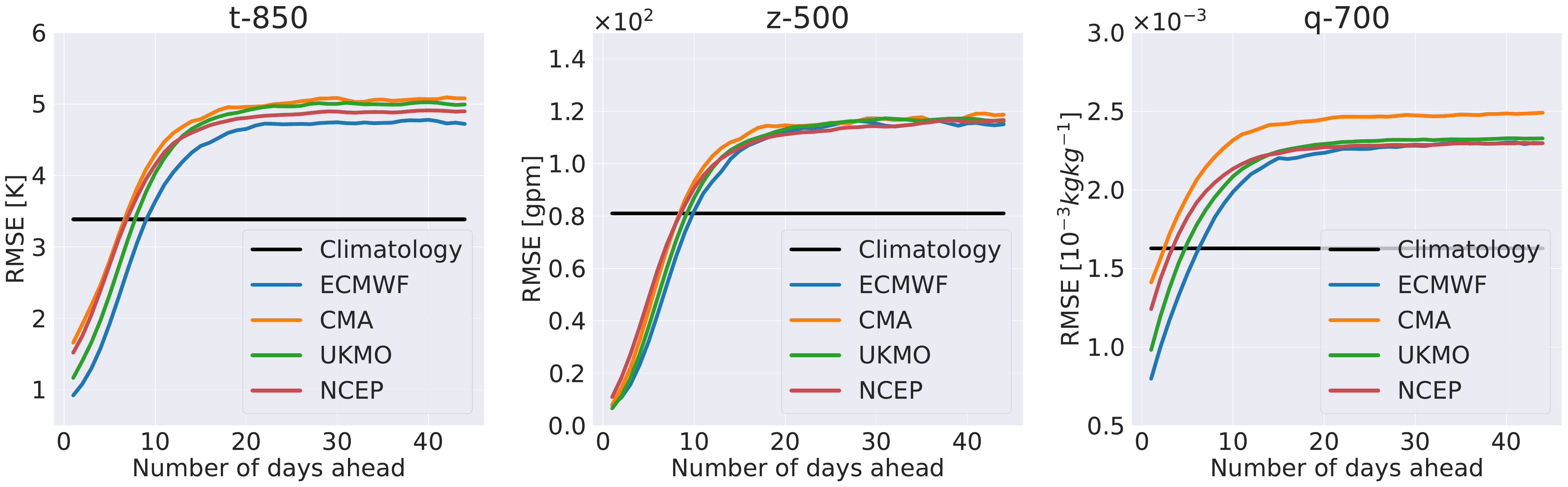}
        \caption{RMSE ($\downarrow$ is better)}
    \end{subfigure}
    \hfill
    \begin{subfigure}{\textwidth}
        \includegraphics[width=\textwidth]{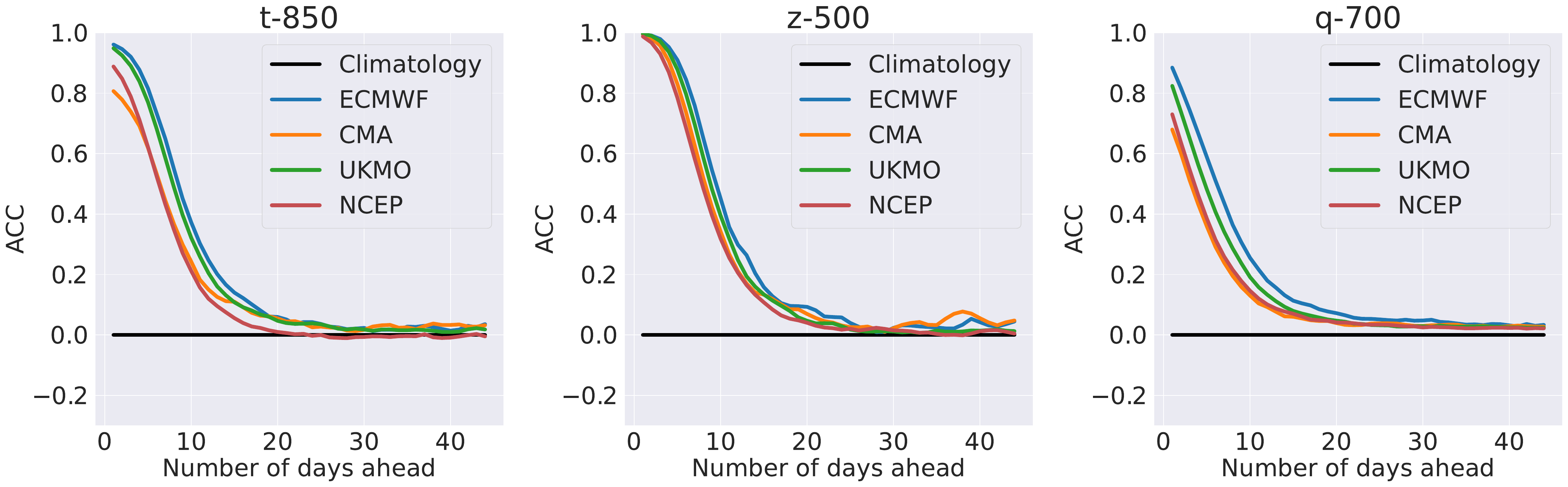}
        \caption{ACC ($\uparrow$ is better)}
    \end{subfigure}
    \hfill
    \begin{subfigure}{\textwidth}
        \includegraphics[width=\textwidth]{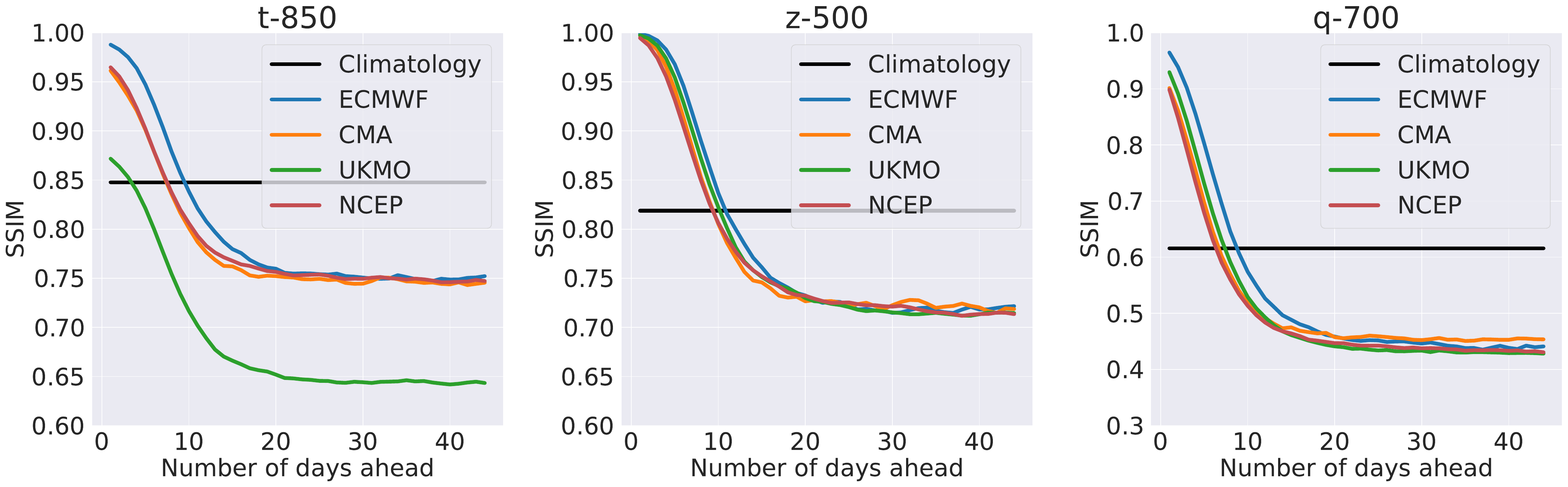}
        \caption{MS-SSIM ($\uparrow$ is better)}
    \end{subfigure}
    
\caption{Evaluation results between baseline climatology (black line) and physics-based control/deterministic forecasts. At longer forecasting horizon, most physics-based control/deterministic forecasts perform worse than climatology.}
\label{fig:all_results_centers}
\end{figure*}

\begin{figure*}[h!]
    \centering
    \begin{subfigure}{\textwidth}
        \includegraphics[width=\textwidth]{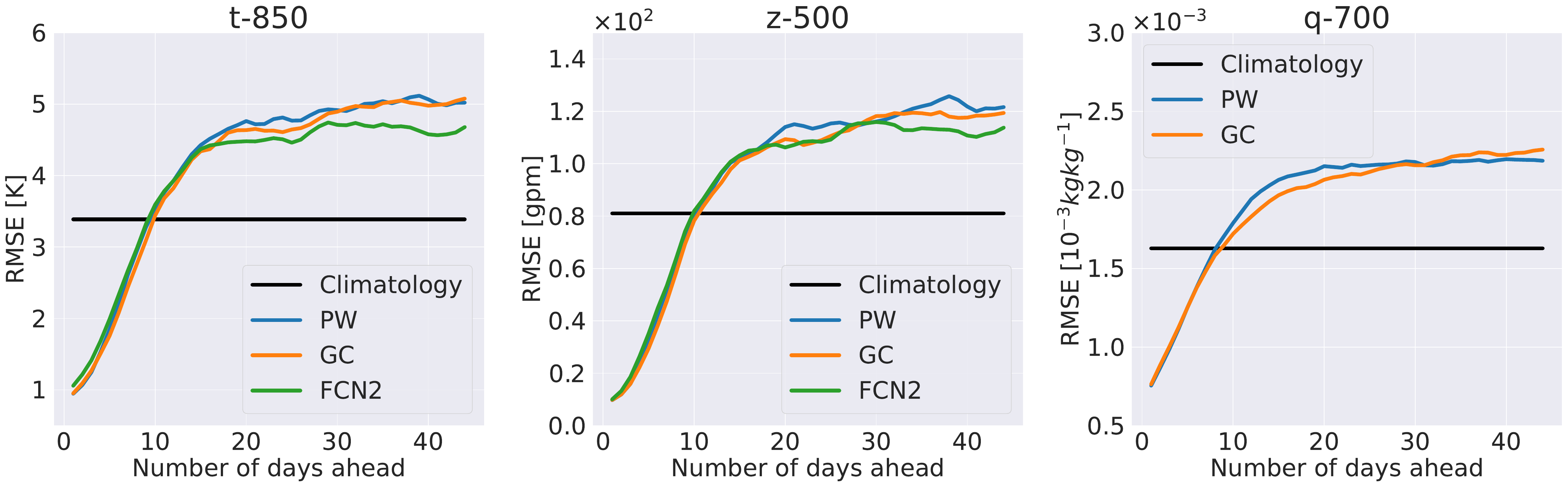}
        \caption{RMSE ($\downarrow$ is better)}
    \end{subfigure}
    \hfill
    \begin{subfigure}{\textwidth}
        \includegraphics[width=\textwidth]{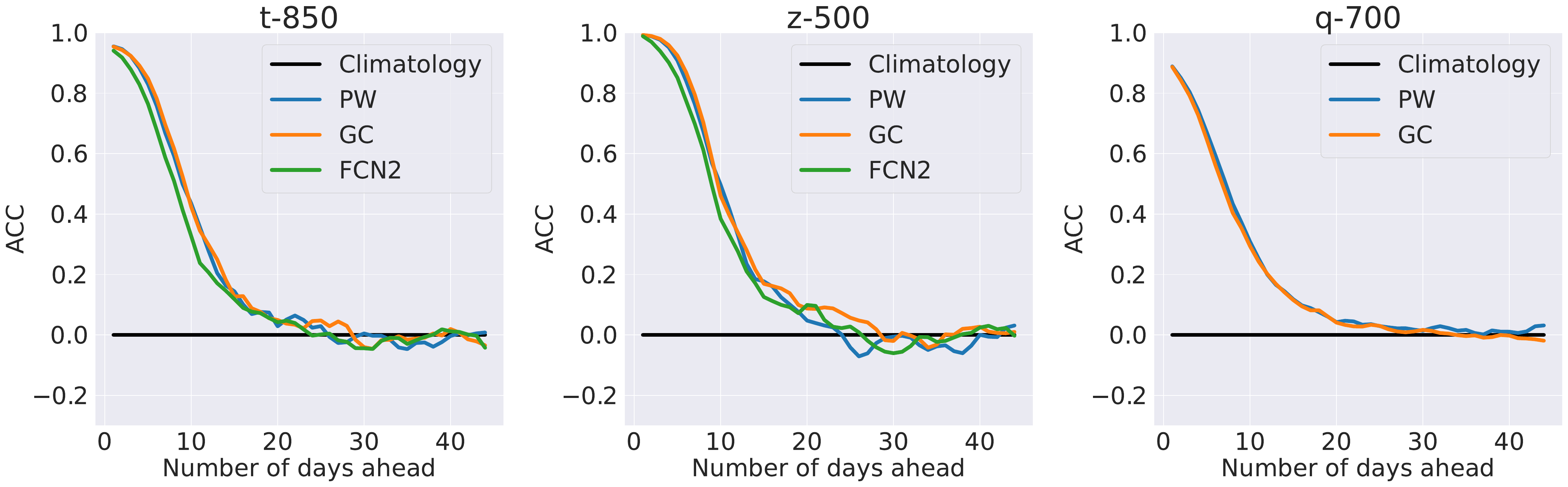}
        \caption{ACC ($\uparrow$ is better)}
    \end{subfigure}
    \hfill
    \begin{subfigure}{\textwidth}
        \includegraphics[width=\textwidth]{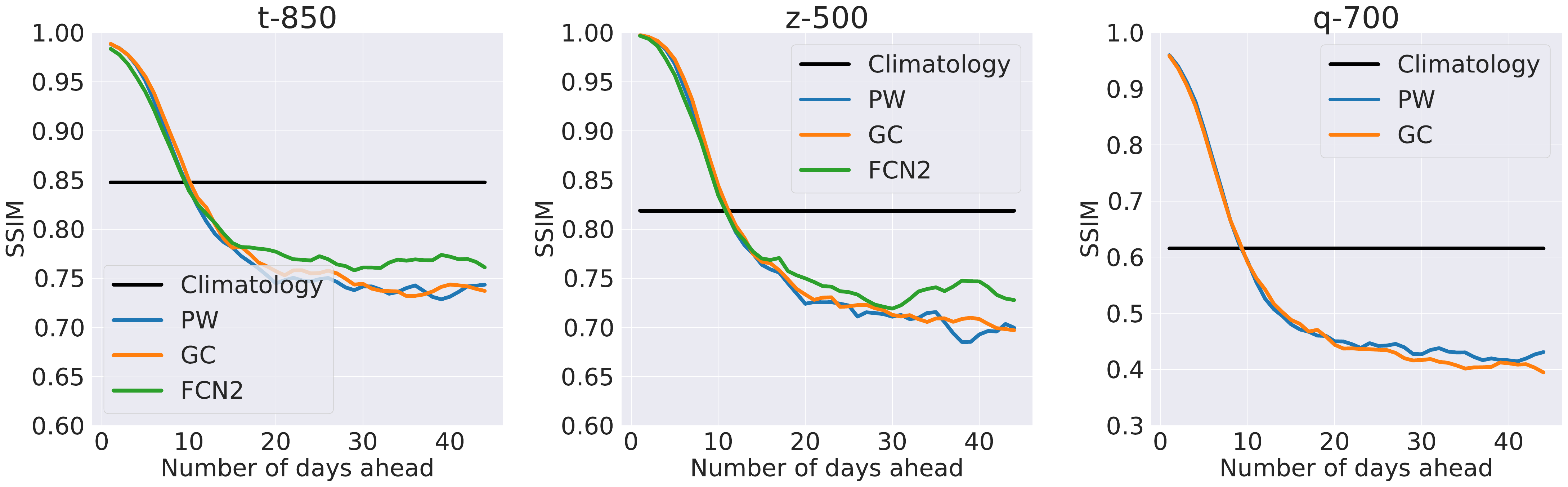}
        \caption{MS-SSIM ($\uparrow$ is better)}
    \end{subfigure}
    
\caption{Evaluation results between baseline climatology (black line) and data-driven models including PanguWeather (PW), GraphCast (GC), and FourCastNetV2 (FCN2). We find that deterministic ML models perform worse than climatology on S2S timescale. \underline{Note}: FCN2 lacks q-700.}
\label{fig:all_results_sota}
\end{figure*}

\noindent\textbf{Spectral Divergence (SpecDiv)} follows principles from Kullback–Leibler (KL) divergence \cite{kullback1951information} where we compute the expectation of the log ratio between target $S^\prime(k)$ and prediction $\hat{S}^\prime(k)$ spectra, and is defined in Equation \ref{eq:specdiv} (see Listing \ref{lst:specdiv} for \textsc{PyTorch} psuedocode). 

\begin{equation}
    \mathcal{M}_{SpecDiv} = \sum_{k} S^{\prime}(k) \cdot \log(S^{\prime}(k) / \hat{S}^{\prime}(k))
    \label{eq:specdiv}
\end{equation}

\noindent\textbf{Spectral Residual (SpecRes)} follows principles from RMSE and adapted from \cite{takamoto2022pdebench} where we compute the root of the expected squared residual, and is defined in Equation \ref{eq:specres} (see Listing \ref{lst:specres} for \textsc{PyTorch} psuedocode).

\begin{equation}
    \mathcal{M}_{SpecRes} = \sqrt{\mathbb{E}_k [(\hat{{S}}^\prime(k) - {S}^\prime(k))^2]}
    \label{eq:specres}
\end{equation}

The expectations are calculated over $\mathbf{K}_q$. For both physics-based metrics, the value will be zero if the power spectra of the forecast is identical to the target, but will increase as discrepancy emerges. Essentially, both metrics measure how well the forecasts \emph{preserve} signals across the frequency spectrum.

\subsection{Probabilistic Metrics}
In addition to the probabilistic version of RMSE, Bias, ACC, MS-SSIM, SpecDiv, and SpecRes where we take their expectation with respect to the ensemble members (Equations \ref{eq:rmse_ens}-\ref{eq:specres_ens}), we also use several probabilistic metrics to evaluate ensemble forecasts critical for long-range S2S prediction.

\textbf{Continuous Ranked Probability Score (CRPS)} evaluates the accuracy of the ensemble distribution against the target. Low CRPS values require forecasts to be reliable, where the predicted uncertainty aligns with the actual uncertainty, and a smaller uncertainty is preferable (Equation \ref{eq:crps}).

\textbf{Continuous Ranked Probability Skill Score (CRPSS)} evaluates the skill of probabilistic forecast relative to climatology variability; CRPSS > 0 suggests skillfulness and vice versa (Equation \ref{eq:crpss}).

\textbf{Spread} quantifies the uncertainty in ensemble forecasts by measuring the variability among ensemble members, which helps to understand the range of possible outcomes and confidence (Equation \ref{eq:spread}).

\textbf{Spread/Skill Ratio} balances the ensemble spread with the forecast skill (e.g., RMSE); ideally, a well-calibrated ensemble should have a spread that matches the forecast skill (Equation \ref{eq:ssr}).

\begin{figure*}[h]
    \centering
    \begin{subfigure}{\textwidth}
        \includegraphics[width=\textwidth]{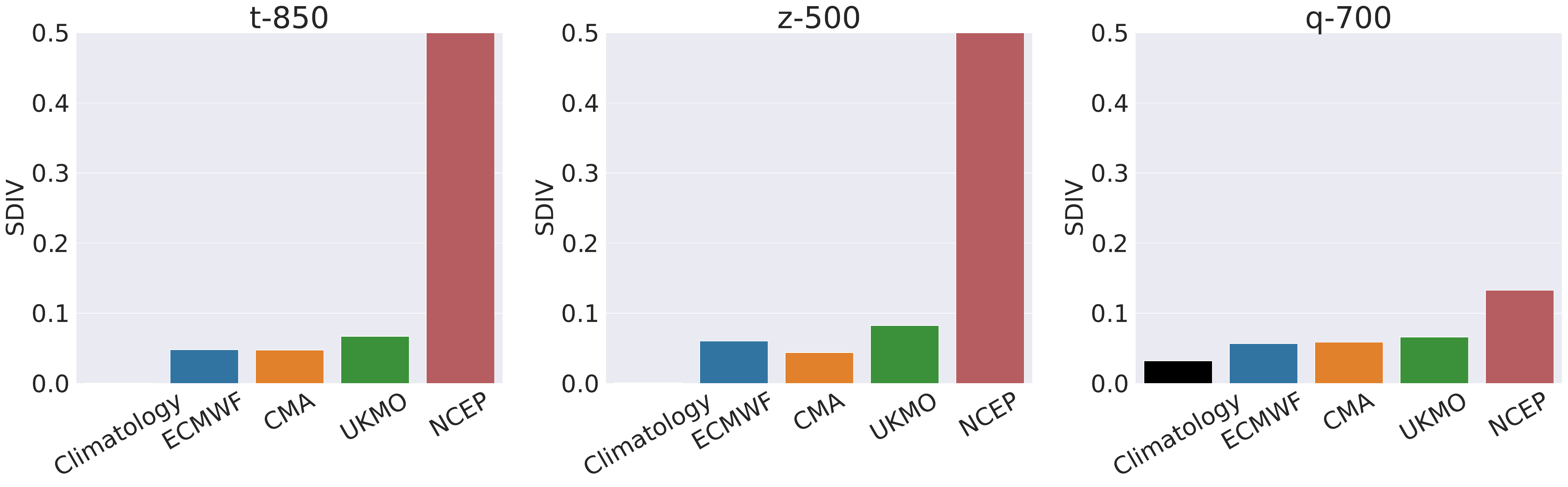}
        \caption{SpecDiv ($\downarrow$ is better) for physics-based models}
    \end{subfigure}
    \hfill
    \begin{subfigure}{\textwidth}
        \includegraphics[width=\textwidth]{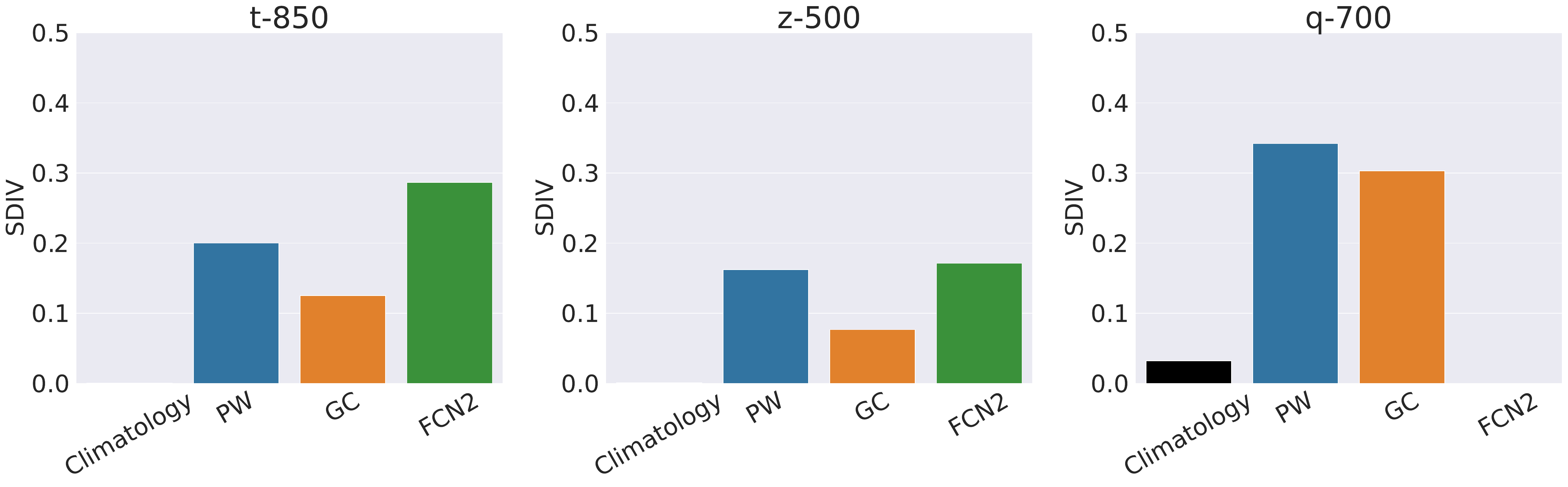}
        \caption{SpecDiv ($\downarrow$ is better) for data-driven models}
    \end{subfigure}
    
\caption{Spectral divergence between (a) physics-based, and (b) data-driven models. Overall, we observe that the latter perform worse than their physics-based counterpart (barring NCEP) on time-averaged spectral divergence. \underline{Note}: FCN2 lacks q-700.}
\label{fig:all_sdiv}
\end{figure*}

\section{Benchmark Results}
Throughout this section, we report headline results on $\hat{\mathbf{X}} \in \{\text{t-850, z-500, q-700}\}$, following Weatherbench v2 \cite{rasp2023weatherbench}. The full benchmark scores are available at \href{https://leap-stc.github.io/ChaosBench}{https://leap-stc.github.io/ChaosBench}. We primarily use four state-of-the-art models for comparison including ViT/ClimaX \cite{nguyen2023climax}, PanguWeather \cite{bi2022pangu}, GraphCast, and FourCastNetV2 \cite{pathak2022fourcastnet} \cite{lam2022graphcast}. However, whenever ablation is performed, we use popular baselines including Lagged Autoencoder \cite{lusch2018deep}, ResNet \cite{rasp2021data}, UNet \cite{rasp2020weatherbench}, and FNO \cite{li2020fourier} trained on 1979-2015 data and validated on 2016-2021 data. All evaluations presented here are done on the held-out 2022 data. The full implementation details are discussed in Appendix \ref{si:data_model}. 

\begin{figure*}[t]
    \centering
    \begin{subfigure}{\textwidth}
        \includegraphics[width=\textwidth]{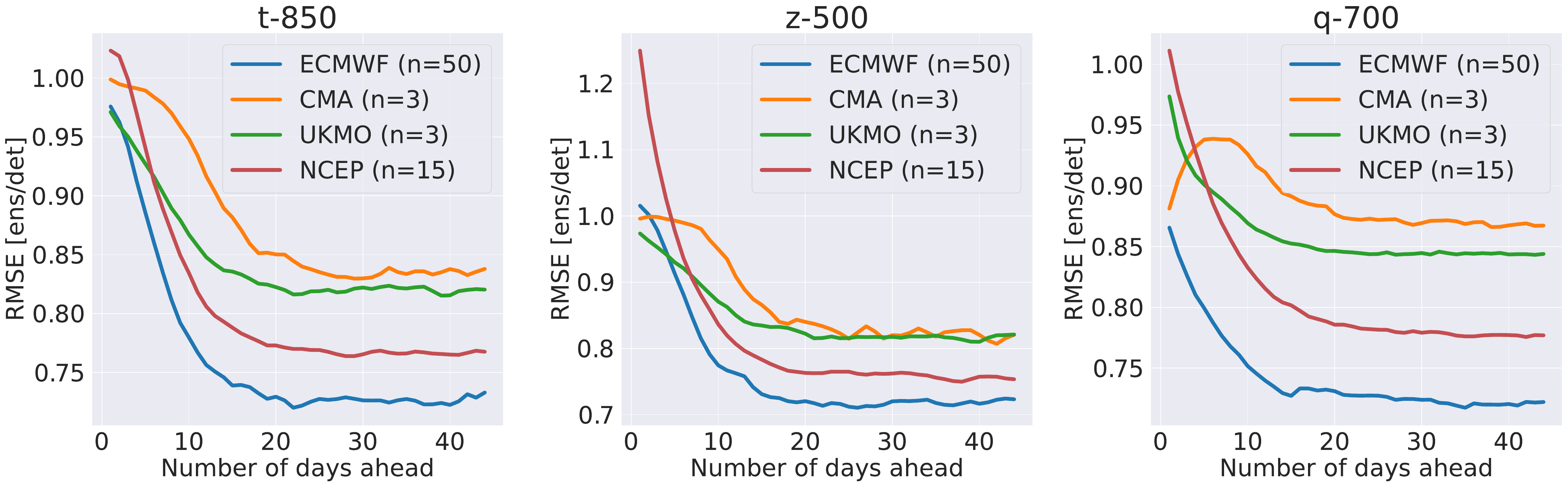}
        \caption{RMSE: ensemble improves deterministic forecasts if $\texttt{ratio} < 1$}
    \end{subfigure}
    \hfill
    \begin{subfigure}{\textwidth}
        \includegraphics[width=\textwidth]{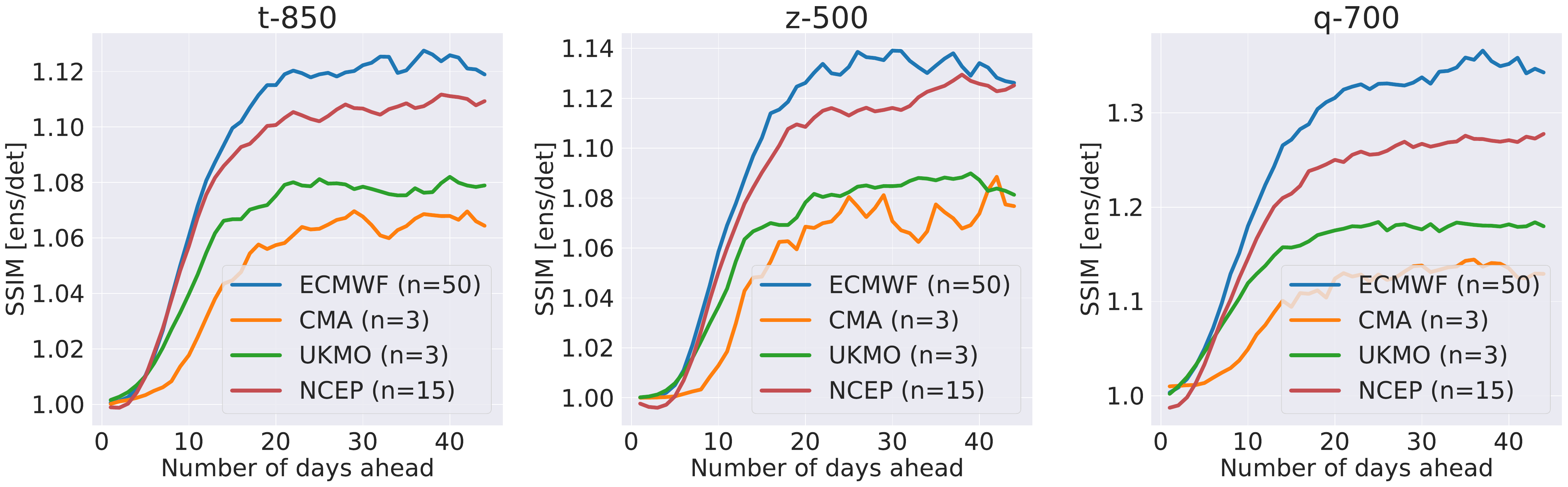}
        \caption{MS-SSIM: ensemble improves deterministic forecasts if $\texttt{ratio} > 1$}
    \end{subfigure}
    
\caption{Metrics ratio e.g., $\texttt{RMSE}_{ens} / \texttt{RMSE}_{det}$ between ensemble and deterministic forecasts, where the former improves the latter by accounting for IC uncertainty that can lead to trajectory divergences. \underline{Note}: $n$ represents the number of ensemble members.}
\label{fig:center_ratio}
\end{figure*}

\begin{table}[h!]
    \centering
    \caption{Performance metrics for SoTAs with different training strategies, at $\Delta t = 44$}
    \begin{tabular}{c|c|c|cccc}
        \toprule
        \textbf{Metrics} & \textbf{Variables} &  \textbf{Reference} & \multicolumn{3}{c}{\textbf{Autoregressive}} & \textbf{Direct} \\
        \cmidrule(r){3-7}
         &  & \textbf{Climatology} & \textbf{PW} & \textbf{GC} & \textbf{FCN2} & \textbf{ViT/ClimaX} \\
        \midrule
        \multirow{3}{*}{\textbf{RMSE} $\downarrow$} 
         & t-850 (K) & 3.39 & 5.85 & 5.87 & 5.11 & \textbf{3.56}\\
         & z-500 (gpm) & 81.0 & 120.9 & 136.0 & 112.4 & \textbf{83.1} \\
         & q-700 ($\times 10^{-3}$) & 1.62 & 2.35 & 2.28 & - & \textbf{1.66} \\
         \midrule
        \multirow{3}{*}{\textbf{MS-SSIM} $\uparrow$} 
         & t-850 & 0.85 & 0.70 & 0.70 & 0.74 & \textbf{0.83} \\
         & z-500 & 0.82 & 0.68 & 0.66 & 0.72 & \textbf{0.81}\\
         & q-700 & 0.62 & 0.43 & 0.45 & - & \textbf{0.59}\\
         \midrule
        \multirow{3}{*}{\textbf{SpecDiv} $\downarrow$}
         & t-850 & 0.01 & 0.25 & \textbf{0.05} & 0.28 & 0.20\\
         & z-500 & 0.01 & 0.33 & \textbf{0.03} & 0.11 & 0.13 \\ 
         & q-700 & 0.03 & \textbf{0.23} & 0.27 & - & 0.28\\
        \bottomrule
    \end{tabular}
    \label{tab:training_strategies}
\end{table}

\textbf{Collapse in Predictive Skill}. As shown in Figure \ref{fig:all_results_centers} (+ \ref{si-fig:all_results_centers}), control forecasts from various operational centers perform worse than climatology at the S2S scale beyond 15 days. A similar phenomenon of skill collapse is evident in data-driven models, as depicted in Figure \ref{fig:all_results_sota} (+ \ref{si-fig:all_results_sota}). Unlike their physics-based counterparts, these forecasts exhibit significantly higher spectral divergence as evidenced in Figure \ref{fig:all_sdiv}, indicating low predictive skill for multi-scale structures over long rollouts. This leads to the blurring artifacts previously discussed. The pervasive lack of predictive skill underscores the notoriously difficult challenge of S2S forecasting and highlights huge potential for improvement.

\textbf{Ensemble Forecasts Account for IC Uncertainty}. Despite the underperformance of deterministic models, many studies have highlighted the potential of ensemble forecasts to account for trajectory divergences caused by IC uncertainties \cite{leith1974theoretical,weyn2021sub,chen2024machine}, also known as the butterfly effect \cite{lorenz1963deterministic}. Figure \ref{si-fig:center_ens} shows that the performance of ensembles across physics-based models improves relative to their deterministic counterparts. For instance, when we take the metrics ratio between ensemble and deterministic forecasts as in Figure \ref{fig:center_ratio} (+ \ref{si-fig:center_ratio}), the ratio of RMSE decreases with lead time, while the ratio of MS-SSIM improves over time with little significant changes in SpecDiv. The extent of improvement also appears to be affected by the number of ensemble members i.e., higher ensemble size $n$ appears to improve skillfulness. We also note similar insights from data-driven ensembling strategy as discussed in Section \ref{si-sec:ml_ensemble}. This highlights the importance of building a well-dispersed ensemble that accounts for long-range divergences for improved S2S predictability.

\textbf{Minimizing Error Propagation Promotes Stability}. Different training and inference strategies have been proposed to improve the accuracy and stability of data-driven weather emulators. Chief among these are the autoregressive and direct approaches \cite{nguyen2023scaling}. The former iteratively cycles through small interval to reach the target lead-time i.e., $\Delta t = N\delta t$ where $N \in \mathbb{Z}^+$ is the number of such compositions, while the latter directly outputs $\Delta t$. As summarized in Table \ref{tab:training_strategies}, we find models trained directly (e.g., ViT/ClimaX) have better performance than those used autoregressively (e.g., PW, GC, FCN2). This suggests that error propagation is a significant source of error, and controlling for stability is key to extend the predictability range of weather emulators. Once stability is achieved, the remaining sources of errors including uncertainties in observation and/or modeling framework can be improved through more data, better model, or both through data assimilation for instance \cite{qu2024deep}. 

\begin{figure*}[t]
    \centering
    \includegraphics[width=\textwidth]{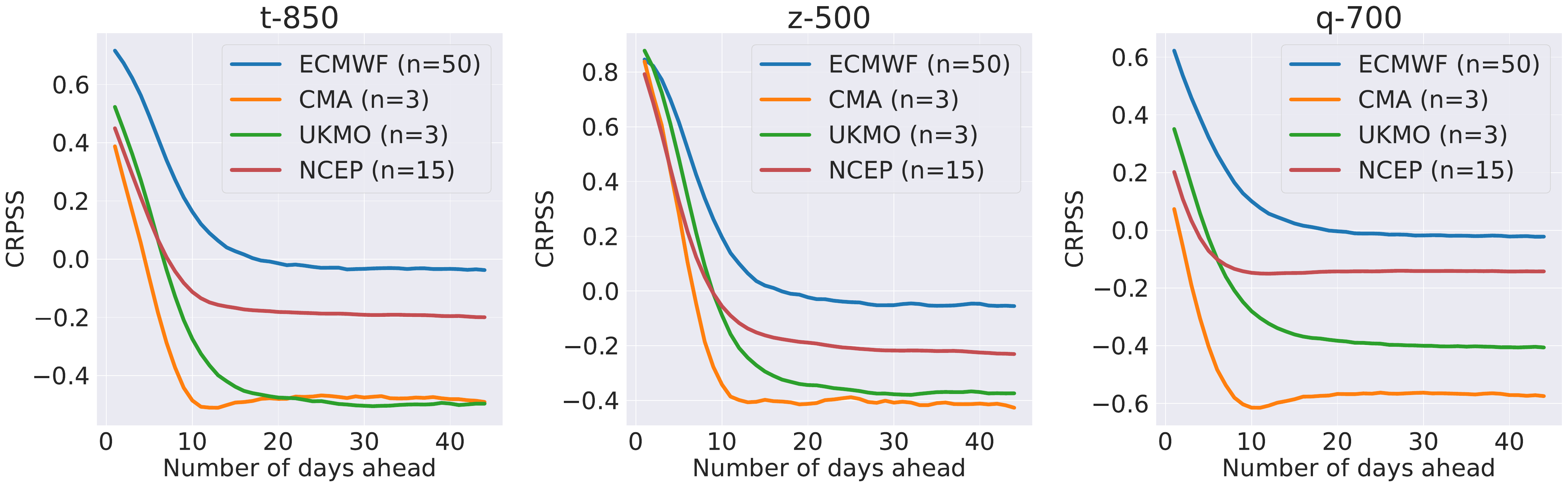}
    
\caption{Probabilistic evaluation on ensemble forecasts indicating current skill limits of 15-20 days; CRPSS $>0$ suggests skills better than climatology variability. \underline{Note}: $n$ represents the number of ensemble members.}
\label{fig:center_probs}
\end{figure*}

% \textbf{Boundary Condition Extends Predictability Range}.

% \begin{table}[h!]
%     \centering
%     \caption{Performance metrics for ResNet and UNet with and without boundary conditions at $\Delta t = 44$}
%     \begin{tabular}{c|c|cc|cc}
%         \toprule
%         \textbf{Metrics} & \textbf{Variables} & \multicolumn{2}{c}{\textbf{ResNet}} & \multicolumn{2}{c}{\textbf{UNet}} \\
%         \cmidrule(r){3-4} \cmidrule(r){5-6}
%          &  & \textbf{w/o BC} & \textbf{with BC} & \textbf{w/o BC} & \textbf{with BC} \\
%         \midrule
%         \multirow{3}{*}{\textbf{RMSE} $\downarrow$} 
%          & t-850 (K) & 12.2 & \textbf{8.96} & 11.8 & 11.8 \\
%          & z-500 (gpm) & 258 & \textbf{172} & \textbf{248} & 250 \\
%          & q-700 ($\times 10^{-3}$) & 2.98 & \textbf{2.95} & 3.17 & 3.24 \\
%          \midrule
%         \multirow{3}{*}{\textbf{ACC} $\uparrow$} 
%          & t-850 & 0.60 & \textbf{0.74} & 0.59 & \textbf{0.65} \\
%          & z-500 & 0.64 & \textbf{0.78} & 0.63 & \textbf{0.67} \\
%          & q-700 & 0.41 & \textbf{0.44} & 0.13 & \textbf{0.31} \\
%          \midrule
%         \multirow{3}{*}{\textbf{MS-SSIM} $\uparrow$}
%          & t-850 & 0.55 & \textbf{0.59} & 0.40 & \textbf{0.42} \\
%          & z-500 & 0.56 & \textbf{0.62} & \textbf{0.46} & 0.41 \\ 
%          & q-700 & 0.25 & \textbf{0.26} & 0.00 & \textbf{0.14} \\
%         \bottomrule
%     \end{tabular}
%     \label{tab:boundaries}
% \end{table}

\textbf{Physical Constraints Yield Improved Performance}. We find models that explicitly incorporate physical knowledge (e.g., learning spectral signals beyond pixel information) have better performance across metrics, such as FNO, as summarized in Table \ref{si-tab:ablation_s2s} given identical parameter budget of $10^6$. This phenomena is unsurprising and has been repeatedly demonstrated in many real-world applications of physics-informed deep learning, for instance.

\textbf{Current Limits of S2S Predictability}. Given our best models, we evaluate the extent of predictability in order to base our next steps. As illustrated in Figure \ref{fig:center_probs} (+ \ref{si-fig:center_probs}), we find that ECMWF high-resolution ensemble, dubbed as the gold standard, still has the best performance in terms of CRPSS (vs ERA5 climatology), with a predictability range of around 15-20 days ahead before its skill collapses to climatology (i.e., CRPSS $\rightarrow$ 0). However, the resurgence of data-driven models are rapidly transforming the field as they are able to efficiently distil knowledge and automatically discover emergent patterns from large-scale, high-dimensional dataset, instead of first reducing them to physical functions with limited set of variables requiring constant calibration as is traditionally done in NWPs. The challenge, therefore, is to extend the predictability range of weather system as a representation of large-scale chaos, and we welcome the machine learning communities to take part in this open effort.  

\section{Conclusion}
We present ChaosBench, a challenging benchmark to extend the predictability range of weather emulators into the S2S timescale where many processes with significant socioeconomic repercussions tend to occur, including extreme events. In addition to providing diverse datasets beyond ERA5 for a full Earth system emulation, we also perform extensive benchmarking on state-of-the-art data-driven and physics-based models alike. Through various ablation, we systematically find that skillfulness can be extended by ensemble forecasting, controlling for exponential error growth, and incorporating physical knowledge in our modeling approaches.

\textbf{Future Work}. Our input datasets have relatively coarse spatiotemporal resolution to match that of physics-based S2S forecasts. Nevertheless, we make the data processing pipeline open-source, allowing users to easily process inputs of the desired resolution (see Section \ref{si-sec:multiresolution} for more details). We are planning for a multi-source  reanalysis products (e.g., MERRA-2 \cite{gelaro2017modern}), leveraging diverse dataset strengths, such as the assimilation of different set of observations. As always, we welcome any contribution from the open-source community to solve this important yet understudied problem. And any comments, feedback, and/or future feature requests can be directed to the corresponding author or through the Github issue tracker at \href{https://github.com/leap-stc/ChaosBench}{https://github.com/leap-stc/ChaosBench}.

\begin{ack}
We would like to thank Matthew Wilson, Tom Andersson, and Dale Durran for the insightful discussion during the earlier version of the manuscript. The authors also acknowledge funding, computing, and storage resources from the NSF Science and Technology Center (STC) Learning the Earth with Artificial Intelligence and Physics (LEAP) (Award \#2019625) and the Department of Energy (DOE) Advanced Scientific Computing Research (ASCR) program
(DE-SC0022255). AG would like to acknowledge support from Google and Schmidt Sciences. Last but definitely not least, we acknowledge the comprehensive S2S database emerging from the joint initiative of the World Weather Research Programme (WWRP) and the World Climate Research Programme (WCRP). The original S2S database is hosted at ECMWF as an extension of the TIGGE database.
\end{ack}

\bibliography{neurips_data_2024} 
\newpage

\appendix
\setcounter{section}{0}
\setcounter{equation}{0}
\setcounter{figure}{0}
\setcounter{table}{0}
\setcounter{lstlisting}{0}

\renewcommand{\thesection}{\Alph{section}}
\renewcommand\thefigure{S\arabic{figure}}
\renewcommand\thetable{S\arabic{table}}
\renewcommand\theequation{S\arabic{equation}}
\renewcommand\thelstlisting{S\arabic{lstlisting}}

\onecolumn
{
    \centering
    \Large
    \textbf{ChaosBench: A Multi-Channel, Physics-Based Benchmark for Subseasonal-to-Seasonal Climate Prediction}\\
    \vspace{0.5em}Supplementary Material \\
    \vspace{0.5em}
    {\normalsize
    \makebox[0.95\textwidth]{\textbf{Juan Nathaniel}\textsuperscript{1,*}, \textbf{Yongquan Qu}\textsuperscript{1}, \textbf{Tung Nguyen}\textsuperscript{2}, \textbf{Sungduk Yu}\textsuperscript{3,5}, \textbf{Julius Busecke}\textsuperscript{1,4},}\\
    \makebox[0.95\textwidth]{\textbf{Aditya Grover}\textsuperscript{2}, \textbf{Pierre Gentine}\textsuperscript{1}}\\
    \makebox[0.95\textwidth]{\textsuperscript{1}Columbia University, \textsuperscript{2}UCLA, \textsuperscript{3}UCI, \textsuperscript{4}LDEO, \textsuperscript{5} Intel Labs}
    }
}
\renewcommand{\thefootnote}{\fnsymbol{footnote}}
\footnotetext[1]{Corresponding author: jn2808@columbia.edu}
\setcounter{page}{1}  % Reset page numbering

\section{Accountability and Reproducibility Statement}
ChaosBench is published under the open source GNU General Public License. Further development and potential updates discussed in the limitations
section will take place on the ChaosBench page. Furthermore, we are committed to maintaining and preserving the ChaosBench benchmark. Ongoing maintenance also includes tracking and resolving issues identified by the broader community after release. User feedback will be closely monitored via the GitHub issue tracker. All assets are hosted on GitHub and HuggingFace, which guarantees reliable and stable storage. 

\textbf{Dataset}: All our dataset, present and future (e.g., with more years, multi-resolution support, etc) are available at \href{https://huggingface.co/datasets/LEAP/ChaosBench}{https://huggingface.co/datasets/LEAP/ChaosBench}.

\textbf{Model Checkpoints}: All of our model checkpoints used for the purposes of ablation in this work are available at \href{https://huggingface.co/datasets/LEAP/ChaosBench/tree/main/logs}{https://huggingface.co/datasets/LEAP/ChaosBench/tree/main/logs}.

\textbf{Code}: Our code and its future extension based on community feedback is accessible at \href{https://github.com/leap-stc/ChaosBench}{https://github.com/leap-stc/ChaosBench}.

\textbf{Documentation}: Finally, our main webpage will keep track of all important updates and latest documentation, and is accessible at \href{https://leap-stc.github.io/ChaosBench}{https://leap-stc.github.io/ChaosBench}.
\newpage

\section{Getting Started}
\label{si:getting_started}
Here, we provide a detailed description on how to prepare the necessary data, perform training, and benchmark your own model. However, we refer users to our webpage \href{https://leap-stc.github.io/ChaosBench}{\texttt{https://leap-stc.github.io/ChaosBench}} for the most updated how-to guides. 

The following sections assume successful cloning of our \texttt{Github} repository \href{https://github.com/leap-stc/ChaosBench}{\texttt{https://github.com/leap-stc/ChaosBench}}. If you find any problems, feel free to contact us or raise an issue. 

\subsection{Data Preparation}

\textbf{First}, navigate to the repository directory and install the necessary dependencies.

\begin{figure}[h]
\begin{lstlisting}[language=bash]
$ cd ChaosBench 
$ pip install -r requirements.txt
\end{lstlisting}
\end{figure}

\textbf{Second}, download the dataset using the following commands.

\begin{figure}[h]
\begin{lstlisting}[language=bash]
$ cd data/
$ wget https://huggingface.co/datasets/LEAP/ChaosBench/resolve/main/process.sh
$ chmod +x process.sh
\end{lstlisting}
\end{figure}

\textbf{Third}, process the following \underline{required} and \underline{optional} dataset. 

\begin{figure}[h]
\begin{lstlisting}[language=bash]
# Required for inputs and climatology (e.g., normalization)
$ ./process.sh era5 
$ ./process.sh lra5
$ ./process.sh oras5
$ ./process.sh climatology

# Optional: control (deterministic) forecasts
$ ./process.sh ukmo
$ ./process.sh ncep
$ ./process.sh cma
$ ./process.sh ecmwf

# Optional: perturbed (ensemble) forecasts
$ ./process.sh ukmo_ensemble
$ ./process.sh ncep_ensemble
$ ./process.sh cma_ensemble
$ ./process.sh ecmwf_ensemble

# Optional: SoTa (deterministic) forecasts
$ ./process.sh panguweather
$ ./process.sh graphcast
$ ./process.sh fourcastnetv2
\end{lstlisting}
\end{figure}
\newpage

\subsection{Training}
We will cover how training can generally be performed, followed by how one can switch between different training strategies by manipulating the config \texttt{.yaml} file.

\textbf{First}, define your model class.

\begin{figure}[h]
\begin{lstlisting}[language=bash]
# An example can be found for e.g. <YOUR_MODEL> == fno

$ touch chaosbench/models/<YOUR_MODEL>.py
\end{lstlisting}
\end{figure}

\textbf{Second}, import and initialize your model in the main \texttt{chaosbench/models/model.py} file, given the pseudocode below.

\begin{figure}[h]
\begin{lstlisting}[language=python]
# Examples for lagged_ae, fno, resnet, unet are provided

import lightning.pytorch as pl
from chaosbench.models import YOUR_MODEL

class S2SBenchmarkModel(pl.LightningModule):

    def __init__(
        self, 
        ...
    ):
        super(S2SBenchmarkModel, self).__init__()
        
        # Initialize your model
        self.model = YOUR_MODEL.BEST_MODEL(...)

        # The rest of model construction logic
      
\end{lstlisting}
\end{figure}

\textbf{Third}, run the \texttt{train.py} script. We recommend using GPUs for training.

\begin{figure}[h]
\begin{lstlisting}[language=bash]
# The _s2s suffix identifies data-driven models

$ python train.py --config_filepath chaosbench/configs/<YOUR_MODEL>_s2s.yaml
\end{lstlisting}
\end{figure}

\newpage
Now you will notice that there is a \texttt{.yaml} file. We define the definition of each field, allowing for greater control over different training strategies.

\begin{figure}[h]
\begin{lstlisting}[language=bash]
# The .yaml file always has two sections: model_args and data_args

model_args:
    model_name: <str>       # Name of your model e.g., 'unet_s2s'
    input_size: <int>       # Input size, default: 60 (ERA5)
    output_size: <int>      # Output size, default: 60 (ERA5)
    learning_rate: <float>  # Learning rate
    num_workers: <int>      # Number of workers
    epochs: <int>           # Number of epochs
    t_max: <int>            # Learning rate scheduler
    only_headline: <bool>   # Only optimized for config.HEADLINE_VARS
    
data_args:
    batch_size: <int>       # Batch size
    train_years: [...]      # Train years e.g., [1979, ...] 
    val_years: [...]        # Val years e.g., [2016, ...]
    n_step: <int, 1>        # Number of autoregressive training steps
    lead_time: <int, 1>     # N-day ahead forecast (for direct scheme)
    land_vars: [...]        # Extra LRA5 vars e.g., ['t2m', ...]
    ocean_vars: [...]       # Extra ORAS5 vars e.g., ['sosstsst', ...]

\end{lstlisting}
\end{figure}

Note,
\begin{enumerate}
    \item If \texttt{only\_headline} is set to \texttt{True}, then the model is optimized only for a subset of variables defined in \texttt{config.HEADLINE\_VARS} (default: \texttt{False}).
    \item If \texttt{n\_step} is set to values greater than 1, the models will train over \( n \)-autoregressive steps (default: 1).
    \item If \texttt{lead\_time} is set to values greater than 1, the models will be able to forecast \( n \)-days ahead. For example, in our direct forecasts, if \texttt{lead\_time} is set to 4, our model will predict the states 4 days into the future (default: 1).
    \item If \texttt{land\_vars} and/or \texttt{ocean\_vars} are set with entries from the acronyms in Tables \ref{si-tab:reanalysis_lra5} and \ref{si-tab:reanalysis_oras5}, these will be used as additional inputs and targets, on top of \texttt{ERA5} variables (default: []).
\end{enumerate}
\newpage

\subsection{Evaluation}
Once training is done, we can perform evaluation depending on the use case. We recommend using GPUs for evaluation.

\textbf{First}, if we have an autoregressive model, we can simply run:

\begin{figure}[h]
\begin{lstlisting}[language=bash]
# Evaluating autoregressive model, e.g., 
# --model_name 'unet_s2s'
# --eval_years 2022
# --version_num 0       ## Checkpoint versions autogenerated in logs/
# --lra5 't2m' 'tp'     ## Additional LRA5 vars to be evaluated
# --oras5 'sosstsst'    ## Additional ORAS5 vars to be evaluated

$ python eval_iter.py --model_name <str> --eval_years <int> --version_num <int> --lra5 [...] --oras5 [...]
\end{lstlisting}
\end{figure}

\textbf{Second}, if we have a collection of models trained specifically for unique \texttt{lead\_time}, we can run:

\begin{figure}[h]
\begin{lstlisting}[language=bash]
# Evaluating direct model with the default sequence of
# lead_time = [1, 5, 10, 15, 20, 25, 30, 35, 40, 44] e.g., 
# --model_name 'unet_s2s'
# --eval_years 2022
# --version_nums 0 4 5 6 7 8 9 10 11 12 
# --lra5 't2m' 'tp'     ## Additional LRA5 vars to be evaluated
# --oras5 'sosstsst'    ## Additional ORAS5 vars to be evaluated

$ python eval_direct.py --model_name <str> --eval_years <int> --version_nums [...] --lra5 [...] --oras5 [...]
\end{lstlisting}
\end{figure}

\textbf{Third}, if we have a probabilistic model that generates ensemble forecasts (e.g., one checkpoint represents one ensemble member) and are supposed to be evaluated with additional probabilistic metrics, we can run:

\begin{figure}[h]
\begin{lstlisting}[language=bash]
# Evaluating ensembles with additional probabilistic metrics e.g., 
# --model_name 'unet_ensemble_s2s'
# --eval_years 2022
# --version_nums 0 1 2  ## One ensemble member per version
# --lra5 't2m' 'tp'     ## Additional LRA5 vars to be evaluated
# --oras5 'sosstsst'    ## Additional ORAS5 vars to be evaluated

$ python eval_ensemble.py --model_name <str> --eval_years <int> --version_nums [...] --lra5 [...] --oras5 [...]
\end{lstlisting}
\end{figure}
\newpage

\subsection{Optional: Processing Multi-Resolution Input}
\label{si-sec:multiresolution}
We open-source the data processing script to allow users to process the inputs given different resolution (highest is $0.25$-degree):

\begin{figure}[h]
\begin{lstlisting}[language=bash]
# Process inputs with e.g., 0.25-degree resolution
$ python scripts/process_atmos.py --resolution 0.25 # ERA5 
$ python scripts/process_ocean.py --resolution 0.25 # ORAS5 
$ python scripts/process_land.py  --resolution 0.25 # LRA5 
\end{lstlisting}
\end{figure}
\newpage

\section{Related Work}
\label{si:related_work}
Here we discuss the criteria used to compare different S2S benchmark. This list is by no means exhaustive and there exists many ways to interpret the different contribution, strength, and scope of each. We refer interested reader to the respective benchmark paper and website.

\noindent \textbf{On Input Variables}. The number of input channels indicates the number of \emph{unique} variables used for training data-driven models. For instance, in the case of SubseasonalClimateUSA, these include tmin, tmax, tmean, precip\_agg, precip\_mean, SST, SIC, z-10, z100, z500, z850, u-250, u-925, v-250, v-925, surface\_P, RH, SSP, precipitable water, PE, DEM, KG, MJO-phase, MJO-amp, ENSO-I, despite them having similar (25) variables across data sources.

\noindent \textbf{On Target Variables and Agencies}. Similarly, the number of target channels represent the variables these benchmarks are aiming for. This is closely related to the number of benchmark agencies, which refers to the number of physics-based simulations used as target, rather than inputs. In the case for SubseasonalClimateUSA, for instance, the number of target channels correspond to two: precipitation and surface temperature, while the number of benchmark agencies is also two: CFSv2 (NCEP) and IFS (ECMWF), despite them using multiple other simulations generated from agencies but as inputs; though evaluated on all in their follow-up work \cite{mouatadid2023adaptive} despite not initially described in the dataset paper. 

\noindent \textbf{On Physics Metrics}. The flag for physics-based metrics indicates whether these benchmarks incorporate not just physical explanation, but also formulate them as scalar and differentiable metrics for future optimization problem.

\noindent \textbf{On Probabilistic Metrics}. The flag for probabilistic metrics indicates whether these benchmarks incorporate probabilistic (e.g., CRPS, CRPSS, Spread, SSR), in addition to deterministic metrics.

\noindent \textbf{On Spatial Extent}. Furthermore, the spatial extent indicates the extent of the target benchmark, rather than of the input dataset. This is because some of the more challenging S2S forecasting task is to get the correct global space-time correlation, and having a full global coverage provides a more complete evaluation.

% \noindent \textbf{On Temporal Extent}. Lastly the temporal extent measures the greatest overlapping window of all input dataset. For instance, in the case of SubseasonalClimateUSA where they have different temporal resolution (25 input variables from 1981-2023, 23 from 1979-2023 (all except SST and SIC), and 6 from 1948-2023), we measure the most common complete years (i.e., 1981 - 2023) as these represent dense observations that can reliably constrain data-driven emulators.
\newpage

\section{ChaosBench}

\subsection{Observations from Reanalysis Products}
\subsubsection{ERA5}
\label{si:reanalysis_era5}

The following table indicates the 48 variables that are inferred by physics-based models. Note that the Input ERA5 observations contains \textbf{ALL} fields, including the unchecked boxes:

\begin{table}[h!]
\centering
\label{si-tab:reanalysis_era5}
\caption{List of ERA5 reanalysis variables}
\begin{tabular}{l|c|c|c|c|c|c|c|c|c|c}
\toprule
\textbf{Parameters/Levels (hPa)} & \textbf{1000} & \textbf{925} & \textbf{850} & \textbf{700} & \textbf{500} & \textbf{300} & \textbf{200} & \textbf{100} & \textbf{50} & \textbf{10} \\ 
\midrule
Geopotential height, z ($gpm$) & \checkmark & \checkmark & \checkmark & \checkmark & \checkmark & \checkmark & \checkmark & \checkmark & \checkmark & \checkmark \\ 
Specific humidity, q ($kg\,kg^{-1}$) & \checkmark & \checkmark & \checkmark & \checkmark & \checkmark & \checkmark & \checkmark &  &  &  \\
Temperature, t ($K$) & \checkmark & \checkmark & \checkmark & \checkmark & \checkmark & \checkmark & \checkmark & \checkmark & \checkmark & \checkmark \\
U component of wind, u ($ms^{-1}$) & \checkmark & \checkmark & \checkmark & \checkmark & \checkmark & \checkmark & \checkmark & \checkmark & \checkmark & \checkmark \\
V component of wind, v ($ms^{-1}$) & \checkmark & \checkmark & \checkmark & \checkmark & \checkmark & \checkmark & \checkmark & \checkmark & \checkmark & \checkmark \\
Vertical velocity, w ($Pas^{-1}$) &  &  &  &  & \checkmark &  &  &  &  & \\
\bottomrule
\end{tabular}
\end{table}

\subsubsection{ORAS5}
\label{si:reanalysis_oras5}
The variables for ORAS5 consist of the following as described in Table \ref{si-tab:reanalysis_oras5}.

\begin{table}[h]
    \centering
    \caption{List of ORAS5 reanalysis variables}
    \begin{tabular}{c|c|c}
        \toprule
        \bf{Acronyms} & \bf{Long Name} & \bf{Units}\\
        \midrule
        \texttt{iicethic} & sea ice thickness & m\\
        \texttt{iicevelu} & sea ice zonal velocity & $ms^{-1}$\\
        \texttt{iicevelv} & sea ice meridional velocity & $ms^{-1}$\\
        \texttt{ileadfra} & sea ice concentration & (0-1)\\
        \texttt{so14chgt} & depth of 14$^\circ$ isotherm & m\\
        \texttt{so17chgt} & depth of 17$^\circ$ isotherm & m\\
        \texttt{so20chgt} & depth of 20$^\circ$ isotherm & m\\
        \texttt{so26chgt} & depth of 26$^\circ$ isotherm & m\\
        \texttt{so28chgt} & depth of 28$^\circ$ isotherm & m\\
        \texttt{sohefldo} & net downward heat flux & $W m^{-2}$\\
        \texttt{sohtc300} & heat content at upper 300m & $J m^{-2}$\\
        \texttt{sohtc700} & heat content at upper 700m & $J m^{-2}$\\
        \texttt{sohtcbtm} & heat content for total water column & $J m^{-2}$\\
        \texttt{sometauy} & meridonial wind stress & $N m^{-2}$\\
        \texttt{somxl010} & mixed layer depth 0.01 & m\\
        \texttt{somxl030} & mixed layer depth 0.03 & m\\
        \texttt{sosaline} & salinity & PSU\\
        \texttt{sossheig} & sea surface height & m\\
        \texttt{sosstsst} & sea surface temperature & $^\circ C$\\
        \texttt{sowaflup} & net upward water flux & $kg/m^2/s$\\
        \texttt{sozotaux} & zonal wind stress & $N m^{-2}$\\
        \bottomrule
    \end{tabular}
    \label{si-tab:reanalysis_oras5}
\end{table}
\newpage

\subsubsection{LRA5}
\label{si:reanalysis_lra5}
The variables for LRA5 consist of the following as described in Table \ref{si-tab:reanalysis_lra5}.

\begin{table}[h]
    \centering
    \caption{List of LRA5 reanalysis variables}
    \begin{tabular}{c|c|c}
        \toprule
        \bf{Acronyms} & \bf{Long Name} & \bf{Units}\\
        \midrule
        \texttt{asn} & snow albedo & (0 - 1)\\
        \texttt{d2m} & 2-meter dewpoint temperature & K\\
        \texttt{e} & total evaporation & m of water equivalent\\
        \texttt{es} & snow evaporation & m of water equivalent\\
        \texttt{evabs} &evaporation from bare soil & m of water equivalent\\
        \texttt{evaow} & evaporation from open water & m of water equivalent \\
        \texttt{evatc} & evaporation from top of canopy & m of water equivalent\\
        \texttt{evavt} & evaporation from vegetation transpiration & m of water equivalent\\
        \texttt{fal} & forecaste albedo & (0 - 1)\\
        \texttt{lai\_hv} & leaf area index, high vegetation & $m^2 m^{-2}$\\
        \texttt{lai\_lv} & leaf area index, low vegetation& $m^2 m^{-2}$\\
        \texttt{pev} & potential evaporation & m\\
        \texttt{ro} & runoff & m\\
        \texttt{rsn} & snow density & $kg m^{-3}$\\
        \texttt{sd} & snow depth & m of water equivalent\\
        \texttt{sde} & snow depth water equivalent & m\\
        \texttt{sf} & snowfall & m of water equivalent\\
        \texttt{skt} & skin temperature & K\\
        \texttt{slhf} & surface latent heat flux & $J m^{-2}$\\
        \texttt{smlt} & snowmelt & m of water equivalent\\
        \texttt{snowc} & snowcover & \%\\
        \texttt{sp} & surface pressure & Pa\\
        \texttt{src} & skin reservoir content & m of water equivalent\\
        \texttt{sro} & surface runoff & m\\
        \texttt{sshf} & surface sensible heat flux & $J m^{-2}$\\
        \texttt{ssr} & net solar radiation & $J m^{-2}$\\
        \texttt{ssrd} & download solar radiation & $J m^{-2}$\\
        \texttt{ssro} & sub-surface runoff & m\\
        \texttt{stl1} & soil temperature level 1 & K\\
        \texttt{stl2} & soil temperature level 2 & K\\
        \texttt{stl3} & soil temperature level 3 & K\\
        \texttt{stl4} & soil temperature level 4 & K\\
        \texttt{str} & net thermal radiation& $J m^{-2}$\\
        \texttt{strd} & downward thermal radiation& $J m^{-2}$\\
        \texttt{swvl1} & volumetric soil water layer 1 & $m^3 m^{-3}$\\
        \texttt{swvl2} & volumetric soil water layer 2& $m^3 m^{-3}$\\
        \texttt{swvl3} & volumetric soil water layer 3& $m^3 m^{-3}$\\
        \texttt{swvl4} & volumetric soil water layer 4& $m^3 m^{-3}$\\
        \texttt{t2m} & 2-meter temperature & K\\
        \texttt{tp} & total precipitation & m\\
        \texttt{tsn} & temperature of snow layer & K\\
        \texttt{u10} & 10-meter u-wind & $ms^{-1}$\\
        \texttt{v10} & 10-meter v-wind & $ms^{-1}$\\
        \bottomrule
    \end{tabular}
    \label{si-tab:reanalysis_lra5}
\end{table}
\newpage

\subsection{Physics-Based Simulations}
\label{si:physics_model}
In this section, we describe in detail the physics-based models used as baselines in ChaosBench. Wherever possible, we discuss specific strategies regarding coupling to the ocean, sea ice, wave, land, initialization and perturbation strategies, specifications of initial/boundary conditions, as well as other numerical considerations to generate forecast. 

\subsubsection{The UK Meteorological Office (UKMO) \cite{williams2015met}}

\begin{itemize}
    \item \textbf{Initialization and Ensemble}. The UKMO model employs the lagged initialization strategy to generate an ensemble of forecasts (4 in this case) at different initialization time to improve prediction stability.\\
    
    \item \textbf{Coupling with ocean} is performed with the Global Ocean 6.0 model \cite{storkey2018uk}, based on NEMO3.6 \cite{mathiot2017explicit} with 0.25 degree horizontal resolution and 75 vertical pressure levels. The ocean model is initialized and calibrated using Nonlinear Evolutionary Model VARiation (NEMOVAR) \cite{mogensen2012nemovar}, a specific data assimilation strategy that uses temperature, salinity profiles, altimeter-derived sea level anomalies to calibrate forecasts. Frequency of coupling is 1-hourly.\\

    \item \textbf{Coupling with sea ice} is performed with the Global Sea Ice 8.1 (CICE5.1.2) model \cite{tsujino2020evaluation}, and again initialized from NEMOVAR.\\

    \item \textbf{Coupling with wave model} is not yet operational.\\

    \item \textbf{Coupling with land surface} is performed with the Joint UK Land Environment Simulator (JULES) \cite{best2011joint}. Soil moisture, soil temperature, and snow are initialized using JULES and forced using the the Japanese 55-year Reanalysis (JRA-55) data \cite{kobayashi2015jra}. The land surface model is paramaterized by land cover type from a combination of satellite (e.g., MODIS LAI \cite{tian2002multiscale}) and  radiometer data (e.g., AVHRR \cite{loveland2000development}). In addition, another parameterization in the form of soil characteristics is derived from the Harmonized World Soil Database \cite{wieder2014regridded}.\\

    \item \textbf{Model grid} uses the Arakawa C-grid \cite{arakawa1997computational} to solve partial differential equations on a spherical surface. In particular, the velocity components (such as zonal and meridional wind) are defined at the center of each face of the grid cells (in the case of a rectilinear grid) or along cell edges (in the case of a curvilinear grid). The scalar quantities such as pressure or temperature are computed at the corners of the grid cells.\\

    \item \textbf{Large-scale dynamics} uses the Semi-Lagrangian approach. It does not strictly follow fluid parcels (i.e., Lagrangian), but it does calculate the value of a field, such as temperature (i.e., Eulerian) by tracing back along the trajectory that a fluid parcel would have taken to reach a specific point at the current time step. This backward trajectory is used to find the origin of the fluid parcel and determine its properties, which are then used to update the model fields. This hybrid approach is therefore termed Semi-Lagrangian. \\
\end{itemize}

\subsubsection{National Centers for Environmental Prediction (NCEP) \cite{saha2010ncep}}
\begin{itemize}
    \item \textbf{Initialization and Ensemble}. The NCEP model adds small perturbation to the atmospheric, oceanic and land analysis at each cycle across 4 ensemble to reduce sensitivity to initial conditions.\\
    
    \item \textbf{Coupling with ocean} is performed with the GFDL Modular Ocean Model version 4 (MOM4) model that has a spatial resolution of 0.5-degree and 0.25-degree in the longitude-latitude directions \cite{griffies2004technical}. There are 40 vertical pressure levels.\\

    \item \textbf{Coupling with sea ice} is also performed with the GFDL Sea Ice Simulator (SIS), which models the thermodynamics and overall dynamics of sea ice \cite{griffies2004technical}. \\
    
    \item \textbf{Coupling with wave model} is not yet operational.\\

    \item \textbf{Coupling with land surface} is performed with 4-layer Noah Land surface model 2.7.1 \cite{ek2003implementation}. Soil moisture, soil temperature, and snow are initialized using Noah and forced using the Climate Forecast System \cite{saha2010ncep} and the Global Land Data Assimilation System \cite{meng2012land} reanalysis data. The land surface model is parameterized by land cover type AVHRR data. In addition, another paramaterization in the form of soil characteristics is derived from the world soil climate database \cite{zobler1986world}.\\
    
    \item \textbf{Model grid} uses the Gaussian grid \cite{hortal1991use}, where the longitude (x-axis) are evenly spaced while the latitudes (y-axis) are not. Instead, they are determined by the roots of the associated Legendre polynomials, which correspond to the Gaussian quadrature points for the sphere. This ensures that the actual area represented by each grid cell is more uniform.\\
    
    \item \textbf{Large-scale dynamics} uses the Spectral approach. It solves partial differential equations by transforming them from the physical space into the spectral domain. In the latter case, the equations are transformed into a series of coefficients that represent the amplitude of waves across scales. The transformations are usually done using Fourier series for periodic domains or spherical harmonics when dealing with the whole Earth's surface \cite{hortal1991use}. This method is especially beneficial for smooth functions and for representing large-scale wave phenomena, such as the Rossby waves, which are important for understanding  weather and climate. 
\end{itemize}

\subsubsection{China Meteorological Administration (CMA) \cite{wu2019beijing}}
\begin{itemize}
    \item \textbf{Initialization and Ensemble}. The CMA model uses the lagged average forecasting (LAF) method across 4 ensemble members to ensure that the mean forecast is less sensitive to initial conditions.\\
    
    \item \textbf{Coupling with ocean} is performed with the GFDL MOM4 model, which has 40 vertical pressure levels \cite{griffies2004technical}. Frequency of coupling is 2-hourly.\\
    
    \item \textbf{Coupling with sea ice} is performed with the GFDL Sea Ice Simulator (SIS), similar to that used by NCEP \cite{griffies2004technical}.\\
    
    \item \textbf{Coupling with wave model} is not yet operational.\\

    \item \textbf{Coupling with land surface} is performed with the Atmosphere-Vegetation Interaction Model version 2 (AVIM2) model \cite{wu2014overview} and the NCAR NCAR Community Land Model version 3.0 (CLMv3) \cite{lawrence2007representing}. Soil moisture, soil temperature, and snow are not initialized directly using reanalysis data, as used by other land surface models. Rather, air-sea-land-ice coupled model is forced by near-surface atmospheric and ocean reanalysis in a long-term integration, and the land initial conditions are produced as a by-product. As a result, the parameterization of land cover type is done by this process, while soil characteristics is derived from the Harmonized World Soil Database \cite{wieder2014regridded}.\\
    
    \item \textbf{Model grid} uses the Gaussian grid \cite{hortal1991use}, similar to that used by the NCEP.\\
    
    \item \textbf{Large-scale dynamics} uses a mixture of Spectral approach for the vorticity, temperature, and surface pressure, as well as Semi-Lagrangian for specific humidity and cloud waters other tracers.
\end{itemize}

\subsubsection{European Center for Medium-Range Weather Forecasts (ECMWF) \cite{documentationpart}}
\begin{itemize}
    \item \textbf{Initialization and Ensemble}. The operational IFS forecast is generated through Singular Vectors (SV) method: it creates a variety of initial conditions by adjusting certain parameters slightly, thus generating different starting points. \\
    
    \item \textbf{Coupling with ocean} is performed with NEMO3.4.1 with 1-degree resolution and 42 vertical pressure levels. Frequency of coupling is 3-hourly.\\

    \item \textbf{Coupling with sea ice} is not operational for this model's version (but it is in the newer generation, though the forecast start-date is much later than 2016). As a result, sea ice initial conditions are persisted up to day 15 and then relaxed to climatology up to day 45.\\
    
    \item \textbf{Coupling with wave model} is performed with ECMWF wave model with 0.5-degree resolution \cite{janssen2005progress}.\\

    \item \textbf{Coupling with land surface} is relatively more complex than the rest, and we refer readers to their \href{https://www.ecmwf.int/en/forecasts/documentation-and-support/changes-ecmwf-model/cy41r1-summary-changes}{documentation}. Regardless, it is based on Land Data Assimilation System (LDAS) that combines heterogenous high-quality dataset from satellite to ground sensors, and integrated with the operational IFS model. The parameterization for land cover type is primarily based on MODIS collection 5 \cite{tian2002multiscale} and soil characteristics from the FAO dominant soil texture class \cite{van1980closed}.\\
    
    \item \textbf{Model grid} uses the Cubic Octohedral grid \cite{malardel2016new}, where the Earth's surface is projected onto a cube. Then, the cube is further subdivided to form an octahedron, where the faces represent finer grid cells. This multi-scale gridding scheme allows for parallelization where processes at different scales could be solved simultaneously.\\
    
    \item \textbf{Large-scale dynamics} uses a mixture of Spectral and Semi-Lagrangian approach, similar to that used by CMA.
\end{itemize}
\newpage

\section{Data-Driven Baseline Models}
\label{si:data_model}
In this section, we describe in detail implementation and hyperparemeter selections of our data-driven  models used as baselines to ChaosBench. Most of the choices are based on the original works that are adapted to weather and climate applications using similar input dataset. All training are performed using 2x NVIDIA A100 GPUs.

\subsection{Lagged Autoencoder (AE)}
We implement lagged AE from \cite{chen2022automated} with 5 $encoder$ blocks and 5 $decoder$ block, with detailed specification in Table \ref{si-tab:lagged_ae}. Each encoder block is comprised of $\textsc{MaxPool2d} \circ (\textsc{Conv2d} \rightarrow \textsc{BatchNorm2d} \rightarrow \textsc{ReLU} \rightarrow \textsc{Conv2d} \rightarrow \textsc{BatchNorm2d} \rightarrow \textsc{ReLU})$. Similarly, the decoder block is comprised of $\textsc{ConvTranspose2d} \rightarrow \textsc{BactNorm2d} \rightarrow \textsc{ReLU}) \bigoplus (\textsc{ConvTranspose2d} \rightarrow \textsc{BactNorm2d} \rightarrow \textsc{Sigmoid}) \circ (\textsc{Conv2d})$.

\begin{table}[h]
    \centering
    \caption{Hyperparameters for Lagged AE}
    \begin{tabular}{c|c}
         \toprule
         \textbf{Hyperparameters} & \textbf{Values} \\
         \midrule
         Channels & [64, 128, 256, 512, 1024]\\
         Encoder Kernel & $3 \times 3$ \\
         Decoder Kernel & $2 \times 2$\\
         Max Pooling Window & $2 \times 2$\\
         Batch Normalization & \textsc{True}\\
         Optimizer & \textsc{AdamW} \cite{loshchilov2017decoupled}\\
         Learning Rate & $\textsc{CosineAnnealing}(10^{-2} \rightarrow 10^{-3})$\\
         Batch Size & 32 \\
         Epochs & 500\\
         Tmax & 500\\
         \bottomrule
    \end{tabular}
    \label{si-tab:lagged_ae}
\end{table}

\subsection{ResNet}
We adapt ResNet implementation from \cite{rasp2021data} using ResNet-50 as feature extractor and 5 $decoder$ blocks, following specification in Table \ref{si-tab:resnet}. Each decoder block is composed of $\textsc{ConvTranspose2d} \rightarrow \textsc{BactNorm2d} \rightarrow \textsc{LeakyReLU}$.

\begin{table}[h]
    \centering
    \caption{Hyperparameters for ResNet}
    \begin{tabular}{c|c}
         \toprule
         \textbf{Hyperparameters} & \textbf{Values} \\
         \midrule
         Backbone & \textsc{ResNet-50}\\
         Decoder Channels & [1024, 512, 256, 128, 64]\\
         Decoder Activation & \textsc{LeakyReLU(0.15)} \\
         Optimizer & \textsc{AdamW}\\
         Learning Rate & $\textsc{CosineAnnealing}(10^{-2} \rightarrow 10^{-3})$\\
         Batch Size & 32 \\
         Epochs & 500\\
         Tmax & 500\\
         \bottomrule
    \end{tabular}
    \label{si-tab:resnet}
\end{table}

\subsection{UNet}
We adapt UNet implementation from \cite{rasp2020weatherbench} using 5 $encoder$ and 5 $decoder$ blocks, with skip connections, following specification in Table \ref{si-tab:unet}. The composition of the encoder and decoder components are similar to those described for Lagged Autoencoder, with the addition of \textsc{skip} connection between each corresponding contracting-expansive path.

\begin{table}[h]
    \centering
    \caption{Hyperparameters for UNet}
    \begin{tabular}{c|c}
         \toprule
         \textbf{Hyperparameters} & \textbf{Values} \\
         \midrule
         Channels & [64, 128, 256, 512, 1024]\\
         Activation & \textsc{LeakyReLU(0.15)}\\
         Encoder Kernel & $3 \times 3$ \\
         Decoder Kernel & $2 \times 2$\\
         Max Pooling Window & $2 \times 2$\\
         Optimizer & \textsc{AdamW}\\
         Learning Rate & $\textsc{CosineAnnealing}(10^{-2} \rightarrow 10^{-3})$\\
         Batch Size & 32 \\
         Epochs & 500\\
         Tmax & 500\\
         \bottomrule
    \end{tabular}
    \label{si-tab:unet}
\end{table}

\subsection{Fourier Neural Operator (FNO)}
\begin{figure*}[h]
    \centering
    \includegraphics[width=0.8\textwidth]{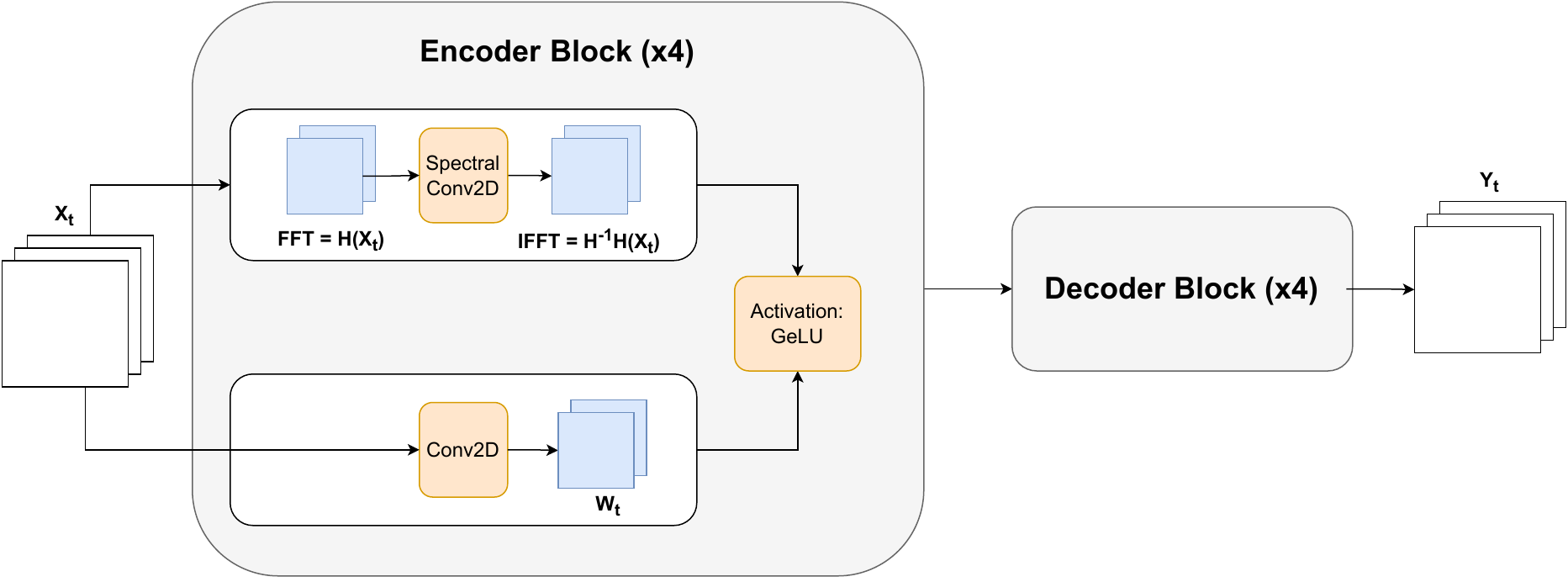}
    \caption{FNO architecture: (1) in the encoder block, we transform our input $X_t$ by convolutional layers both in the Fourier and physical domains, before we concatenate and apply non-linear \textsc{GeLU} activation function. The (2) decoder block is then applying deconvolutional operation to the latent features to generate forecast $Y_t$.}
    \label{si-fig:fno}
\end{figure*}

We adapt FNO implementation from \cite{li2020fourier}, following specification in Table \ref{si-tab:fno} and illustrated in \ref{si-fig:fno}. We implement the encoder-decoder structure, where we (1) first transform our input $\mathbf{X}_t$ by convolutional layers both in the Fourier (applying fast fourier transform; FFT) and physical domains, before we concatenate both (applying inverse FFT for the former convolved features), and apply non-linear \textsc{GeLU} activation function \cite{hendrycks2016gaussian}. We select only the first 4 main Fourier modes to make the number of trainable parameters comparable with the other data-driven baseline models. The (2) decoder block then applies deconvolutional operation to the latent features to generate output $Y_t$.

\begin{table}[h]
    \centering
    \caption{Hyperparameters for FNO}
    \begin{tabular}{c|c}
         \toprule
         \textbf{Hyperparameters} & \textbf{Values} \\
         \midrule
         Non-Spectral Channels & [64, 128, 256, 512, 1024]\\
         Spectral Channel & [64, 128, 256, 512, 1024]\\
         Activation & \textsc{GeLU}\\
         Fourier Modes  & (4,4) \\
         Optimizer & \textsc{AdamW}\\
         Learning Rate & $\textsc{CosineAnnealing}(10^{-2} \rightarrow 10^{-3})$\\
         Batch Size & 32 \\
         Epochs & 500\\
         Tmax & 500\\
         \bottomrule
    \end{tabular}
    \label{si-tab:fno}
\end{table}

\subsection{ClimaX}
ClimaX is based on the ViT model \cite{dosovitskiy2020image} with variational positional embedding in variable-time space. We use ClimaX model as is described and implemented in the original paper and is pre-trained using CMIP6 \cite{nguyen2023climax}. We fine-tune the original pre-trained model given our training setup.

\subsection{PanguWeather, FourCastNetV2, GraphCast}
We perform inference using their latest checkpoints using the API provided here: \href{https://github.com/ecmwf-lab/ai-models}{https://github.com/ecmwf-lab/ai-models}.

For this work, we process the forecasts at biweekly temporal resolution. In the codebase, we provide the script for further flexibility, for instance:

\begin{figure}[h]
\begin{lstlisting}[language=bash]
# Process biweekly, 1.5-degree forecasts for the year 2022

## Panguweather
$ python scripts/process_sota.py --model_name panguweather --years 2022 

## Graphcast
$ python scripts/process_sota.py --model_name graphcast --years 2022 

## FourCastNetV2
$ python scripts/process_sota.py --model_name fourcastnetv2 --years 2022 
\end{lstlisting}
\end{figure}
\newpage

\section{Evaluation Metrics}
\subsection{Deterministic Metrics}
\label{si:vision_metrics}
We describe in detail the four primary vision-based metrics used for this benchmark, including RMSE, Bias, ACC, and MS-SSIM.

\subsubsection{Root Mean-Squared Error (RMSE)}
As described in the main text, we apply latitude-adjustment to RMSE computation.

\begin{equation}
    \label{eq:rmse}
    \mathcal{M}_{RMSE} = \sqrt{\frac{1}{|\boldsymbol{\theta}||\boldsymbol{\gamma}|} \sum_{i=1}^{|\boldsymbol{\theta}|}\sum_{j=1}^{|\boldsymbol{\gamma}|} w(\theta_i) (\hat{\mathbf{Y}}_{i,j} - \mathbf{Y}_{i,j})^2}
\end{equation}

\subsubsection{Bias}
Similarly, we apply latitude-adjustment to Bias computation. 

\begin{equation}
    \label{eq:bias}
    \mathcal{M}_{Bias} = \frac{1}{|\boldsymbol{\theta}||\boldsymbol{\gamma}|} \sum_{i=1}^{|\boldsymbol{\theta}|}\sum_{j=1}^{|\boldsymbol{\gamma}|} w(\theta_i) (\hat{\mathbf{Y}}_{i,j} - \mathbf{Y}_{i,j})
\end{equation}

\subsubsection{Anomaly Correlation Coefficient (ACC)}
\label{si:acc}
We remove the indexing for a more compact representation where the summation is performed over each grid cell $(i,j)$. The predicted and observed anomalies at each grid-cell are denoted by $A_{\hat{\mathbf{Y}}_{i,j}} = \hat{\mathbf{Y}}_{i,j} - C$ and $A_{\mathbf{Y}_{i,j}} = \mathbf{Y}_{i,j} - C$, where $C$ is the observational climatology. We apply latitude-adjustment to ACC computation.

\begin{equation}
    \label{eq:acc}
    \mathcal{M}_{ACC} = \frac{\sum w(\theta)[A_{\hat{\mathbf{Y}}} \cdot A_{\mathbf{Y}}]}{\sqrt{\sum w(\theta)A^2_{\hat{\mathbf{Y}}}\sum w(\theta)A^2_{\mathbf{Y}}}}
\end{equation}

\subsubsection{Multi-scale Structural Similarity Index Measure (MS-SSIM)}
\label{si:ms-ssim}
Let $\mathbf{Y}$ and $\hat{\mathbf{Y}}$ be two images to be compared, and let $\mu_\mathbf{Y}$, $\sigma^2_\mathbf{Y}$ and $\sigma_{\mathbf{Y}\hat{\mathbf{Y}}}$ be the mean of $\mathbf{Y}$, the variance of $\mathbf{Y}$, and the covariance of $\mathbf{Y}$ and $\hat{\mathbf{Y}}$, respectively. The luminance, contrast and structure comparison measures are defined as follows:
\begin{equation}
    l(\mathbf{Y}, \hat{\mathbf{Y}})=\frac{2 \mu_\mathbf{Y} \mu_{\hat{\mathbf{Y}}}+C_1}{\mu_\mathbf{Y}^2+\mu_{\hat{\mathbf{Y}}}^2+C_1},
\end{equation}
\begin{equation}
    c(\mathbf{Y}, \hat{\mathbf{Y}})=\frac{2 \sigma_\mathbf{Y} \sigma_{\hat{\mathbf{Y}}}+C_2}{\sigma_\mathbf{Y}^2+\sigma_{\hat{\mathbf{Y}}}^2+C_2},
\end{equation}
\begin{equation}
    s(\mathbf{Y}, \hat{\mathbf{Y}})=\frac{\sigma_{ \mathbf{Y}\hat{\mathbf{Y}}}+C_3}{\sigma_\mathbf{Y} \sigma_{\hat{\mathbf{Y}}}+C_3},
\end{equation}
where $C_1$, $C_2$ and $C_3$ are constants given by
\begin{equation}
C_1 = (K_1L)^2, C_2 = (K_2L)^2, \text{ and } C_3 = C_2/2.
\end{equation}
$L=255$ is the dynamic range of the gray scale images, and $K_1\ll 1$ and $K_2 \ll 1$ are two small constants. To compute the MS-SSIM metric across multiple scales, the images are successively low-pass filtered and down-sampled by a factor of 2. We index the original image as scale 1, and the desired highest scale as scale $M$. At each scale, the contrast comparison and structure comparison are computed and denoted as $c_j(\mathbf{Y},\hat{\mathbf{Y}})$ and $s_j(\mathbf{Y},\hat{\mathbf{Y}})$ respectively. The luminance comparison is only calculated at the last scale $M$, denoted by $l_M(\mathbf{Y},\hat{\mathbf{Y}})$. Then, the MS-SSIM metric is defined by
\begin{equation}
\label{eq:ms-ssim}
\mathcal{M}_{MS-SSIM} =  [l_M(\mathbf{Y}, \hat{\mathbf{Y}})]^{\alpha_M}\cdot\prod_{j=1}^M[c_j(\mathbf{Y}, \hat{\mathbf{Y}})]^{\beta_j}[s_j(\mathbf{Y}, \hat{\mathbf{Y}})]^{\gamma_j}
\end{equation}
where $\alpha_M$, $\beta_j$ and $\gamma_j$ are parameters. We use the same set of parameters as in \cite{wang2003multiscale}: $K_1 = 0.01$, $K_2 = 0.03$, $M=5$, $\alpha_5=\beta_5=\gamma_5=0.1333$, $\beta_4=\gamma_4=0.2363$, $\beta_3=\gamma_3=0.3001$, $\beta_2=\gamma_2=0.2856$, $\beta_1=\gamma=0.0448$. The predicted and ground truth images of physical variables are re-scaled to 0-255 prior to the calculation of their MS-SSIM values.

\subsection{Physics-Based Metrics}
\label{si:physics_metrics}
In this section, we describe in detail the definition and implementation of our physics-based metrics, including \textsc{PyTorch} psuedocode implementation.

Let $\mathbf{Y}$ be a 2D image of size $h \times w$ for a physical variables at a specific time, variable, and level. Let $f(x,y)$ be the intensity of the pixel at position $(x,y).$  First, we compute the 2D Fourier transform of the image by
\begin{equation}
    F(k_x, k_y) = \sum_{x=0}^{w-1}\sum_{y=0}^{h-1}f(x,y)\cdot e^{-2\pi i \left(k_xx/w+ k_yy/h\right)},
    \label{eq:DFT}
\end{equation}
where $k_x$ and $k_y$ correspond to the wavenumber components in the horizontal and vertical directions, respectively, and $i$ is the imaginary unit. The power at each wavenumber component $(k_x, k_y)$ is given by the square of the magnitude spectrum of $F(k_x, k_y)$, that is,
\begin{equation}
    S(k_x, k_y) = \vert F(k_x, k_y) \vert ^2 = \texttt{Re}[F(k_x, k_y)]^2 + \texttt{Im}[F(k_x, k_y)]^2.
\end{equation}
\\
The scalar wavenumber is defined as:
\begin{equation}
    k=\sqrt{k_x^2 + k_y^2},
    \label{eq:k}
\end{equation}
which represents the magnitude of the spatial frequency vector, indicating how rapidly features change spatially regardless of direction. Then, the energy distribution at a spatial frequency corresponding to k is defined as
\begin{equation}
    S(k)=\sum_{(k_x,k_y):\sqrt{k_x^2+k_y^2}=k}S(k_x, k_y).
    \label{eq:Ek}
\end{equation}
Given the spatial energy frequency distribution for observations $E(k)$ and predictions $\hat{S}(k)$ , we perform normalization for each over $\mathbf{K}_q$, the set of wavenumbers corresponding to high-frequency components of energy distribution, as defined in Equation \ref{eq:norm_ek}. This is to ensure that the sum of the component sums up to 1 which exhibits pdf-like property. 

\begin{equation}
    S^\prime(k) = \frac{S(k)}{\sum_{k\in\mathbf{K}_q}S(k)}, \quad
    \hat{S}^\prime(k) = \frac{\hat{S}(k)}{\sum_{k\in\mathbf{K}_q}\hat{S}(k)},\quad k\in\mathbf{K}_q
    \label{eq:norm_ek}
\end{equation}
\newpage

\begin{figure}[t!]
\begin{lstlisting}[language=Python, label={lst:specdiv}, caption=Psuedocode for computing SpecDiv using \textsc{PyTorch}]
import torch
import torch.nn as nn
    
class SpectralDiv(nn.Module):
    """
    Compute Spectral divergence given the top-k percentile wavenumber (higher k means higher frequency)
    """
    def __init__(
        self,   
        percentile=0.9,
        input_shape=(121,240)
    ):
        super(SpectralDiv, self).__init__()
        
        self.percentile = percentile
        
        # Compute the discrete Fourier Transform sample frequencies for a signal of size
        nx, ny = input_shape
        kx = torch.fft.fftfreq(nx) * nx
        ky = torch.fft.fftfreq(ny) * ny
        kx, ky = torch.meshgrid(kx, ky)

        # Construct discretized k-bins
        self.k = specify_k_bins(...)
        
        # Get k-percentile index
        self.k_percentile_idx = int(len(self.k) * self.percentile)
        
    def forward(self, predictions, targets):
        
        # Preprocess data, including handling of missing values, etc
        predictions = preprocess_data(...)
        targets = preprocess_data(...)
        
        # Compute along mini-batch
        predictions, targets = torch.nanmean(predictions, dim=0), torch.nanmean(targets, dim=0)
        
        # Transform prediction and targets onto the Fourier space and compute the power
        predictions_power = torch.fft.fft2(predictions)
        predictions_power = torch.abs(predictions_power)**2
        
        targets_power = torch.fft.fft2(targets)
        targets_power = torch.abs(targets_power)**2
        
        # Normalize as pdf
        predictions_Sk = predictions_power / torch.nansum(predictions_power)
        targets_Sk = targets_power / torch.nansum(targets_power)

        # Compute spectral Sk divergence
        div = torch.nansum(targets_Sk * torch.log(torch.clamp(targets_Sk / predictions_Sk, min=1e-9)))
        
        return div
\end{lstlisting}
\end{figure}
\clearpage

\begin{figure}[t!]
\begin{lstlisting}[language=Python, label={lst:specres}, caption=Psuedocode for computing SpecRes using \textsc{PyTorch}]
import torch
import torch.nn as nn
    
class SpectralRes(nn.Module):
    """
    Compute Spectral residual given the top-k percentile wavenumber (higher k means higher frequency)
    """
    def __init__(
        self,   
        percentile=0.9,
        input_shape=(121,240)
    ):
        super(SpectralRes, self).__init__()
        
        self.percentile = percentile
        
        # Compute the discrete Fourier Transform sample frequencies for a signal of size
        nx, ny = input_shape
        kx = torch.fft.fftfreq(nx) * nx
        ky = torch.fft.fftfreq(ny) * ny
        kx, ky = torch.meshgrid(kx, ky)

        # Construct discretized k-bins
        self.k = specify_k_bins(...)
        
        # Get k-percentile index
        self.k_percentile_idx = int(len(self.k) * self.percentile)
        
    def forward(self, predictions, targets):
        
        # Preprocess data, including handling of missing values, etc
        predictions = preprocess_data(...)
        targets = preprocess_data(...)
        
        # Compute along mini-batch
        predictions, targets = torch.nanmean(predictions, dim=0), torch.nanmean(targets, dim=0)
        
        # Transform prediction and targets onto the Fourier space and compute the power
        predictions_power = torch.fft.fft2(predictions)
        predictions_power = torch.abs(predictions_power)**2
        
        targets_power = torch.fft.fft2(targets)
        targets_power = torch.abs(targets_power)**2
        
        # Normalize as pdf
        predictions_Sk = predictions_power / torch.nansum(predictions_power)
        targets_Sk = targets_power / torch.nansum(targets_power)

        # Compute spectral Sk residual
        res = torch.sqrt(torch.nanmean(torch.square(predictions_Sk - targets_Sk)))
        
        return res
\end{lstlisting}
\end{figure}
\clearpage

\subsection{Probabilistic Metrics}
\label{si:probabilistic_metrics}
Here, we broadly define $n \in N$ as an ensemble member, and $N \in \mathbb{R}$ the total number of ensemble members.

\subsubsection{Deterministic Extension}
This includes the ensemble version of deterministic and physics-based metrics, including RMSE, Bias, ACC, MS-SSIM, SpecDiv, and SpecRes.

\begin{equation}
    \label{eq:rmse_ens}
    \mathcal{M}_{RMSE}^{ens} = \frac{1}{N}\sum_{n=1}^N\mathcal{M}_{RMSE}^n
\end{equation}

\begin{equation}
    \label{eq:bias_ens}
    \mathcal{M}_{Bias}^{ens} = \frac{1}{N}\sum_{n=1}^N\mathcal{M}_{Bias}^n
\end{equation}

\begin{equation}
    \label{eq:acc_ens}
    \mathcal{M}_{ACC}^{ens} = \frac{1}{N}\sum_{n=1}^N\mathcal{M}_{ACC}^n
\end{equation}

\begin{equation}
    \label{eq:ssim_ens}
    \mathcal{M}_{MS-SSIM}^{ens} = \frac{1}{N}\sum_{n=1}^N\mathcal{M}_{MS-SSIM}^n
\end{equation}

\begin{equation}
    \label{eq:specdiv_ens}
    \mathcal{M}_{SpecDiv}^{ens} = \frac{1}{N}\sum_{n=1}^N\mathcal{M}_{SpecDiv}^n
\end{equation}

\begin{equation}
    \label{eq:specres_ens}
    \mathcal{M}_{SpecRes}^{ens} = \frac{1}{N}\sum_{n=1}^N\mathcal{M}_{SpecRes}^n
\end{equation}

\subsubsection{CRPS}
CRPS measures the accuracy of probabilistic forecasts by integrating the square of the difference between the cumulative distribution function (CDF) of the forecast and the CDF of the observed data over all possible outcomes. It can be thought of as probabilistic MAE, where a smaller value is desirable and a deterministic forecast reduces to MAE. We first apply latitude-adjustments for the forecasts and target fields.

\begin{equation}
    \label{eq:crps}
    \mathcal{M}_{CRPS}(F, x) = \int_{-\infty}^{\infty} (F(y) - H(y - x))^2 \, dy
\end{equation}

where \( F(y) \) is the CDF of the forecast, \( H(y - x) \) is the Heaviside step function at the observed value \( x \), and \( y \) ranges over all possible outcomes. 

\subsubsection{CRPSS}
CRPSS measures the skillfulness of an ensemble forecasts, with positive being skillful, zero unskilled, and negative being worse than baseline climatology.

\begin{equation}
    \label{eq:crpss}
    \mathcal{M}_{CRPSS} = 1 - \frac{CRPS_{\text{forecast}}}{CRPS_{\text{climatology}}}
\end{equation}

\subsubsection{Spread}
We apply latitude-adjusted spread of the ensemble members, and \texttt{std} is the standard deviation operator. 

\begin{equation}
    \label{eq:spread}
    \mathcal{M}_{Spread} = \frac{1}{|\boldsymbol{\theta}||\boldsymbol{\gamma}|} \sum_{i=1}^{|\boldsymbol{\theta}|} \sum_{j=1}^{|\boldsymbol{\gamma}|} \texttt{std} \left( \{ w(\theta_i) \hat{\mathbf{Y}}^n_{i,j} \}_{n=1}^N \right)
\end{equation}

\subsubsection{Spread/Skill Ratio (SSR)}
We use ensemble RMSE as the skill in the SSR computation.

\begin{equation}
    \label{eq:ssr}
    \mathcal{M}_{SSR} = \frac{\mathcal{M}_{Spread}}{\mathcal{M}_{RMSE}^{ens}}
\end{equation}
\clearpage

\section{Extended Results}
We provide extended results accompanying the main text.

\begin{figure*}[h]
    \centering
    \begin{subfigure}{0.7\textwidth}
        \includegraphics[width=\textwidth]{imgs/center_rmse.pdf}
        \caption{RMSE ($\downarrow$ is better)}
    \end{subfigure}
    \hfill
    \begin{subfigure}{0.7\textwidth}
        \includegraphics[width=\textwidth]{imgs/center_acc.pdf}
        \caption{ACC ($\uparrow$ is better)}
    \end{subfigure}
    \hfill
    \begin{subfigure}{0.7\textwidth}
        \includegraphics[width=\textwidth]{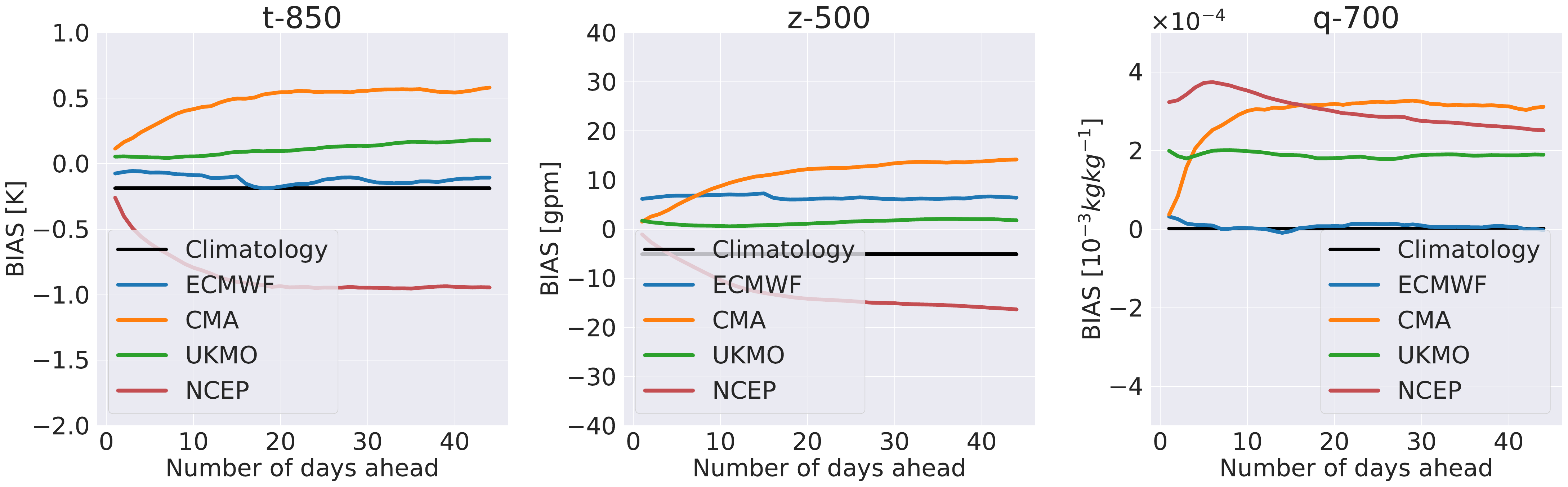}
        \caption{Bias ($\rightarrow 0$ is better)}
    \end{subfigure}
    \hfill
    \begin{subfigure}{0.7\textwidth}
        \includegraphics[width=\textwidth]{imgs/center_ssim.pdf}
        \caption{MS-SSIM ($\uparrow$ is better)}
    \end{subfigure}
    \hfill
    \begin{subfigure}{0.7\textwidth}
        \includegraphics[width=\textwidth]{imgs/center_sdiv.pdf}
        \caption{SpecDiv ($\downarrow$ is better)}
    \end{subfigure}
    
\caption{Evaluation results between baseline climatology (black line) and physics-based control (deterministic) forecasts. At longer forecasting horizon, most physics-based deterministic forecasts perform worse than climatology while maintaining structures as evidenced from their low SpecDiv (barring NCEP).}
\label{si-fig:all_results_centers}
\end{figure*}

\begin{figure*}[h]
    \centering
    \begin{subfigure}{0.8\textwidth}
        \includegraphics[width=\textwidth]{imgs/sota_rmse.pdf}
        \caption{RMSE ($\downarrow$ is better)}
    \end{subfigure}
    \hfill
    \begin{subfigure}{0.8\textwidth}
        \includegraphics[width=\textwidth]{imgs/sota_acc.pdf}
        \caption{ACC ($\uparrow$ is better)}
    \end{subfigure}
    \hfill
    \begin{subfigure}{0.8\textwidth}
        \includegraphics[width=\textwidth]{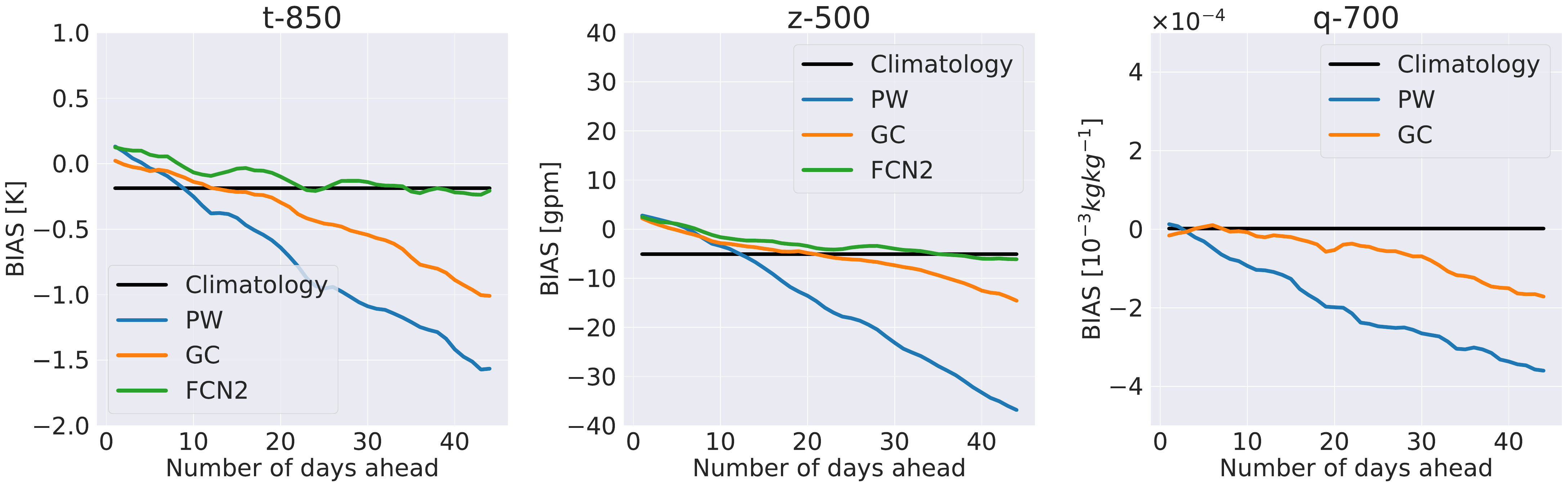}
        \caption{Bias ($\rightarrow 0$ is better)}
    \end{subfigure}
    \hfill
    \begin{subfigure}{0.8\textwidth}
        \includegraphics[width=\textwidth]{imgs/sota_ssim.pdf}
        \caption{MS-SSIM ($\uparrow$ is better)}
    \end{subfigure}
    \hfill
    \begin{subfigure}{0.8\textwidth}
        \includegraphics[width=\textwidth]{imgs/sota_sdiv.pdf}
        \caption{SpecDiv ($\downarrow$ is better)}
    \end{subfigure}
    
\caption{Evaluation results between baseline climatology (black line) and data-driven models including PanguWeather (PW), FourCastNetV2 (FCN2), and GraphCast (GC). Overall, we observe that data-driven models perform significantly worse than climatology on S2S timescale. They also perform poorly on physics-based metrics indicating the lack of predictive power on multi-scale structures. \underline{Note}: FCN2 lacks q-700 and climatology naturally has low SpecDiv (direct observations).}
\label{si-fig:all_results_sota}
\end{figure*}

\begin{figure*}[h]
    \centering
    \begin{subfigure}{0.8\textwidth}
        \includegraphics[width=\textwidth]{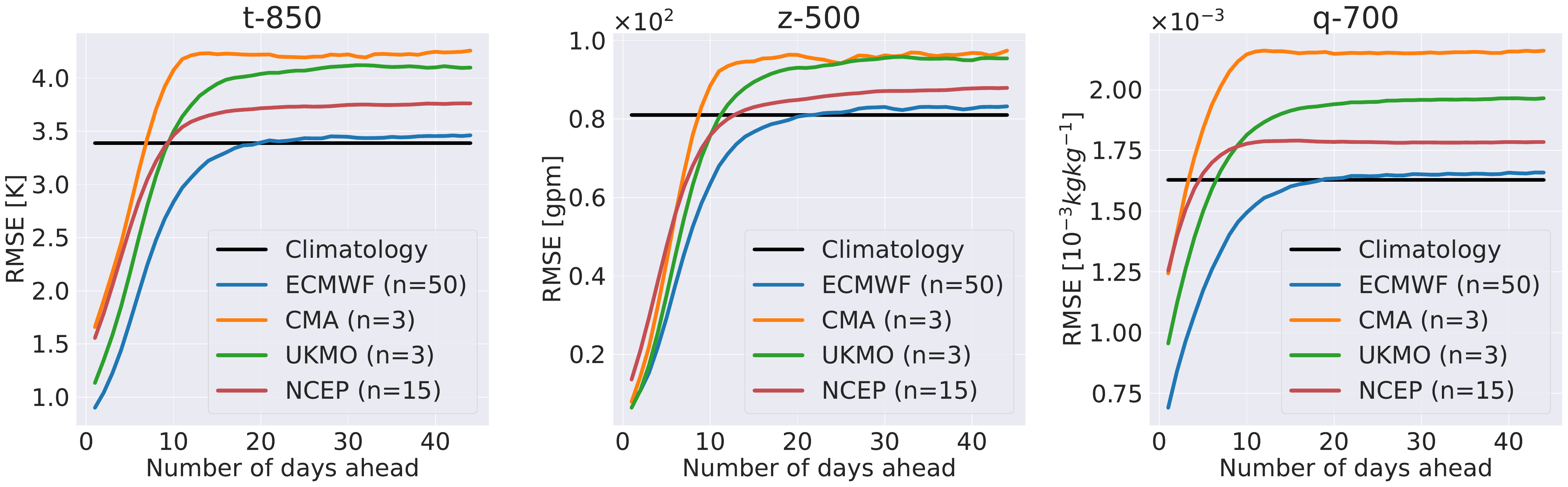}
        \caption{RMSE ($\downarrow$ is better)}
    \end{subfigure}
    \hfill
    \begin{subfigure}{0.8\textwidth}
        \includegraphics[width=\textwidth]{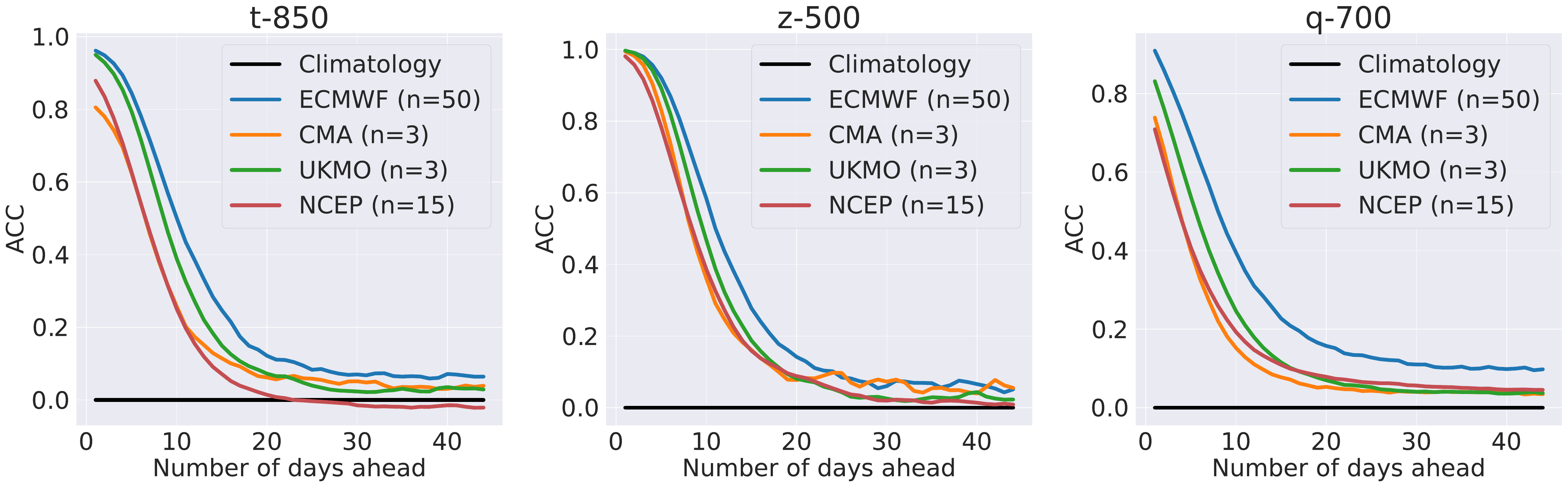}
        \caption{ACC ($\uparrow$ is better)}
    \end{subfigure}
    \hfill
    \begin{subfigure}{0.8\textwidth}
        \includegraphics[width=\textwidth]{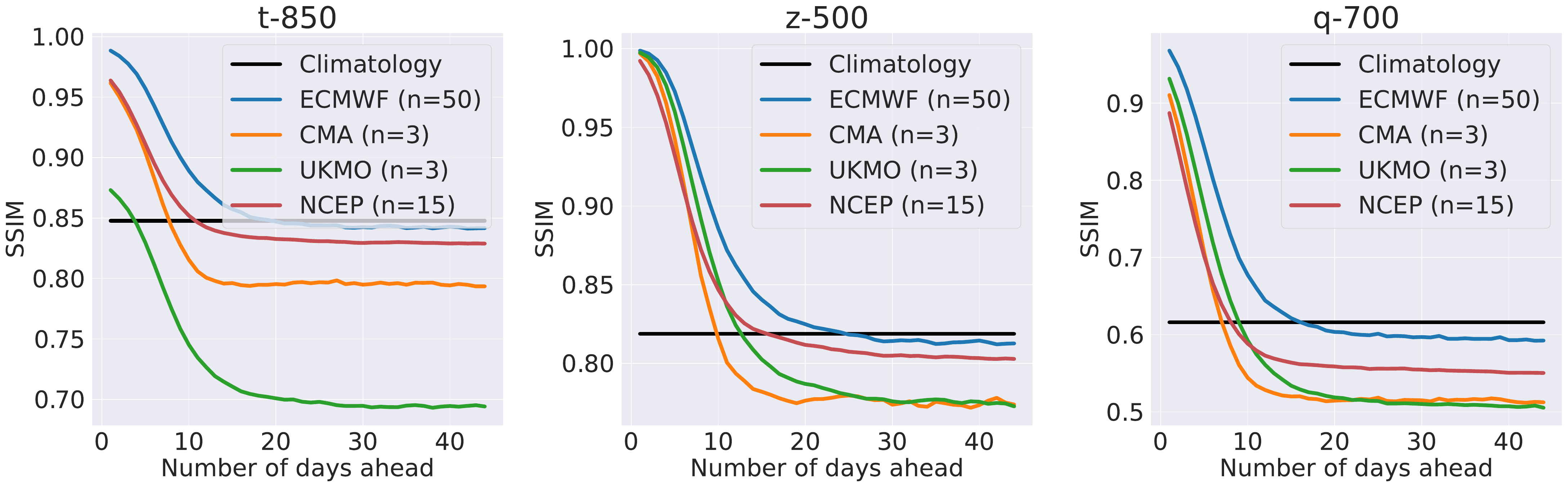}
        \caption{MS-SSIM ($\uparrow$ is better)}
    \end{subfigure}
    \hfill
    \begin{subfigure}{0.8\textwidth}
        \includegraphics[width=\textwidth]{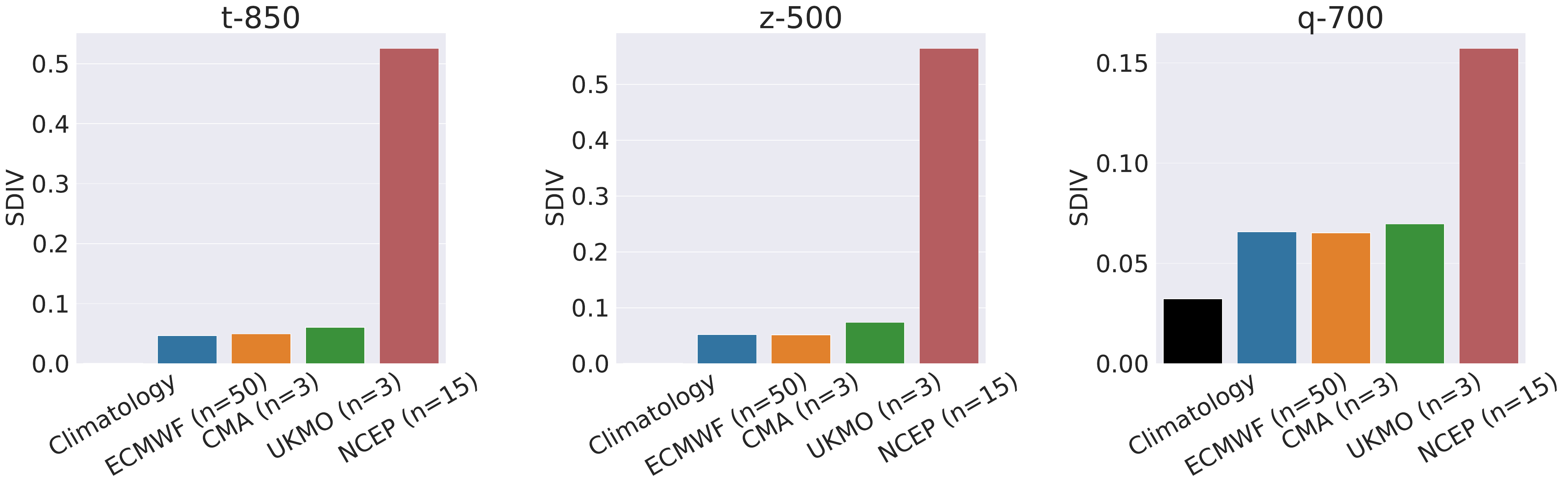}
        \caption{SpecDiv ($\downarrow$ is better)}
    \end{subfigure}
    
\caption{Evaluation results between baseline climatology (black line) and physics-based ensembles from ECMWF, CMA, UKMO, NCEP. Overall, we observe that ensemble forecasts perform better than their deterministic counterparts.}
\label{si-fig:center_ens}
\end{figure*}

\begin{figure*}[h]
    \centering
    \begin{subfigure}{0.8\textwidth}
        \includegraphics[width=\textwidth]{imgs/center_ratio_rmse.pdf}
        \caption{RMSE: ensemble improves deterministic forecasts if $\texttt{ratio} < 1$}
    \end{subfigure}
    \hfill
    \begin{subfigure}{0.8\textwidth}
        \includegraphics[width=\textwidth]{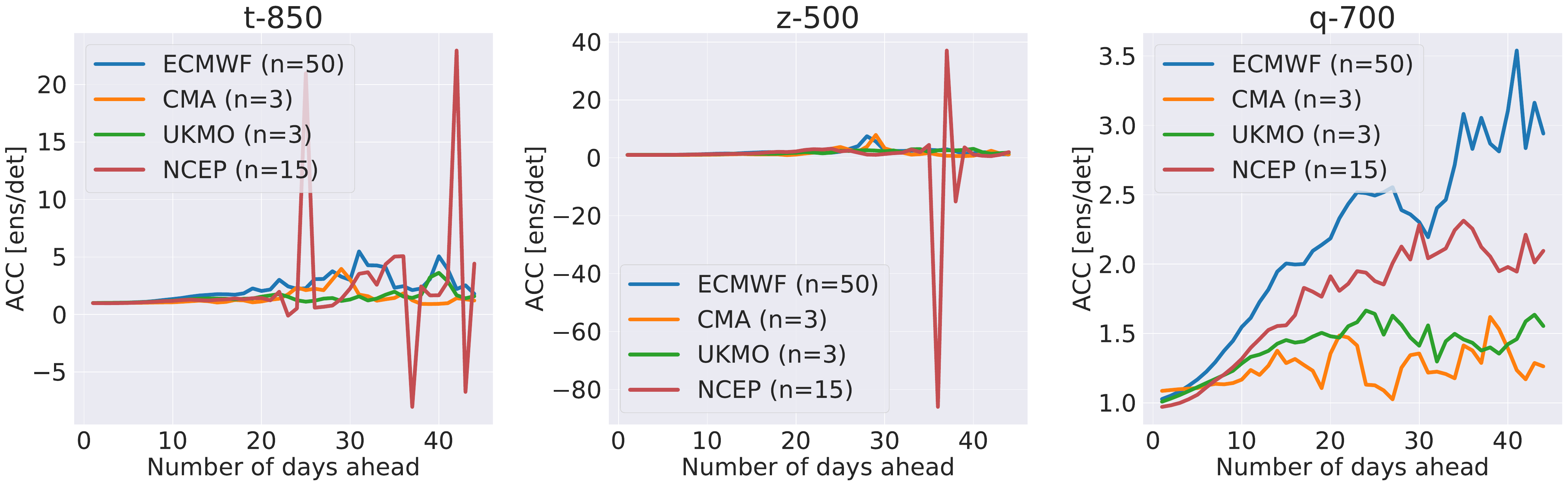}
        \caption{ACC: ensemble improves deterministic forecasts if $\texttt{ratio} > 1$}
    \end{subfigure}
    \hfill
    \begin{subfigure}{0.8\textwidth}
        \includegraphics[width=\textwidth]{imgs/center_ratio_ssim.pdf}
        \caption{MS-SSIM: ensemble improves deterministic forecasts if $\texttt{ratio} > 1$}
    \end{subfigure}
    \hfill
    \begin{subfigure}{0.8\textwidth}
        \includegraphics[width=\textwidth]{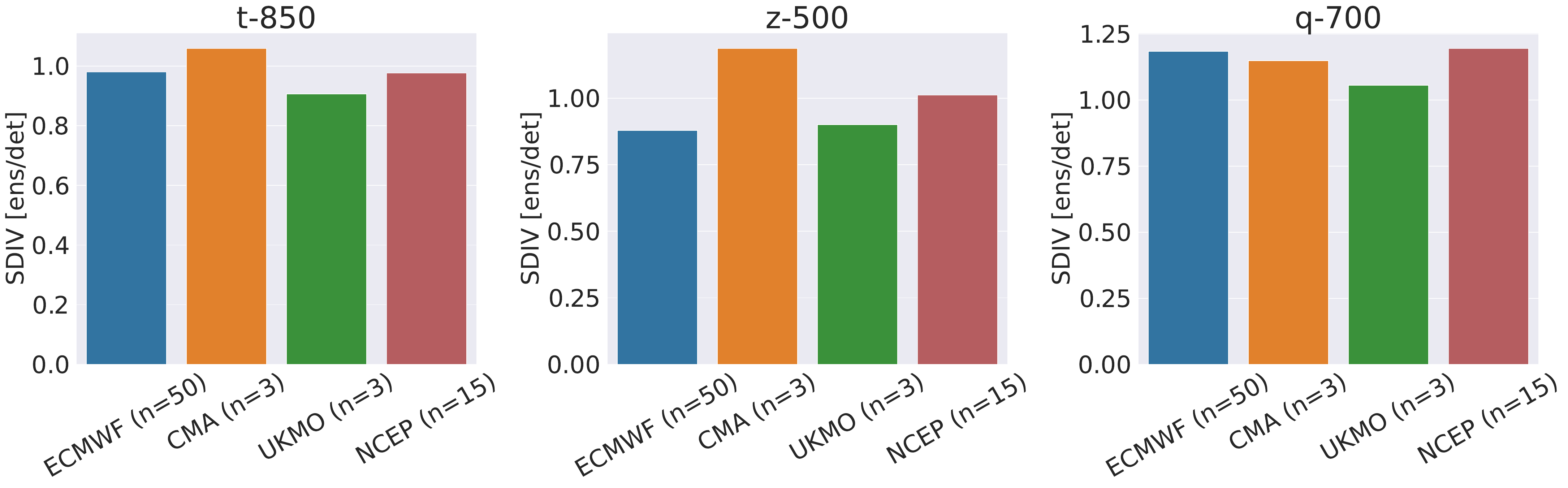}
        \caption{SpecDiv: ensemble improves deterministic forecasts if $\texttt{ratio} < 1$}
    \end{subfigure}
    
\caption{Metrics ratio e.g., $\texttt{RMSE}_{ens} / \texttt{RMSE}_{det}$ between ensemble and deterministic forecasts, where the former improves the latter by accounting for IC uncertainty that can lead to long-range instability and trajectory divergences. \underline{Note}: $n$ represents the number of ensemble members. The ratio for \texttt{ACC} fluctuates as the scalar value approaches 0.}
\label{si-fig:center_ratio}
\end{figure*}

\begin{figure*}[t]
    \centering
    \begin{subfigure}{0.8\textwidth}
        \includegraphics[width=\textwidth]{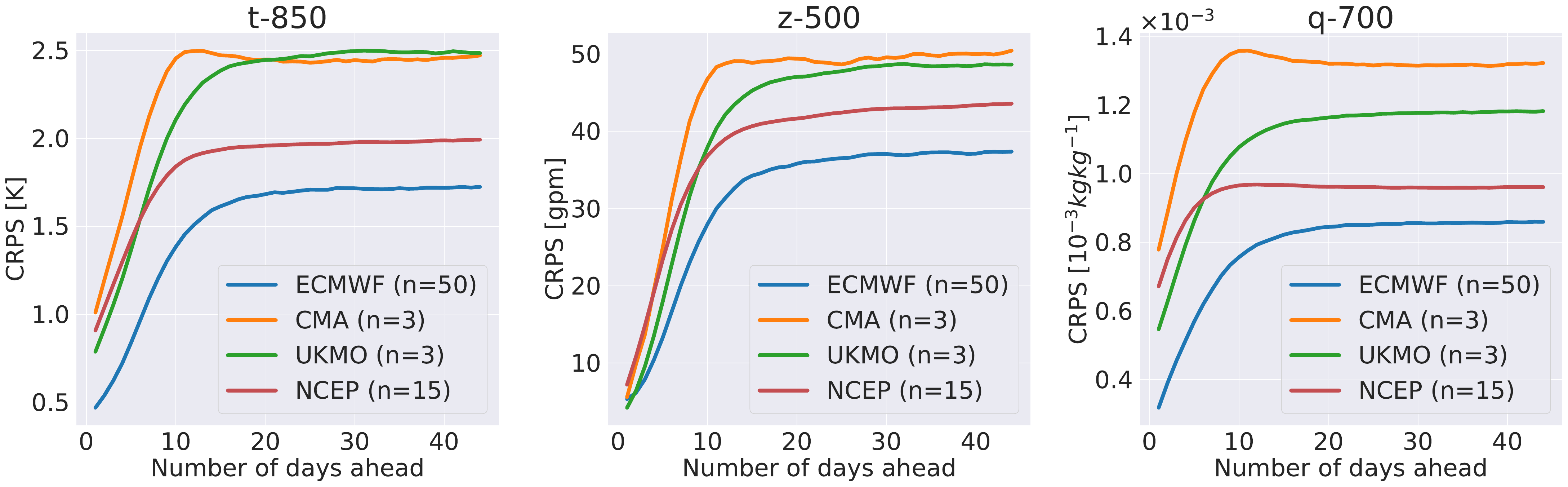}
        \caption{CRPS ($\downarrow$ is better)}
    \end{subfigure}
    \hfill
    \begin{subfigure}{0.8\textwidth}
        \includegraphics[width=\textwidth]{imgs/center_probs_crpss.pdf}
        \caption{CRPSS ($>0$ is better)}
    \end{subfigure}
    \hfill
    \begin{subfigure}{0.8\textwidth}
        \includegraphics[width=\textwidth]{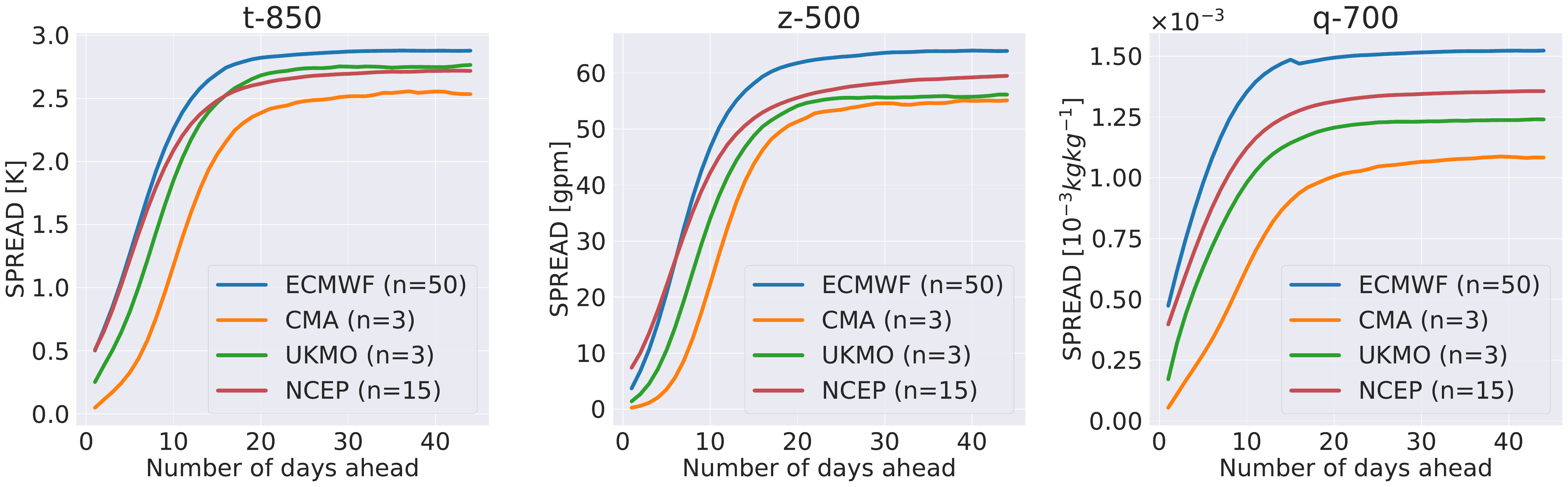}
        \caption{Spread}
    \end{subfigure}
    \hfill
    \begin{subfigure}{0.8\textwidth}
        \includegraphics[width=\textwidth]{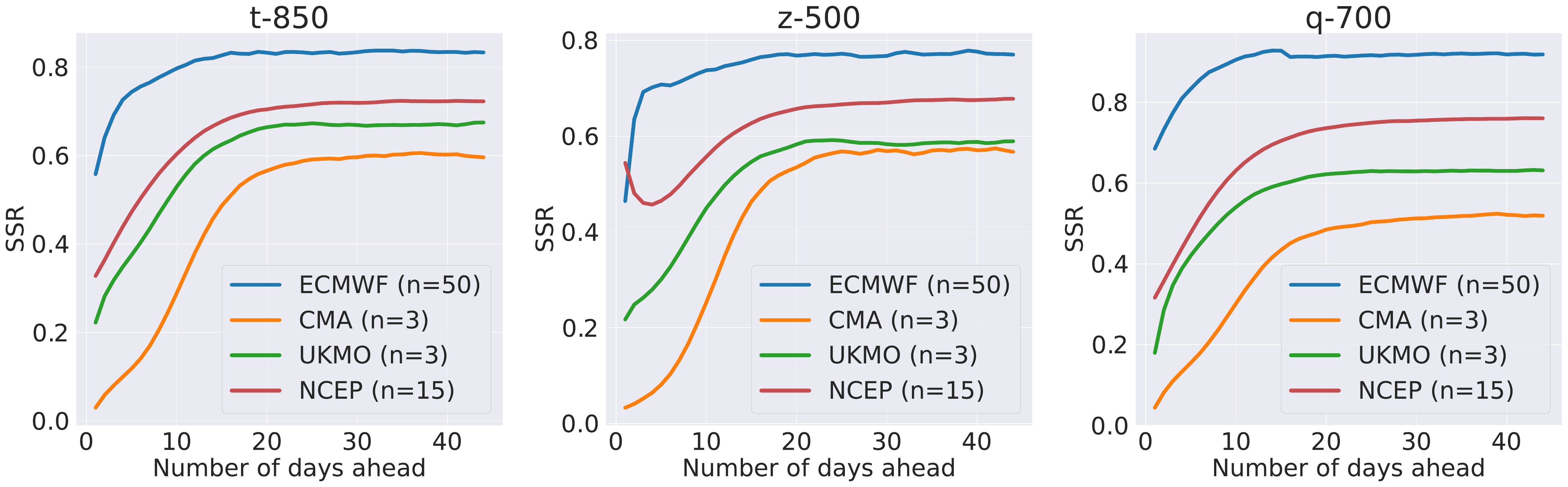}
        \caption{SSR ($>0$ is better)}
    \end{subfigure}
    
\caption{Probabilistic evaluation on ensemble forecasts indicating current skill limits of 15 days. \underline{Note}: $n$ represents the number of ensemble members.}
\label{si-fig:center_probs}
\end{figure*}
\clearpage

\subsection{Effects of Different Autoregressive Training Steps; \texttt{lead\_time}}
\begin{figure*}[h]
    \centering
    \begin{subfigure}{0.8\textwidth}
        \includegraphics[width=\textwidth]{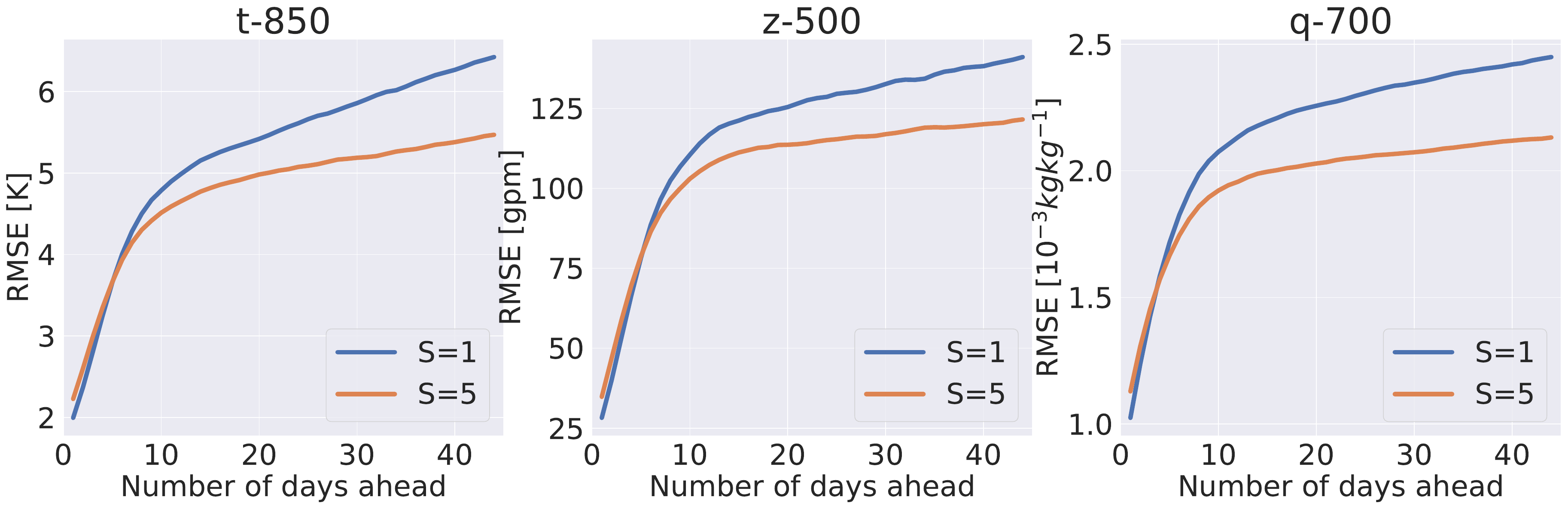}
        \caption{RMSE ($\downarrow$ is better)}
    \end{subfigure}
    \hfill
    \begin{subfigure}{0.8\textwidth}
        \includegraphics[width=\textwidth]{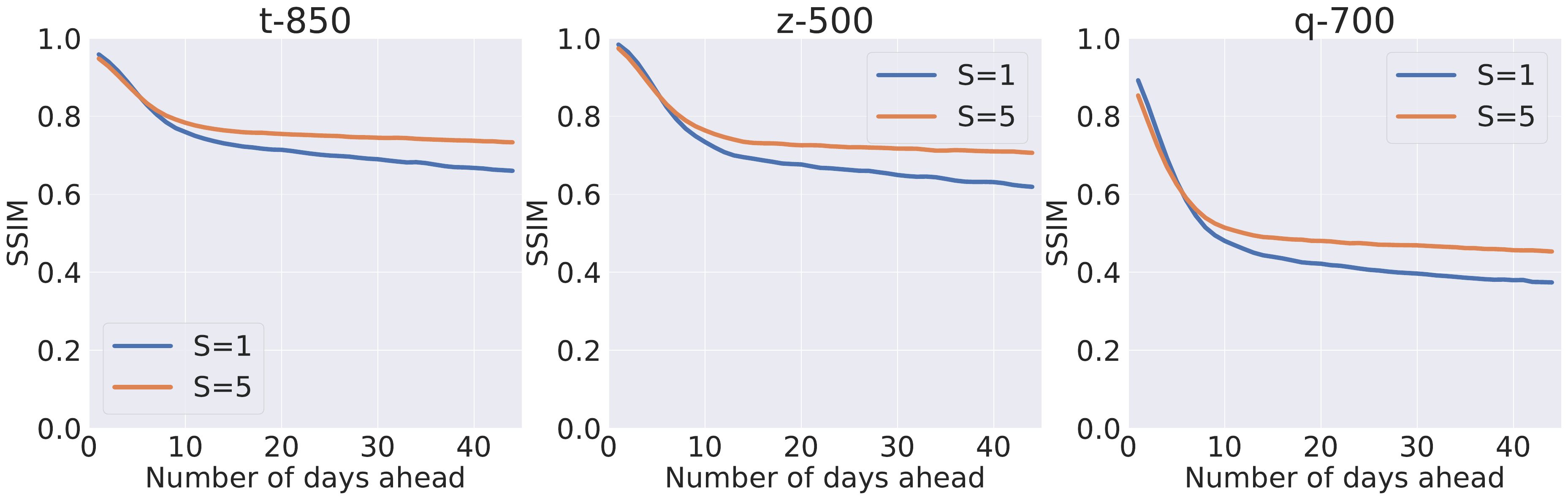}
        \caption{MS-SSIM ($\uparrow$ is better)}
    \end{subfigure}
    \hfill
    \begin{subfigure}{0.8\textwidth}
        \includegraphics[width=\textwidth]{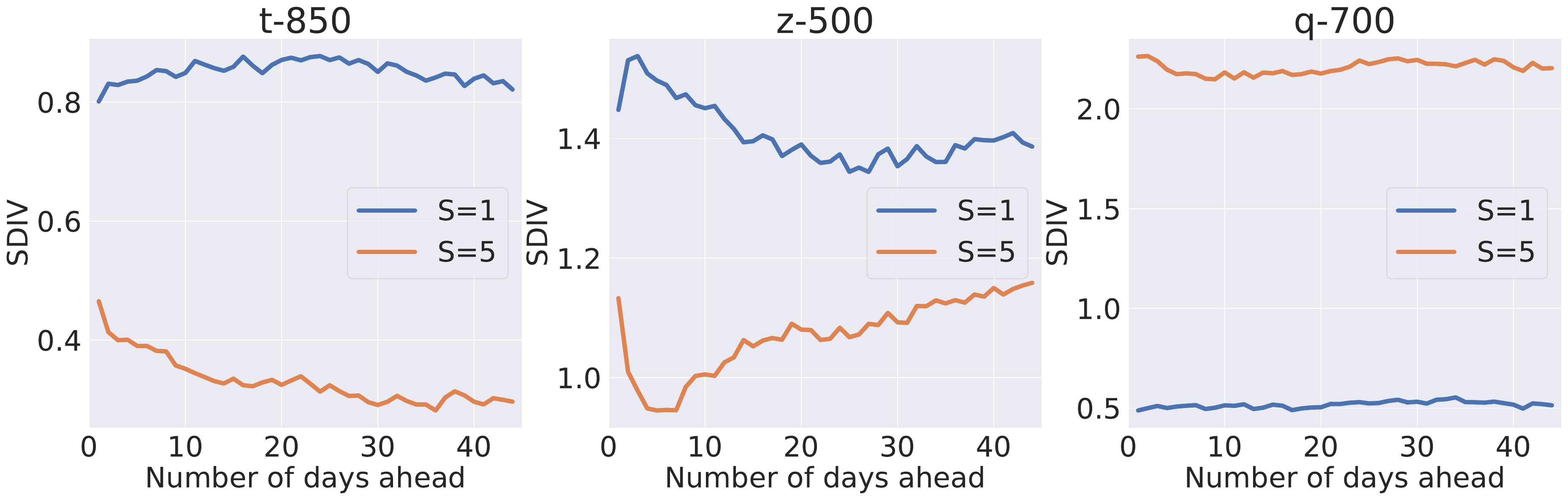}
        \caption{SpecDiv ($\downarrow$ is better)}
    \end{subfigure}
    
\caption{Ablation results for incorporating temporal information in an autoregressive scheme for long-range forecast using UNet models. The x-axis represents the number of forecasting days for t-850, z-500, q-700 representative tasks. Blue and orange lines illustrate autoregressive scheme with $s=1$ and $s=5$ respectively. Overall we observe that incorporating temporal information improve the vision-based metrics even at longer forecasting timesteps. However, the converse trend is true where incorporating temporal context makes S2S forecast worse off in some physics-based scores.}

\label{si-fig:autoreg}
\end{figure*}

We showcased more results for autoregressive training strategy. In this case, we performed autoregressive training using either 1 or 5 iterative steps (\texttt{n\_step}; $s$). As illustrated in Figure \ref{si-fig:autoreg}, we observe that incorporating temporal information improve the vision-based metrics even at longer forecasting timesteps, with lower RMSE, higher MS-SSIM. However, the converse trend is true incorporating temporal context makes S2S forecast worse off in some physics-based scores. The modified loss function for training a model with multiple autoregressive steps is:

\begin{equation}
\mathcal{L} = \frac{1}{|S|}\sum_{i=1}^s\mathcal{L}(\mathbf{\hat{Y}_{t+s_i}}\mathbf{Y_{t+s_i}}), \forall s_i \in S
\end{equation}

Here $S = \{1, \cdots, s\}$ and $s \in \mathbb{N}^+$ is the autoregressive steps. For this work, we set $s=5$. 

\clearpage
\subsection{Effects of Subset Optimization; \texttt{headline\_vars}}

In many cases, we seek to train data-driven models so that they are able to perform well on \emph{all} states by optimizing for the full state of the next forecasting timestep $t+1$, that is,
$$\phi^\ast = \operatorname*{argmin}_\phi \mathcal{L}(\mathbf{\hat{Y}}_{t+1}, \mathbf{Y}_{t+1})$$
where $\mathcal{L}$ is any loss function. This task is especially useful for building emulators that act as surrogates for the more expensive physics-based NWP models \cite{nguyen2023climax}. 

Although the first task is useful for learning the full complex interaction between variables, it is relatively difficult due to the intrinsic high-dimensionality of the  data. As a result, we introduce a second task that allows for the optimization on a subset of variables of interest ($\mathbf{Y'} \in \mathbf{Y}$): $$\phi^\ast=\operatorname*{argmin}_\phi\mathcal{L}(\mathbf{\hat{Y}'}_{t+1}, \mathbf{Y'}_{t+1})$$ 

Here $\mathbf{Y}^\prime_{t+1} = \{\text{t-850, z-500, q-700}\}$, and we train them using 5 autoregressive steps i.e., $\texttt{n\_step}=5$.

\begin{table*}[h]
    \centering
    \caption{Long-range forecasting ($\Delta t=44$) results on select metrics and target variables between physics-based and data-driven models. Results are for Task 1 (full) and Task 2 (sparse). \emph{(*) Baseline model that uses privileged information (observations) to make prediction.}}
    \begin{threeparttable}
    \begin{tabular}{c|ccc|ccc|ccc}
        \toprule
        & \multicolumn{3}{c}{RMSE $\downarrow$} & \multicolumn{3}{c}{MS-SSIM $\uparrow$} & \multicolumn{3}{c}{SpecDiv $\downarrow$}\\
        \midrule
        Models & \makecell{T850 \\($K$)} & \makecell{Z500 \\($gpm$)} & \makecell{Q700 \\($\times 10^{-3}$)} & T850 & Z500 & Q700& T850 & Z500 & Q700\\
        \midrule
        
        Climatology\tnote{*} & \textbf{3.39} & \textbf{81.0} & \textbf{1.62} & \textbf{0.85} & \textbf{0.82} & \textbf{0.62} & \textbf{0.01} & \textbf{0.01} & \textbf{0.03} \\
        
        Persistence\tnote{*} & 5.88 & 127.8 & 2.47 & 0.71 & 0.69 & 0.41 & 0.02 & 0.03 & 0.05 \\
        
        UKMO & 5.00 & 116.2 & 2.32 & 0.64 & 0.71 & 0.43 & 0.06 & 0.09 & 0.07\\
        
        NCEP & 4.90 & 116.7 & 2.30 & 0.75 & 0.71 & 0.43 & 0.53 & 0.55 & 0.10\\
        
        CMA & 5.08 & 118.7 & 2.49 & 0.75 & 0.72 & 0.45 & 0.05 & 0.04 & 0.06\\
        
        ECMWF & 4.72 & 115.1 & 2.30 & 0.75 & 0.72 & 0.44 & 0.06 & 0.07 & 0.06\\
        
        \midrule
        & \multicolumn{9}{c}{Task 1: Full Dynamics Prediction} \\
        \midrule
        
        Lagged AE & 5.55 & 122.4 & 2.03  & 0.74 & 0.71 & 0.47 & 0.18 & 2.44 & 0.21\\
        
        ResNet & 5.67 & 125.3 & 2.07 & 0.73 & 0.70 & 0.47 & 0.21 & 0.37 & 0.26\\
        
        UNet & 5.47 & 121.5 & 2.13 & 0.73 & 0.71 & 0.45 & 0.30 & 1.16 & 2.20 \\
        
        FNO & \textbf{5.06} & \textbf{112.5}  & \textbf{1.95} & \textbf{0.75} & \textbf{0.73} & \textbf{0.51} & \textbf{0.18} & \textbf{0.11} & \textbf{0.10} \\
        
        \midrule
        & \multicolumn{9}{c}{Task 2: Sparse Dynamics Prediction} \\
        \midrule
        
        Lagged AE & 5.39 & 119.0 & 2.12  & 0.75 & 0.73 & 0.48 & 0.52 & 1.41 & 0.29\\
        
        ResNet & 5.80 & 124.1 & 2.18 & 0.74 & 0.72 & 0.46 & 0.33 & 1.22 & 0.09\\
        
        UNet & 5.57 & 120.2 & 2.18 & 0.74 & 0.71 & 0.45 & 1.20 & 1.08 & \textbf{0.07} \\
        
        FNO & \textbf{4.73} & \textbf{101.8}  & \textbf{1.91} & \textbf{0.79} & \textbf{0.76} & \textbf{0.52} & \textbf{0.18} & \textbf{0.23} & 0.21 \\
        \bottomrule
    \end{tabular}
    \end{threeparttable}
    \label{si-tab:ablation_s2s}
\end{table*}

\noindent Overall, we find models that attempt to preserve spectral structures (e.g., FNO) perform better on all metrics, deterministic and physics-based. Also, Task 2 (sparse) appears to be easier than Task 1 (full). Nonetheless, they are still performing worse than climatology.
\clearpage

\subsection{Effects of Ensemble Forecasts}
\label{si-sec:ml_ensemble}
This section provides additional results for data-driven ensemble approach, and follow similar evaluation process as the physics-based counterpart.

\begin{figure*}[h]
    \centering
    \begin{subfigure}{0.8\textwidth}
        \includegraphics[width=\textwidth]{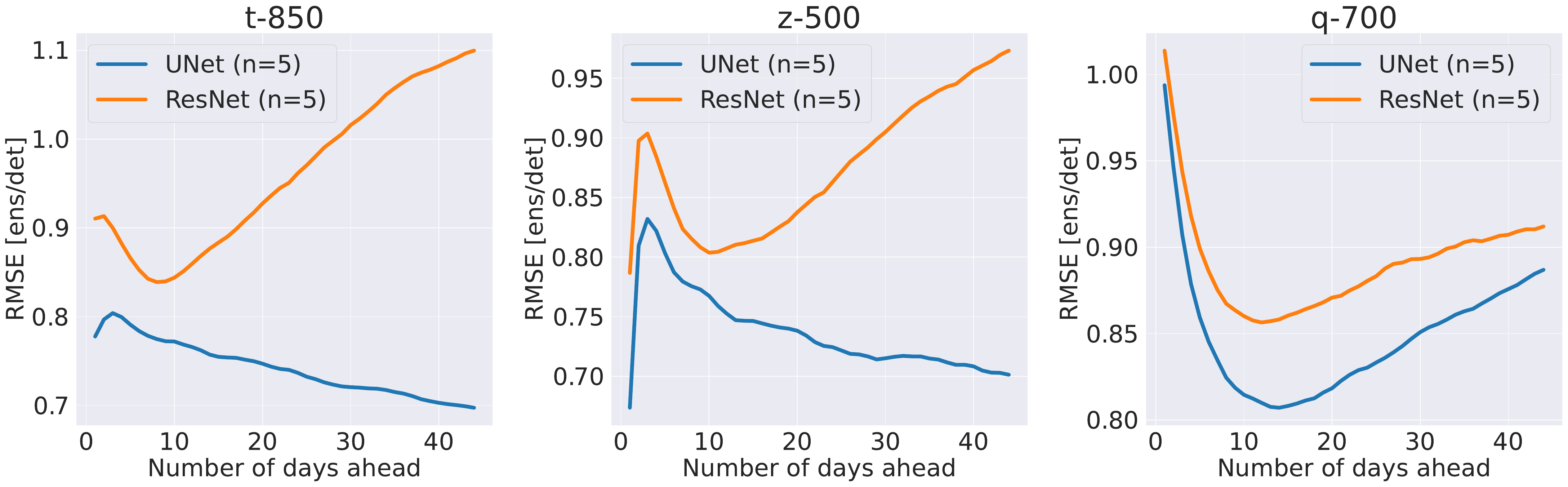}
        \caption{RMSE: ensemble improves deterministic forecasts if $\texttt{ratio} < 1$}
    \end{subfigure}
    \hfill
    \begin{subfigure}{0.8\textwidth}
        \includegraphics[width=\textwidth]{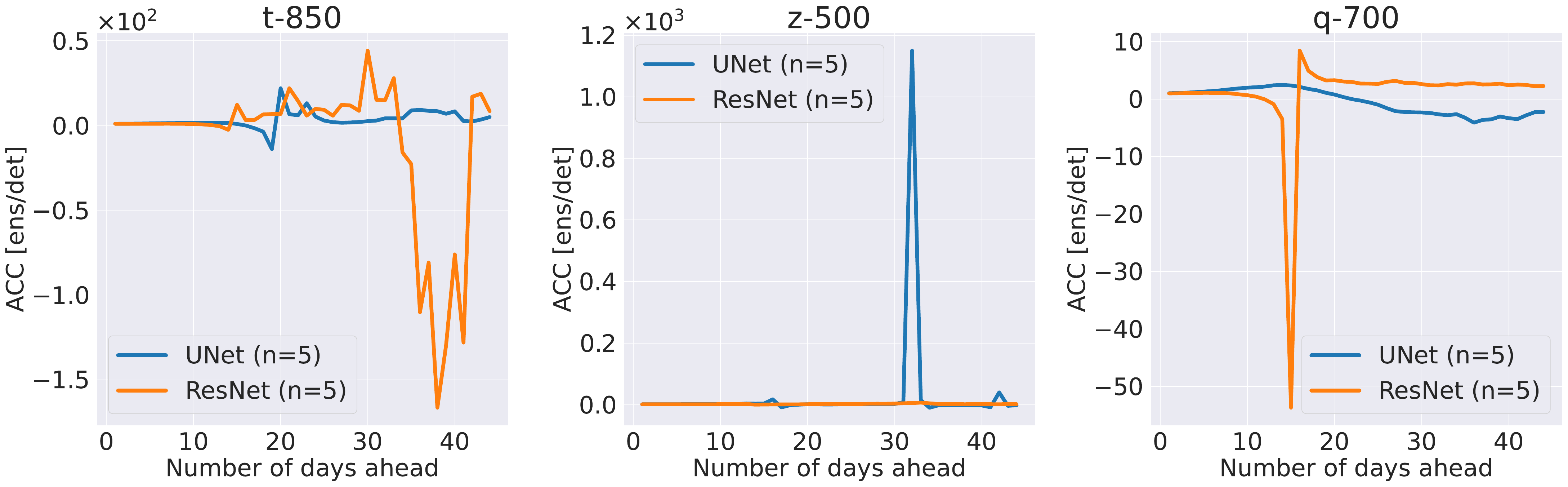}
        \caption{ACC: ensemble improves deterministic forecasts if $\texttt{ratio} > 1$}
    \end{subfigure}
    \hfill
    \begin{subfigure}{0.8\textwidth}
        \includegraphics[width=\textwidth]{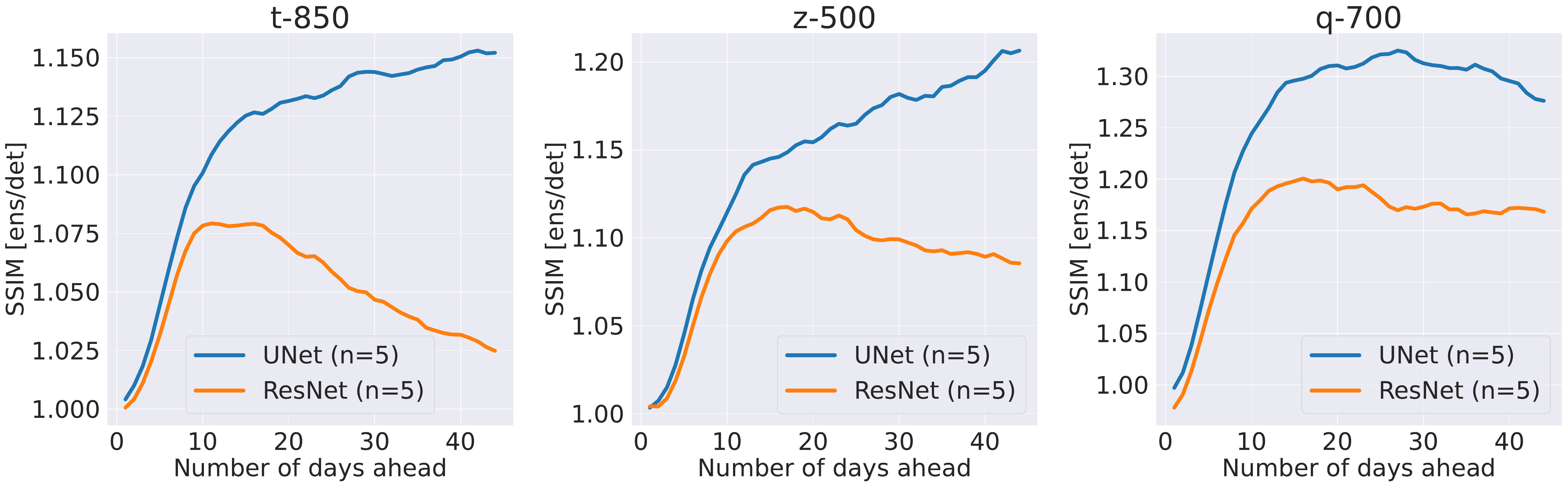}
        \caption{MS-SSIM: ensemble improves deterministic forecasts if $\texttt{ratio} > 1$}
    \end{subfigure}
    \hfill
    \begin{subfigure}{0.8\textwidth}
        \includegraphics[width=\textwidth]{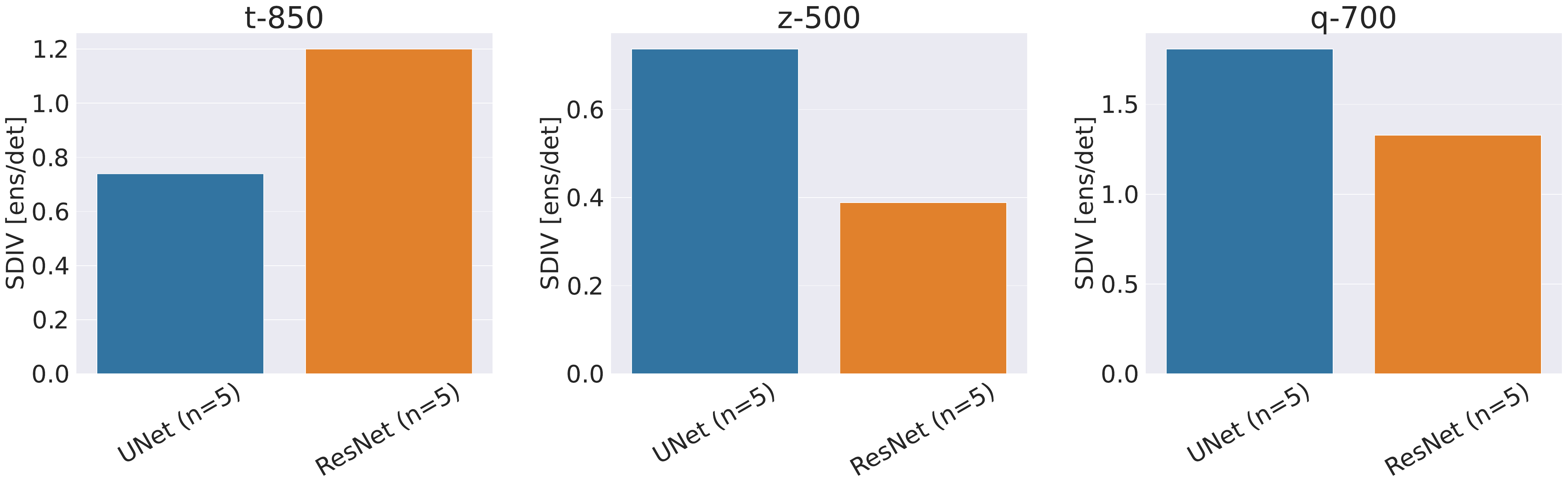}
        \caption{SpecDiv: ensemble improves deterministic forecasts if $\texttt{ratio} < 1$}
    \end{subfigure}
    
\caption{Metrics ratio e.g., $\texttt{RMSE}_{ens} / \texttt{RMSE}_{det}$ between ensemble and deterministic forecasts, where the former improves the latter by accounting for IC uncertainty that can lead to long-range instability and trajectory divergences. \underline{Note}: $n$ represents the number of ensemble members.}
\label{si-fig:ml_ratio}
\end{figure*}

\begin{figure*}[h]
    \centering
    \begin{subfigure}{0.8\textwidth}
        \includegraphics[width=\textwidth]{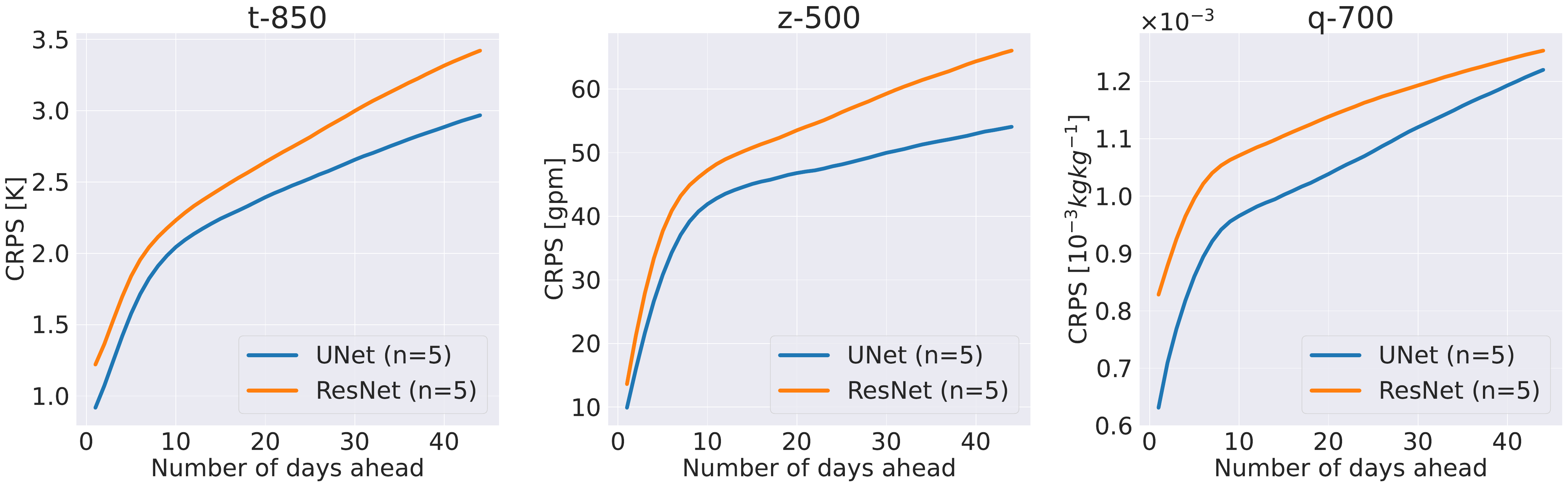}
        \caption{CRPS ($\downarrow$ is better)}
    \end{subfigure}
    \hfill
    \begin{subfigure}{0.8\textwidth}
        \includegraphics[width=\textwidth]{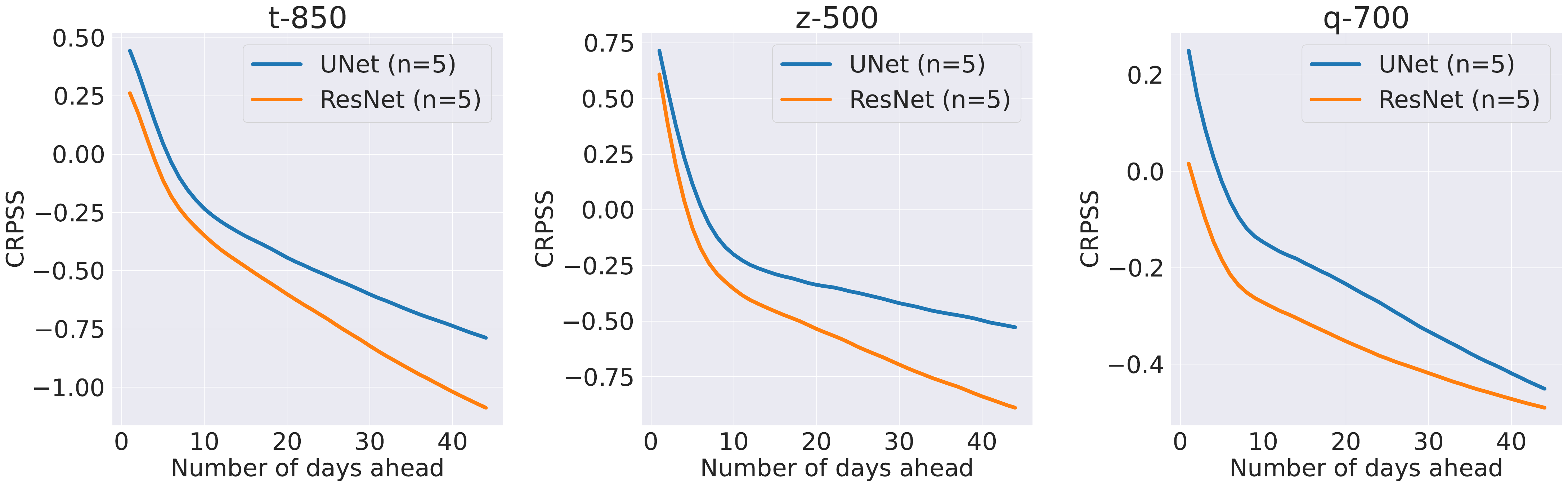}
        \caption{CRPSS ($>0$ is better)}
    \end{subfigure}
    \hfill
    \begin{subfigure}{0.8\textwidth}
        \includegraphics[width=\textwidth]{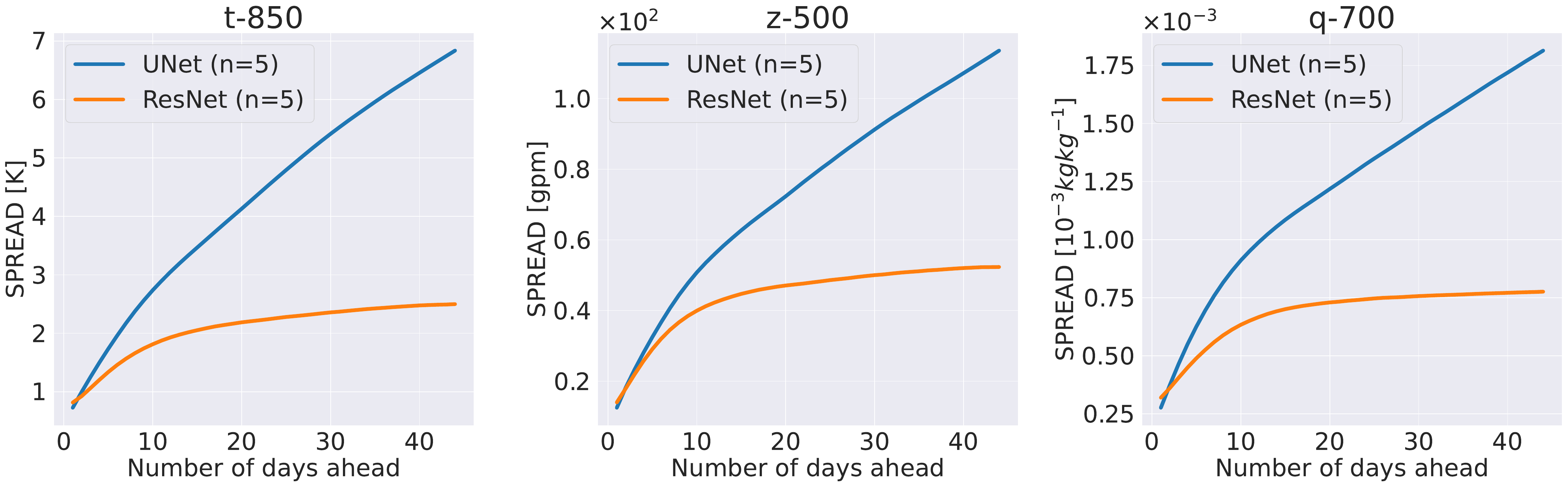}
        \caption{Spread}
    \end{subfigure}
    \hfill
    \begin{subfigure}{0.8\textwidth}
        \includegraphics[width=\textwidth]{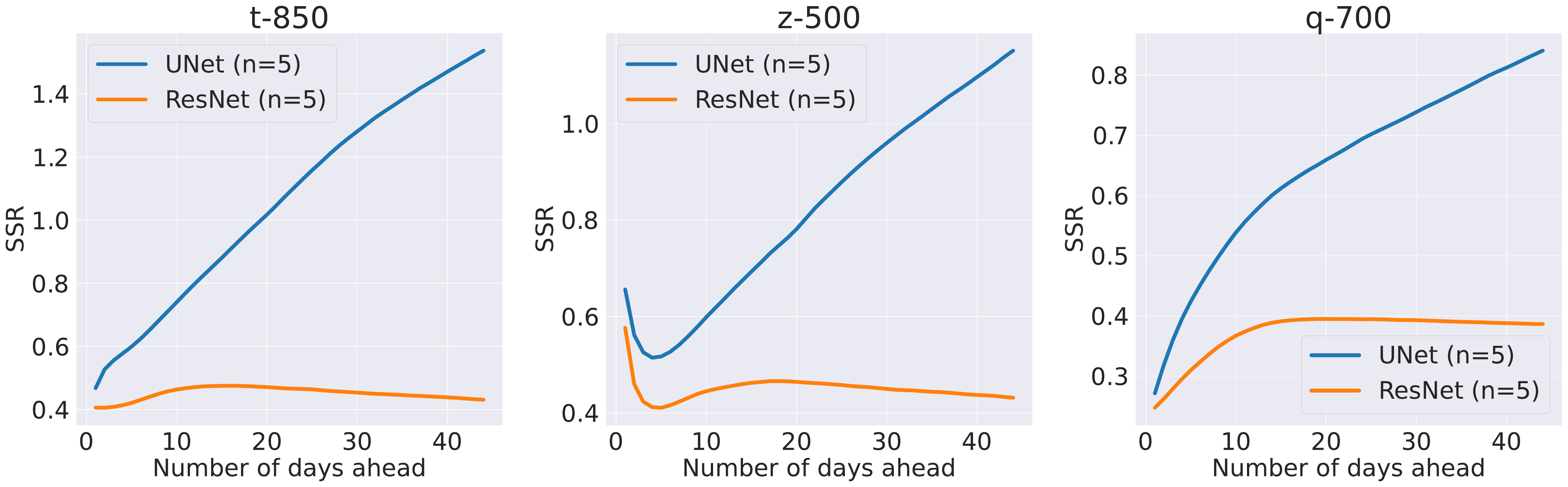}
        \caption{SSR ($>0$ is better)}
    \end{subfigure}
    
\caption{Probabilistic evaluation on ensemble forecasts. \underline{Note}: $n$ represents the number of ensemble members.}
\label{si-fig:ml_probs}
\end{figure*}
\clearpage

\subsection{Power Spectra}
\begin{figure*}[h]
    \centering
    \begin{subfigure}{\textwidth}
        \includegraphics[width=\textwidth]{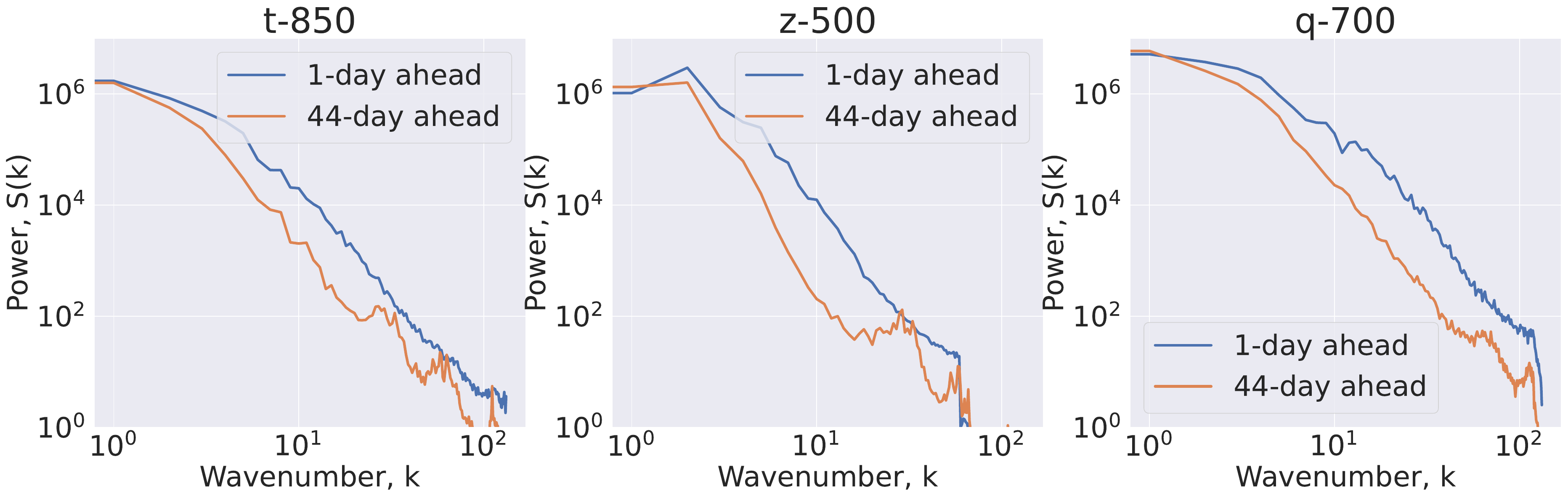}
        \caption{Task 1}
    \end{subfigure}
    \hfill
    \begin{subfigure}{\textwidth}
        \includegraphics[width=\textwidth]{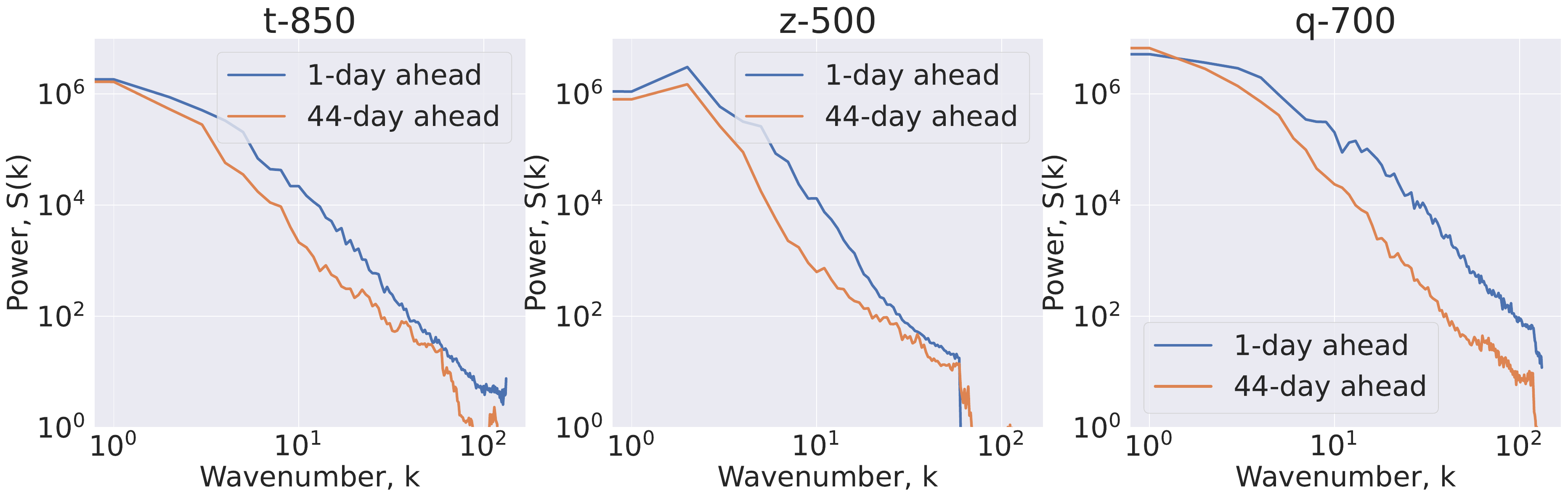}
        \caption{Task 2}
    \end{subfigure}
    
\caption{Power spectra for ViT/ClimaX demonstrating energy decay/divergence especially for high $k$ as lead time grows.}
\label{si-fig:specdiv_climax}
\end{figure*}
\newpage

\subsection{Qualitative Evaluation}
\begin{figure*}[h]
    \centering
    \begin{subfigure}{0.48\textwidth}
        \includegraphics[width=\textwidth]{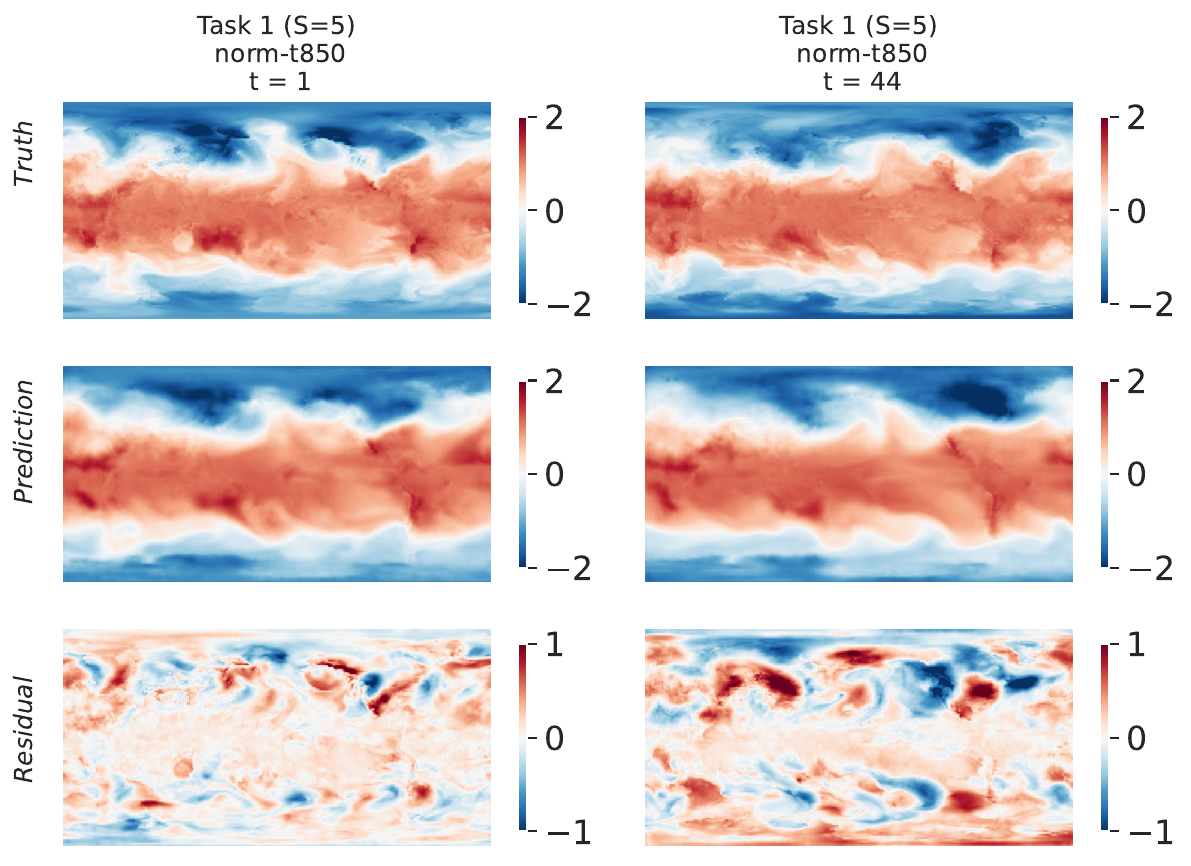}
        \caption{Task 1}
    \end{subfigure}
    \hfill
    \begin{subfigure}{0.48\textwidth}
        \includegraphics[width=\textwidth]{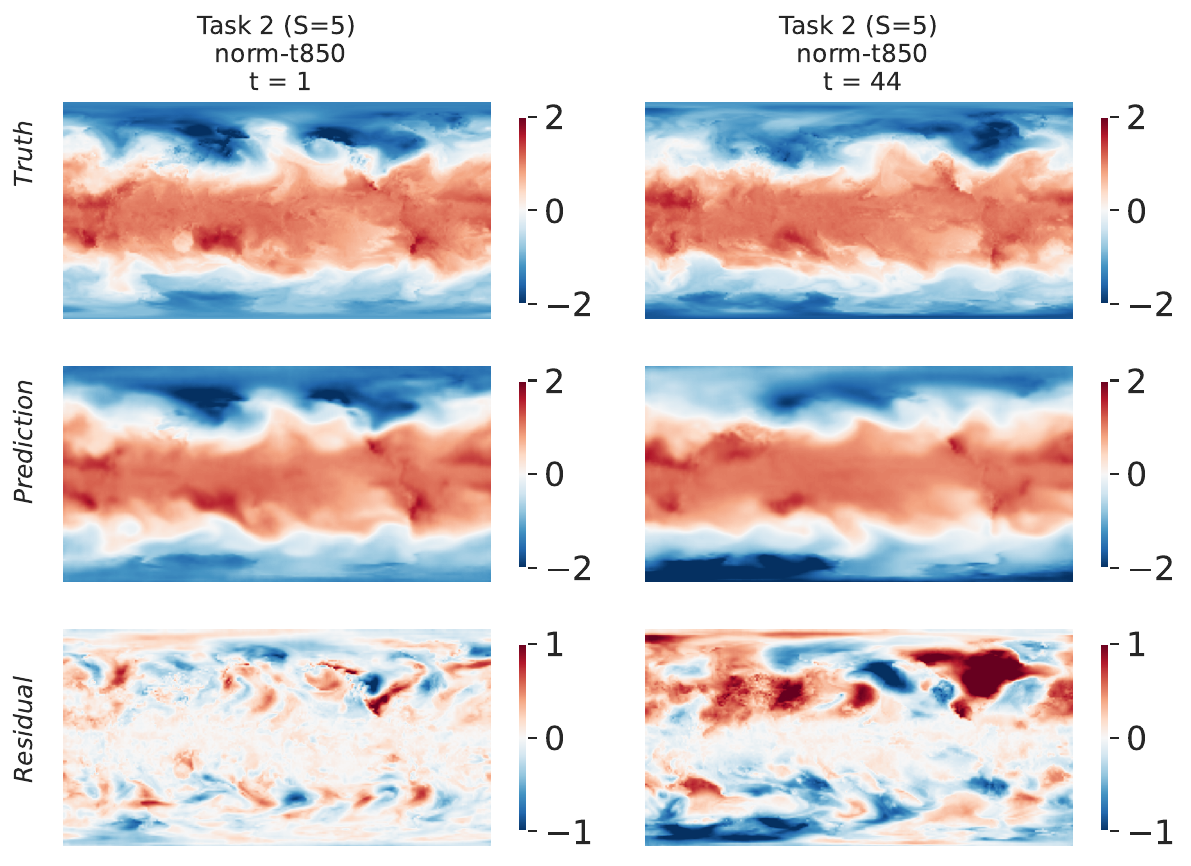}
        \caption{Task 2}
    \end{subfigure}
    
\caption{Normalized t@850-hpa qualitative results for UNet-autoregressive (S=5).}
\label{si-fig:preds_t850_s5}
\end{figure*}

\begin{figure*}[h]
    \centering
    \begin{subfigure}{0.48\textwidth}
        \includegraphics[width=\textwidth]{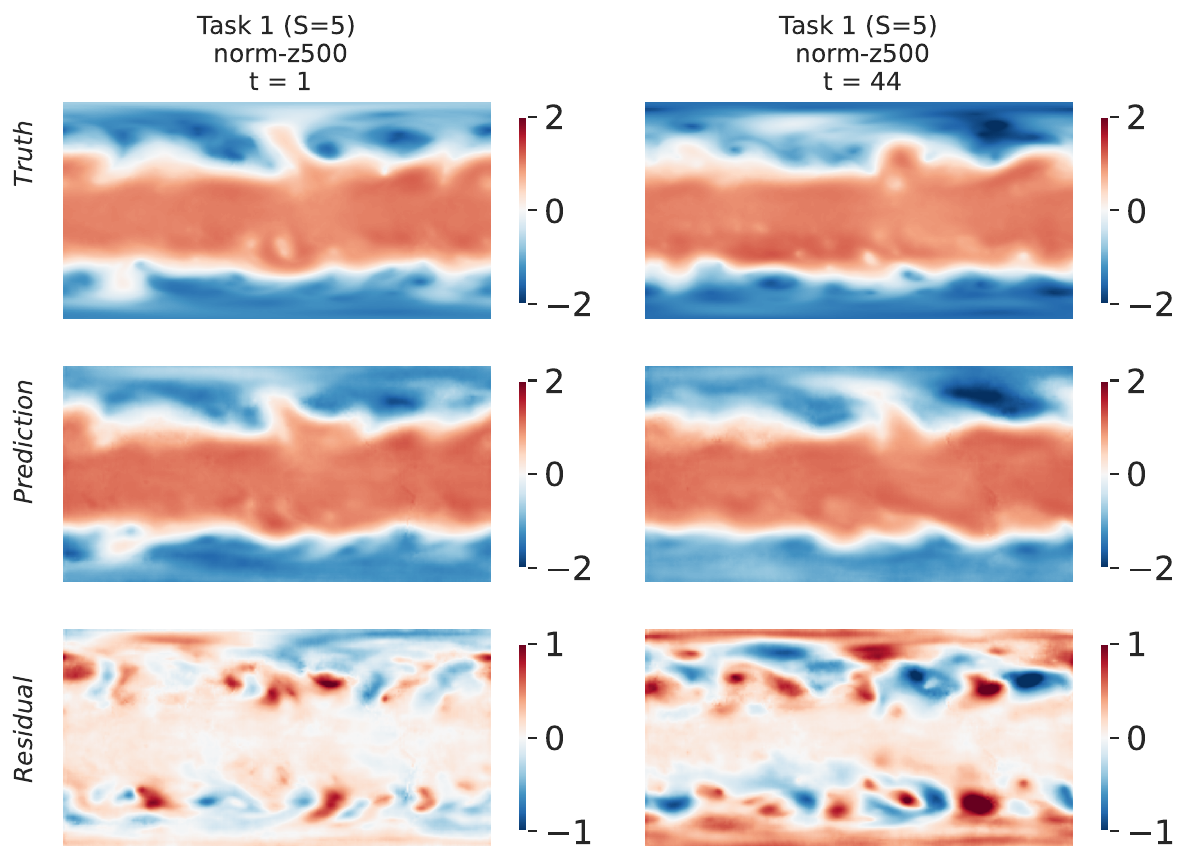}
        \caption{Task 1}
    \end{subfigure}
    \hfill
    \begin{subfigure}{0.48\textwidth}
        \includegraphics[width=\textwidth]{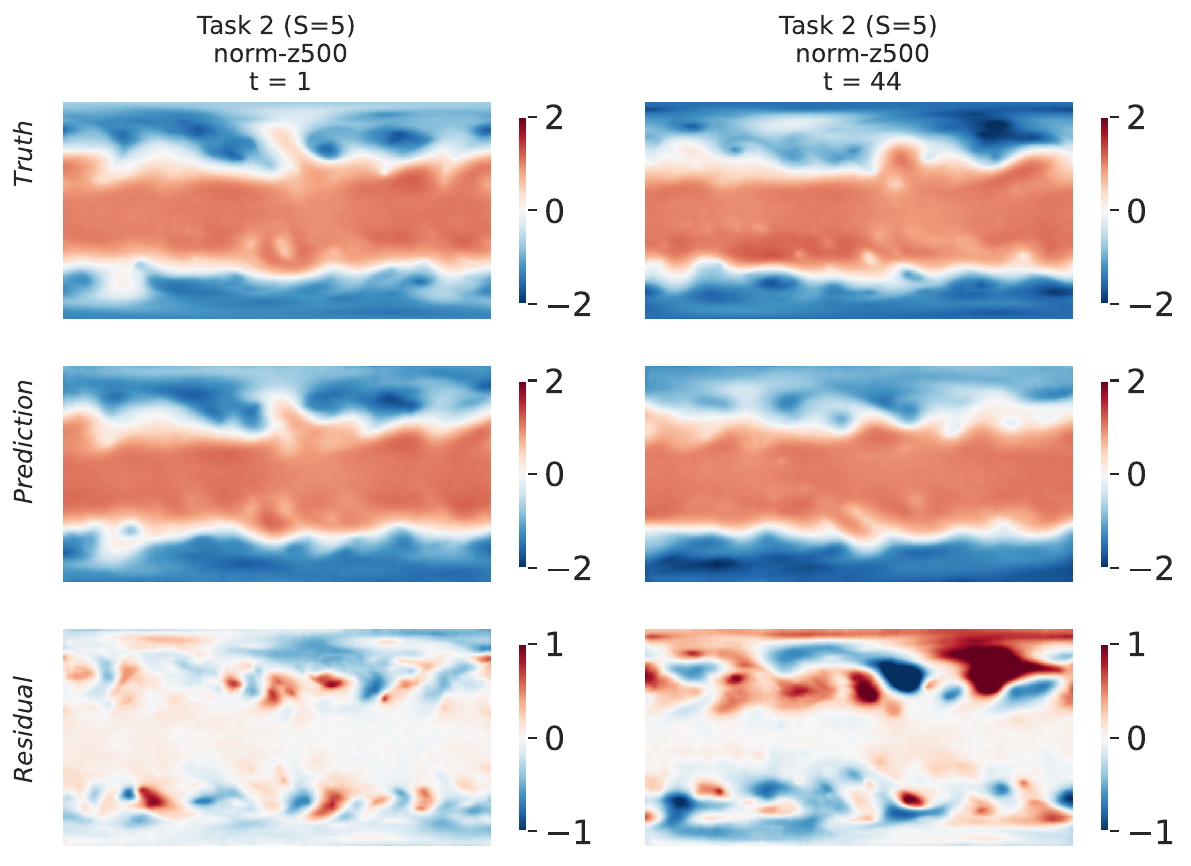}
        \caption{Task 2}
    \end{subfigure}
    
\caption{Normalized z@500-hpa qualitative results for UNet-autoregressive (S=5).}
\label{si-fig:preds_z500_s5}
\end{figure*}

\begin{figure*}[h]
    \centering
    \begin{subfigure}{0.48\textwidth}
        \includegraphics[width=\textwidth]{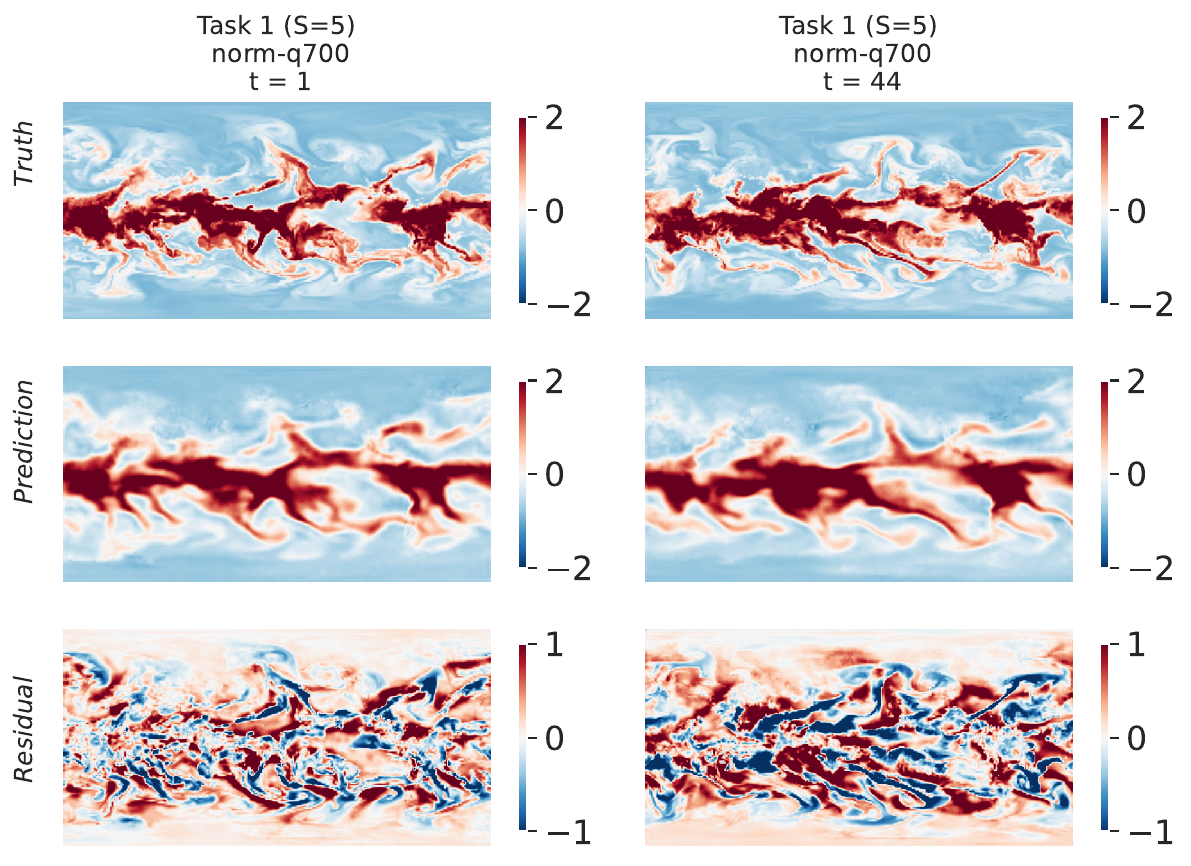}
        \caption{Task 1}
    \end{subfigure}
    \hfill
    \begin{subfigure}{0.48\textwidth}
        \includegraphics[width=\textwidth]{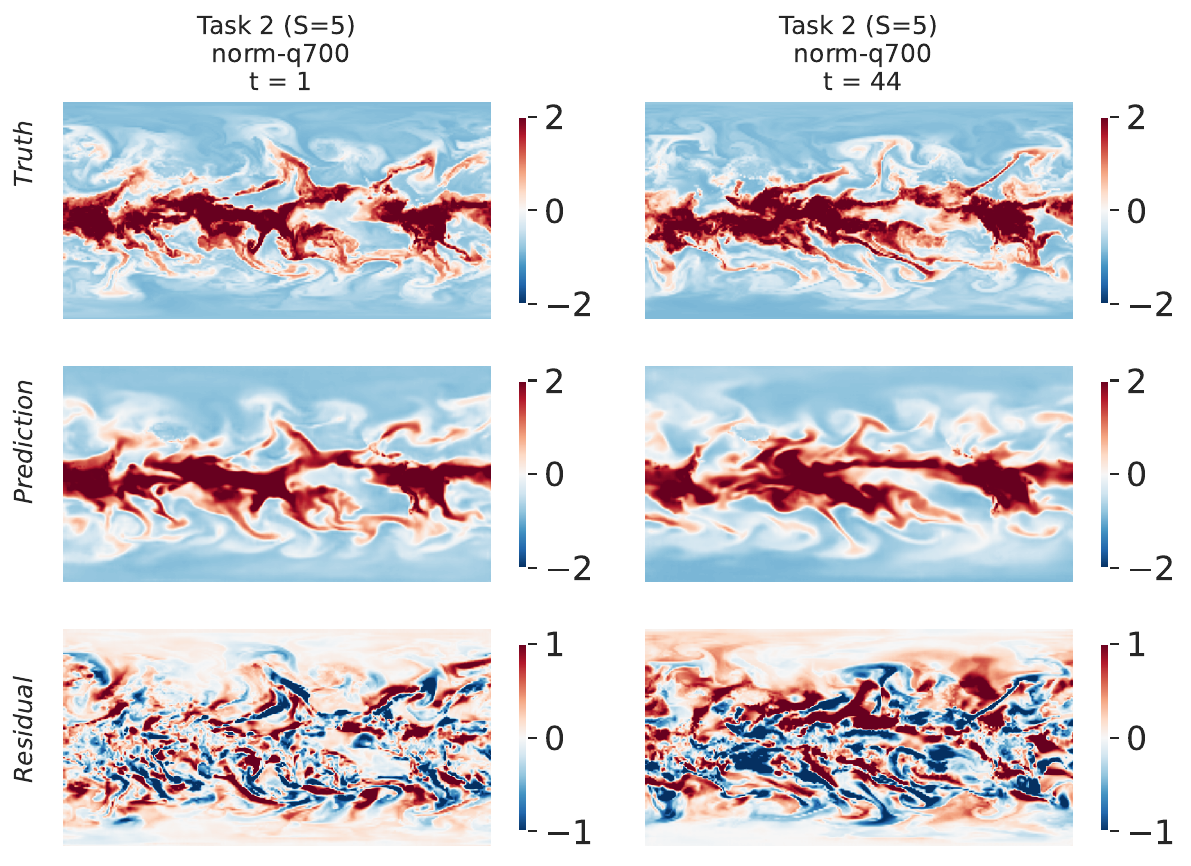}
        \caption{Task 2}
    \end{subfigure}
    
\caption{Normalized q@700-hpa qualitative results for UNet-autoregressive (S=5).}
\label{si-fig:preds_q700_s5}
\end{figure*}

\begin{figure*}[h]
    \centering
    \begin{subfigure}{0.48\textwidth}
        \includegraphics[width=\textwidth]{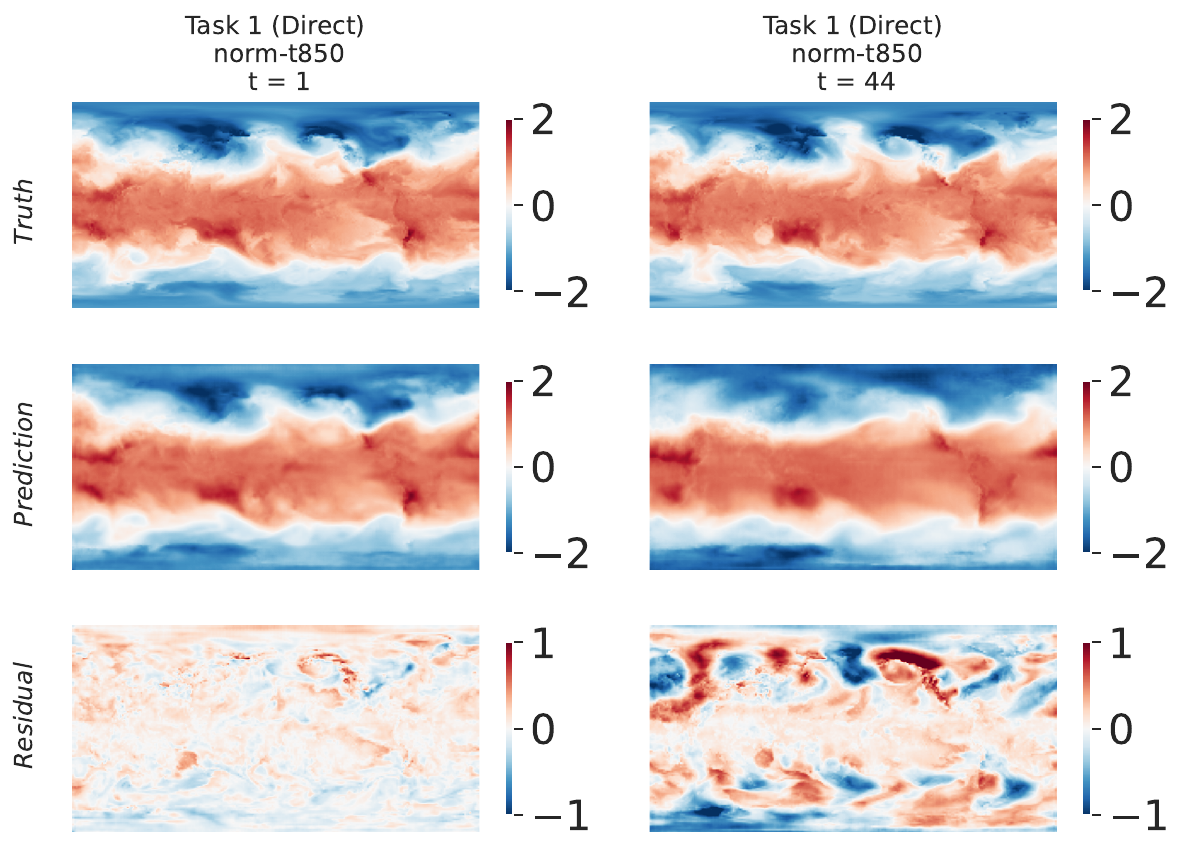}
        \caption{Task 1}
    \end{subfigure}
    \hfill
    \begin{subfigure}{0.48\textwidth}
        \includegraphics[width=\textwidth]{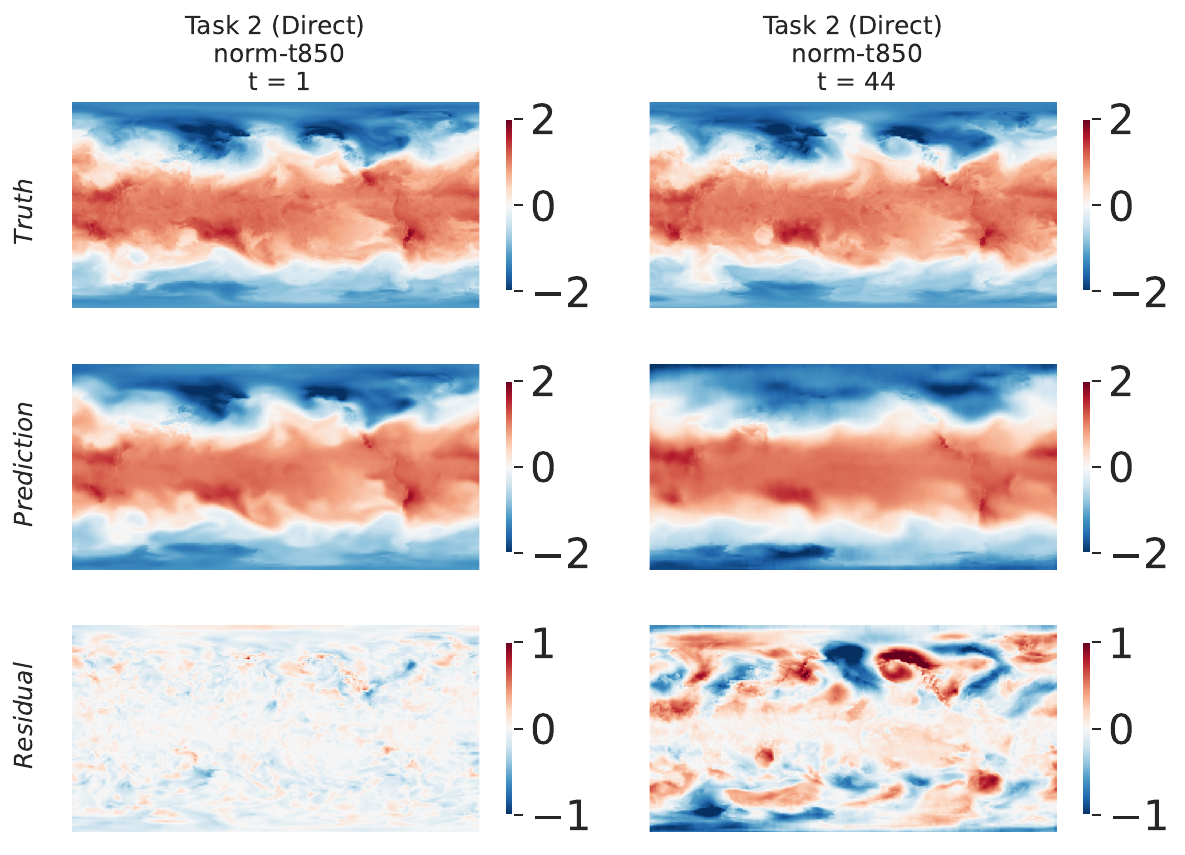}
        \caption{Task 2}
    \end{subfigure}
    
\caption{Normalized t@850-hpa qualitative results for UNet-direct.}
\label{si-fig:preds_t850_unet_direct}
\end{figure*}

\begin{figure*}[h]
    \centering
    \begin{subfigure}{0.48\textwidth}
        \includegraphics[width=\textwidth]{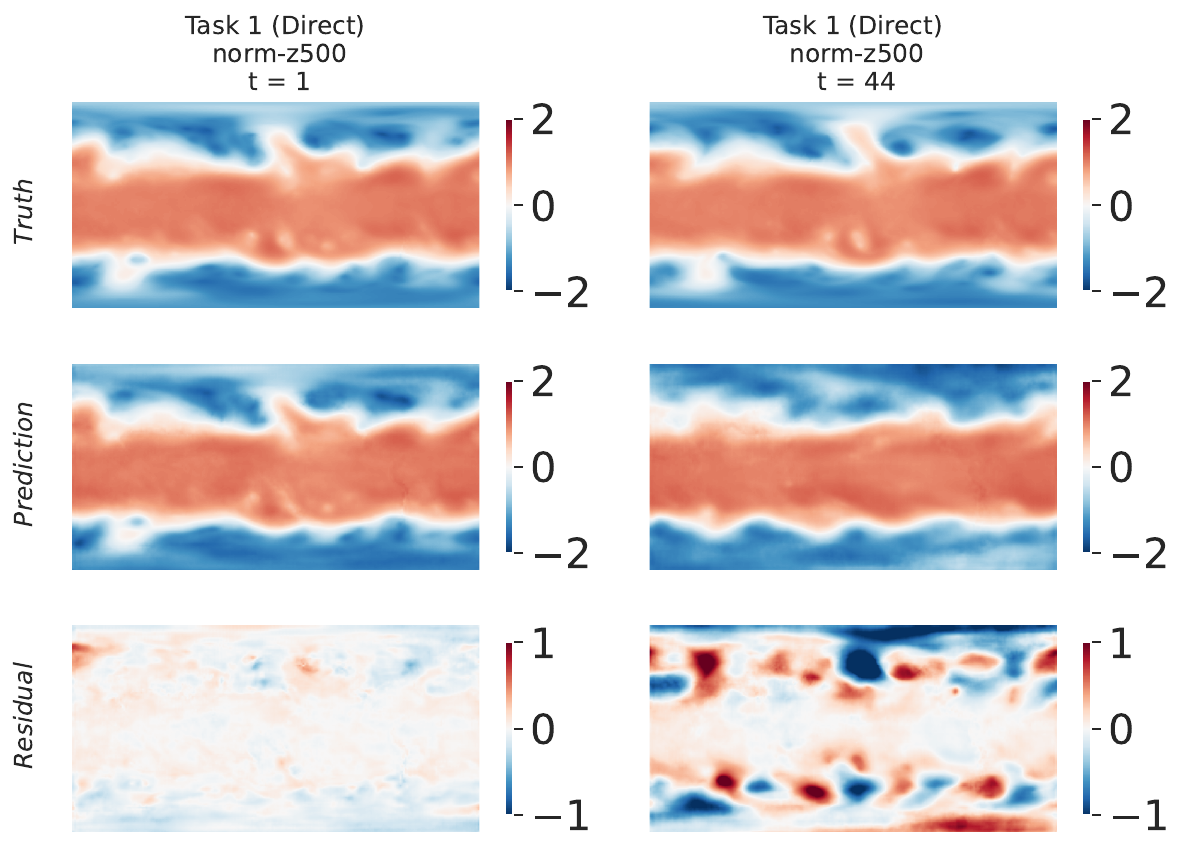}
        \caption{Task 1}
    \end{subfigure}
    \hfill
    \begin{subfigure}{0.48\textwidth}
        \includegraphics[width=\textwidth]{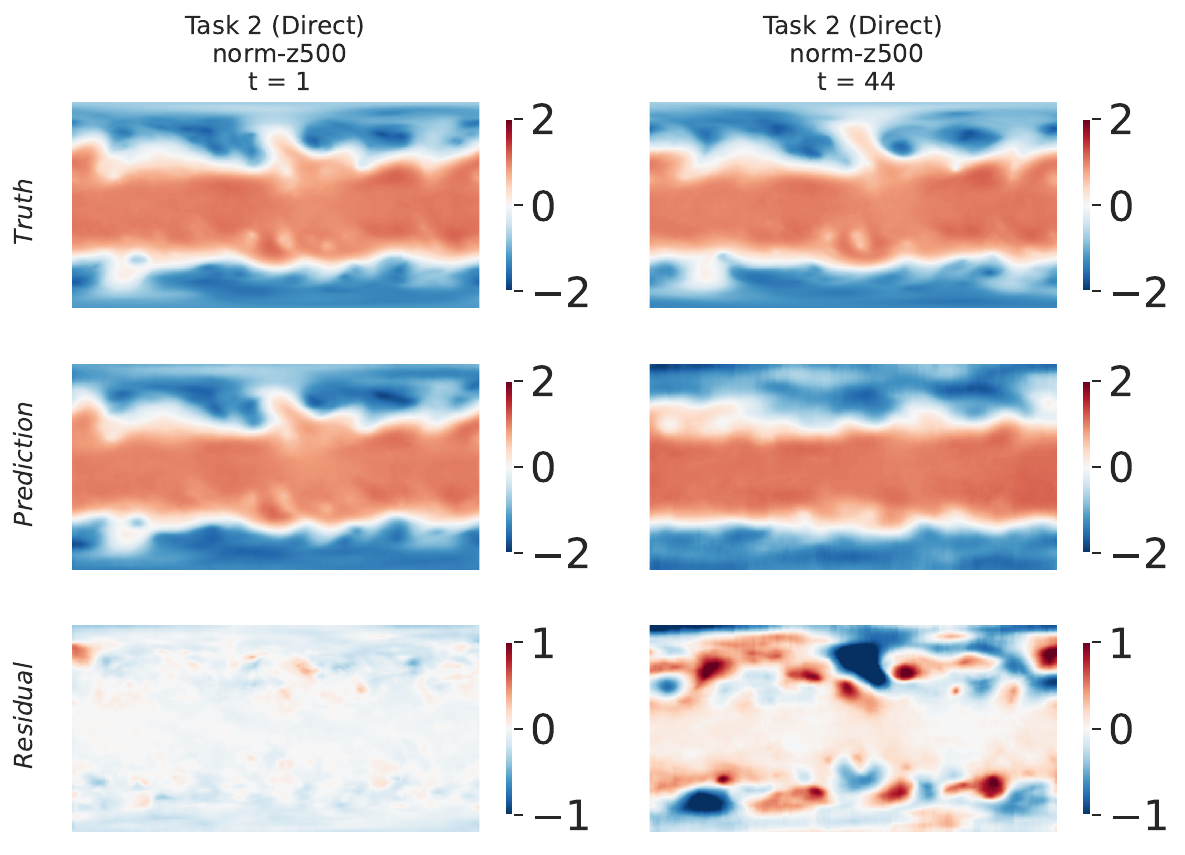}
        \caption{Task 2}
    \end{subfigure}
    
\caption{Normalized z@500-hpa qualitative results for UNet-direct.}
\label{si-fig:preds_z500_unet_direct}
\end{figure*}
\newpage

\begin{figure*}[h]
    \centering
    \begin{subfigure}{0.48\textwidth}
        \includegraphics[width=\textwidth]{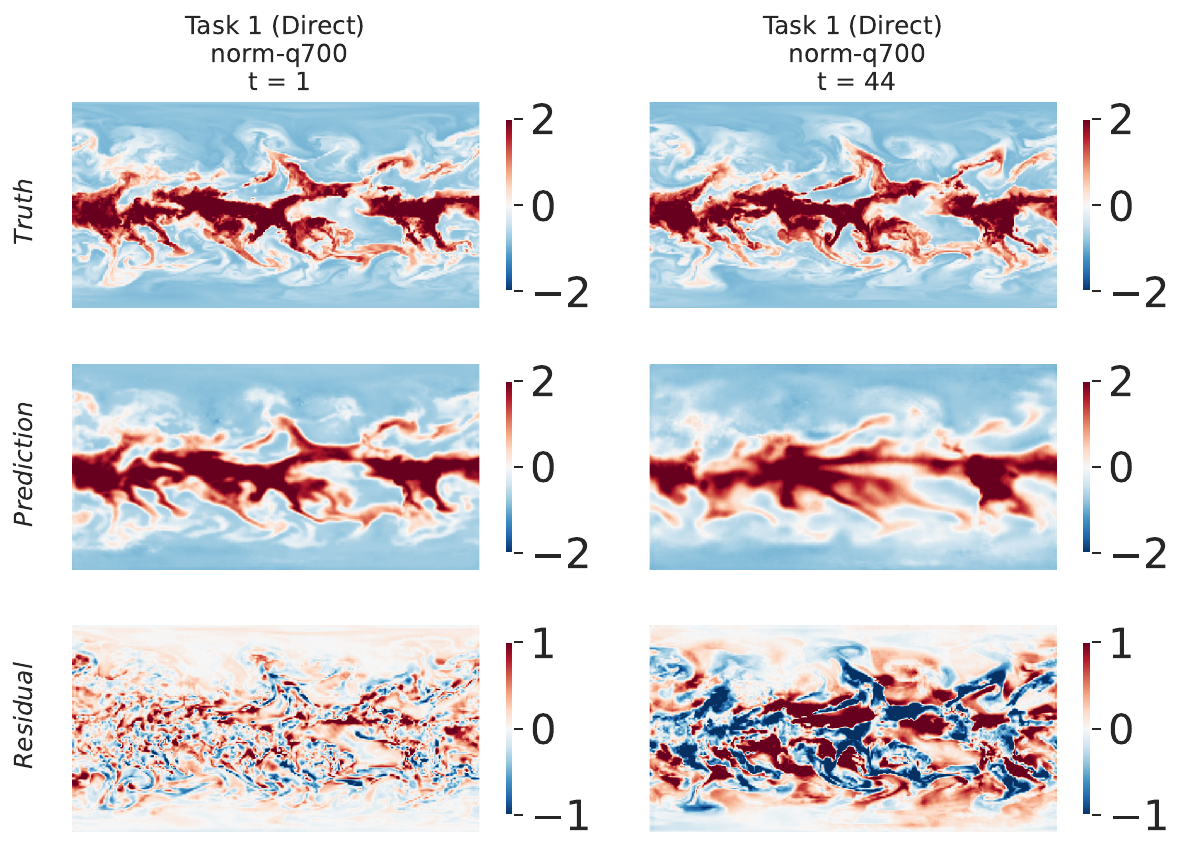}
        \caption{Task 1}
    \end{subfigure}
    \hfill
    \begin{subfigure}{0.48\textwidth}
        \includegraphics[width=\textwidth]{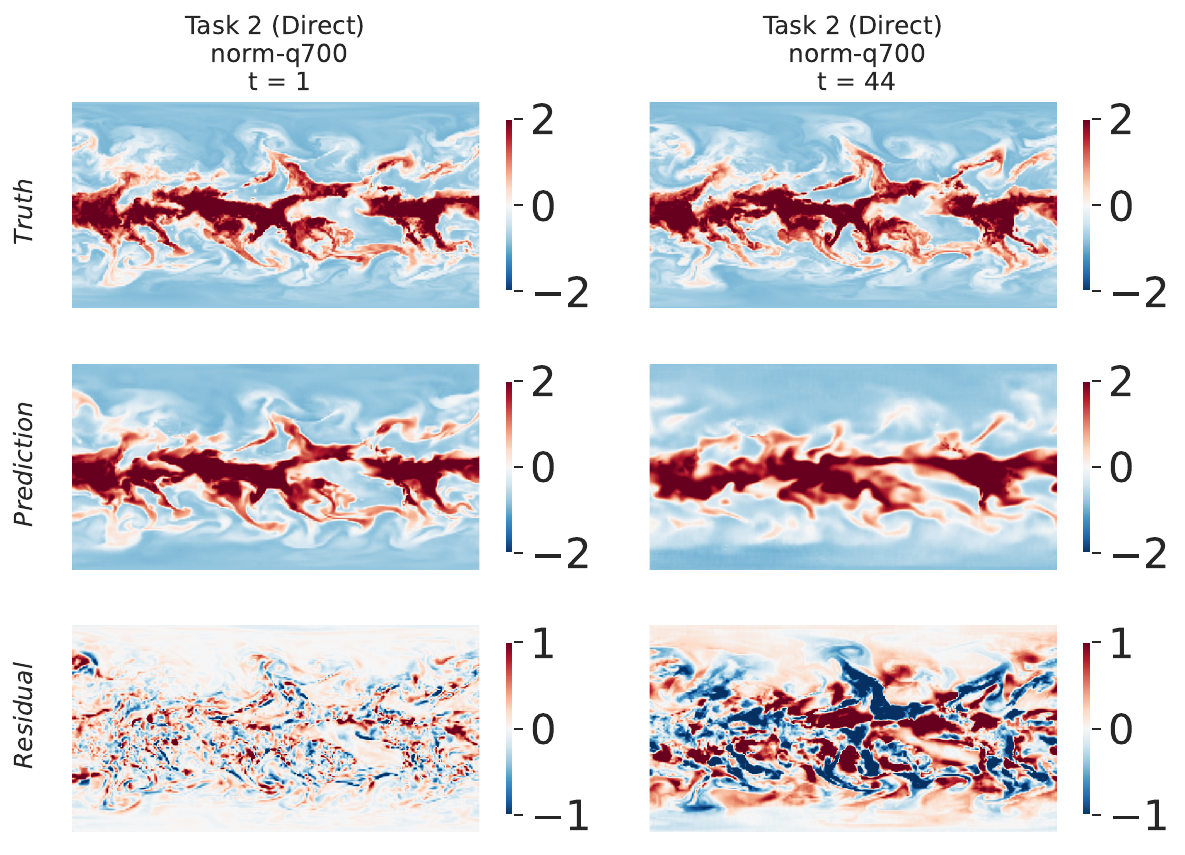}
        \caption{Task 2}
    \end{subfigure}
    
\caption{Normalized q@700-hpa qualitative results for UNet-direct.}
\label{si-fig:preds_q700_unet_direct}
\end{figure*}

\begin{figure*}[h]
    \centering
    \begin{subfigure}{0.48\textwidth}
        \includegraphics[width=\textwidth]{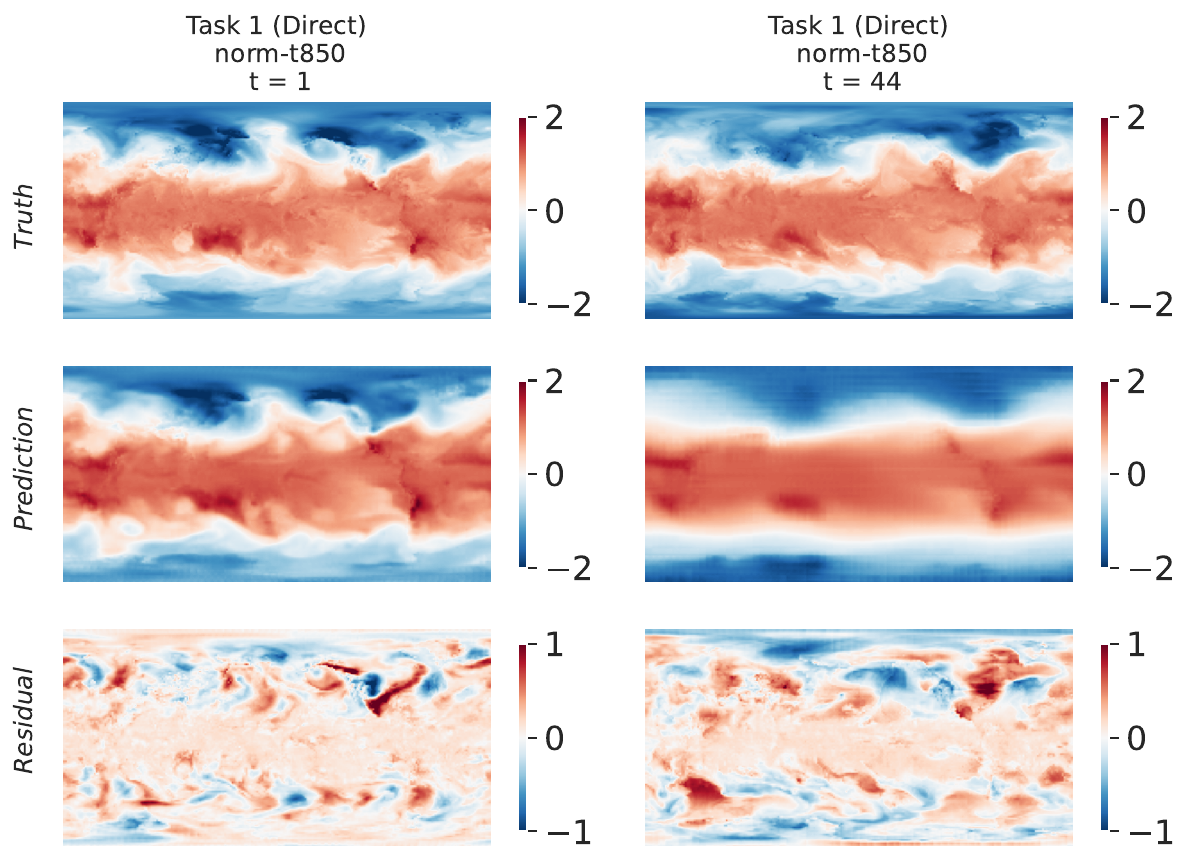}
        \caption{Task 1}
    \end{subfigure}
    \hfill
    \begin{subfigure}{0.48\textwidth}
        \includegraphics[width=\textwidth]{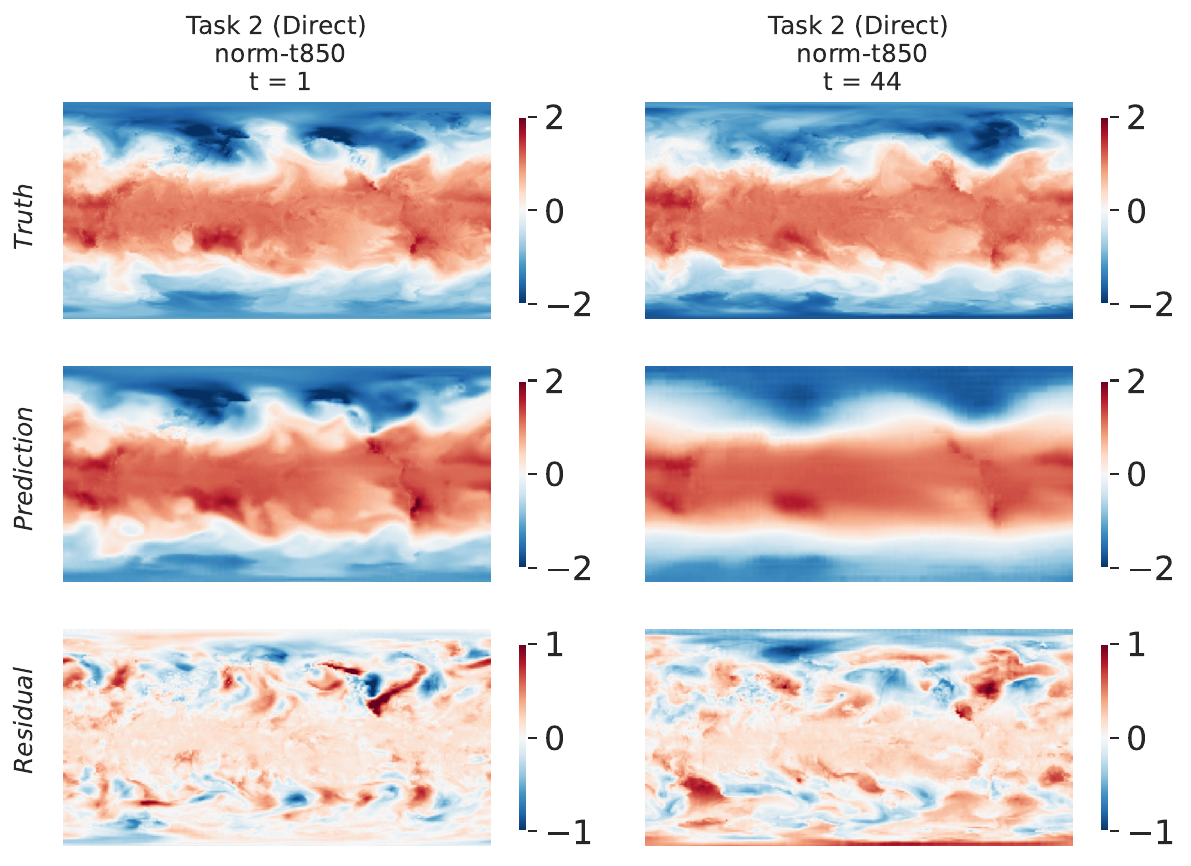}
        \caption{Task 2}
    \end{subfigure}
    
\caption{Normalized t@850-hpa qualitative results for ClimaX-direct.}
\label{si-fig:preds_t850_climax_direct}
\end{figure*}

\begin{figure*}[h]
    \centering
    \begin{subfigure}{0.48\textwidth}
        \includegraphics[width=\textwidth]{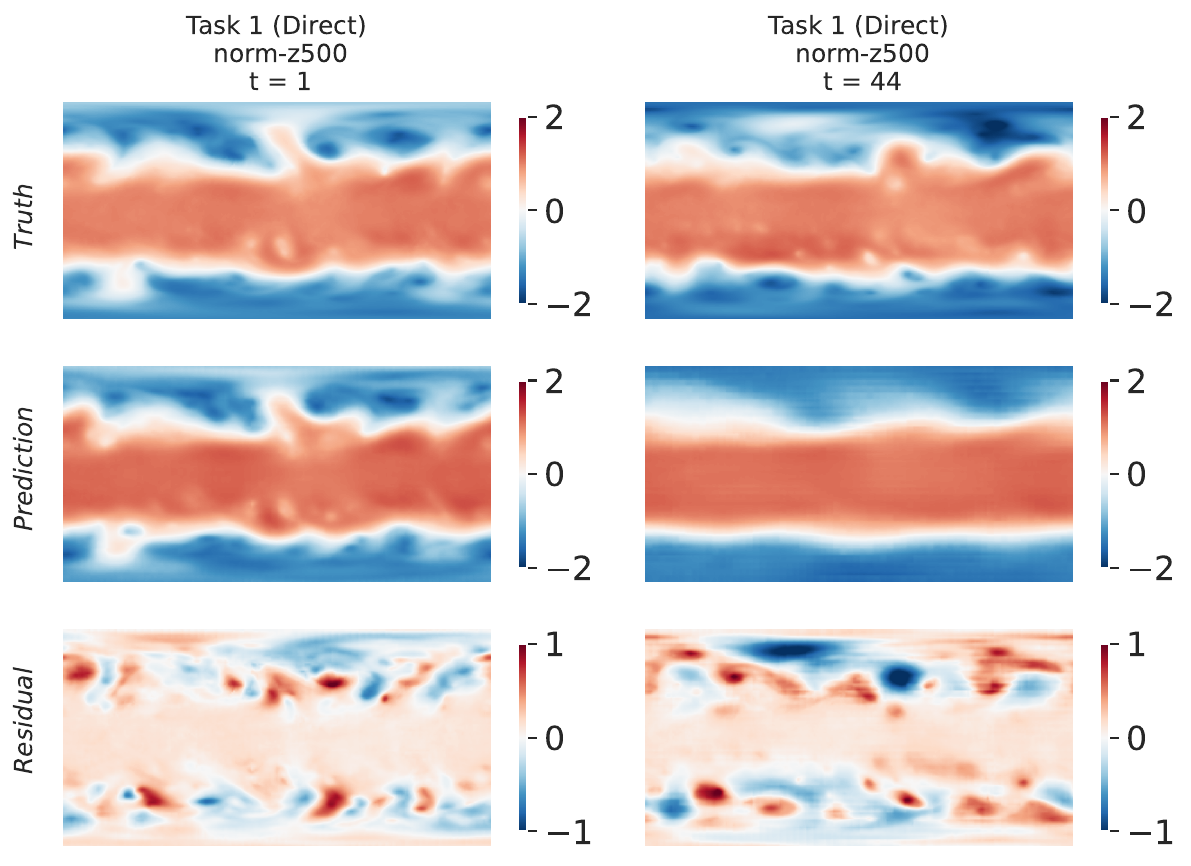}
        \caption{Task 1}
    \end{subfigure}
    \hfill
    \begin{subfigure}{0.48\textwidth}
        \includegraphics[width=\textwidth]{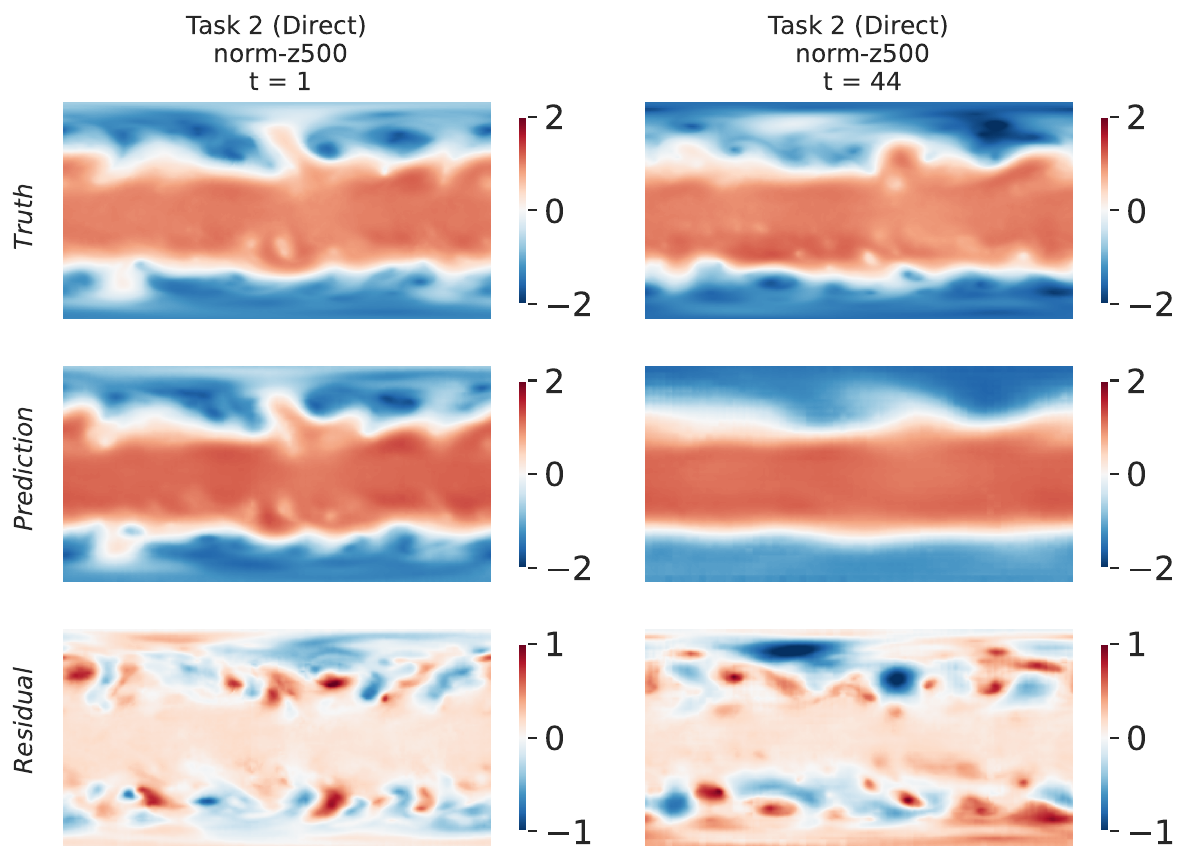}
        \caption{Task 2}
    \end{subfigure}
    
\caption{Normalized z@500-hpa qualitative results for ClimaX-direct.}
\label{si-fig:preds_z500_climax_direct}
\end{figure*}

\begin{figure*}[h]
    \centering
    \begin{subfigure}{0.48\textwidth}
        \includegraphics[width=\textwidth]{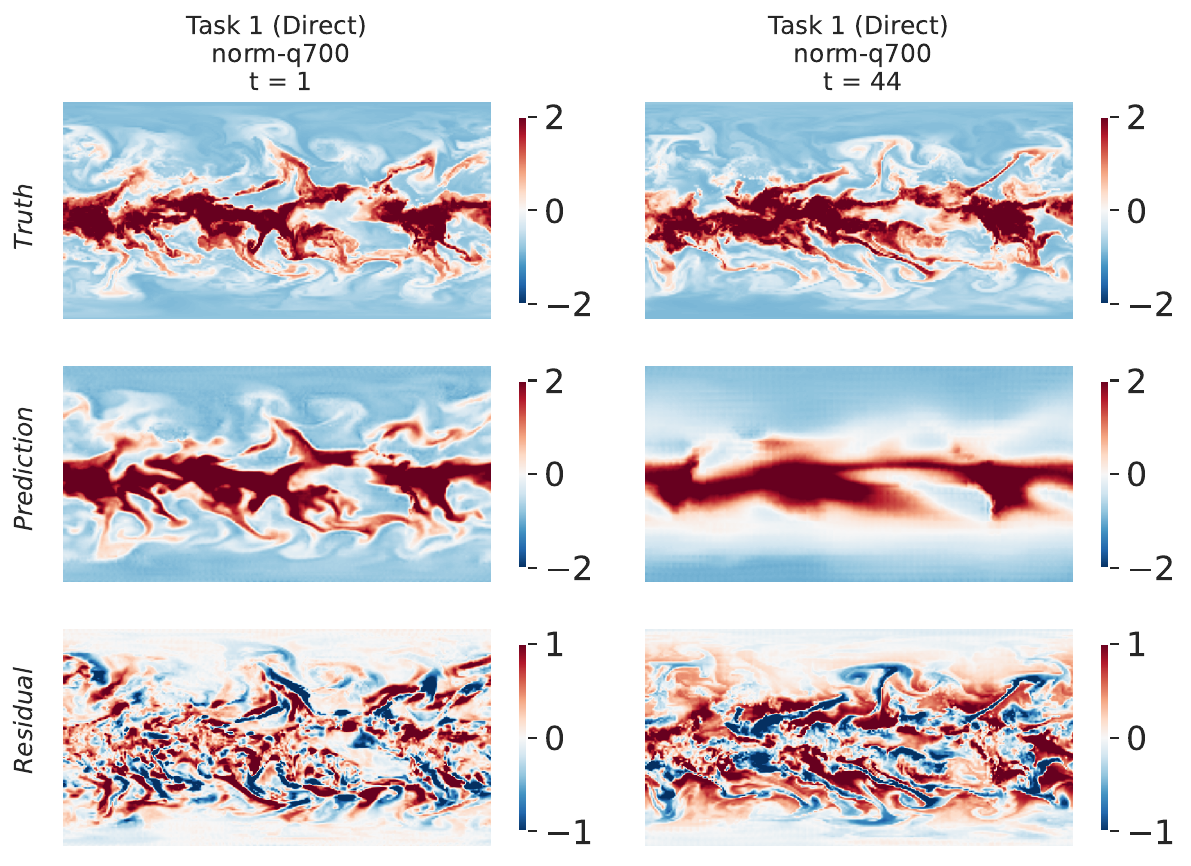}
        \caption{Task 1}
    \end{subfigure}
    \hfill
    \begin{subfigure}{0.48\textwidth}
        \includegraphics[width=\textwidth]{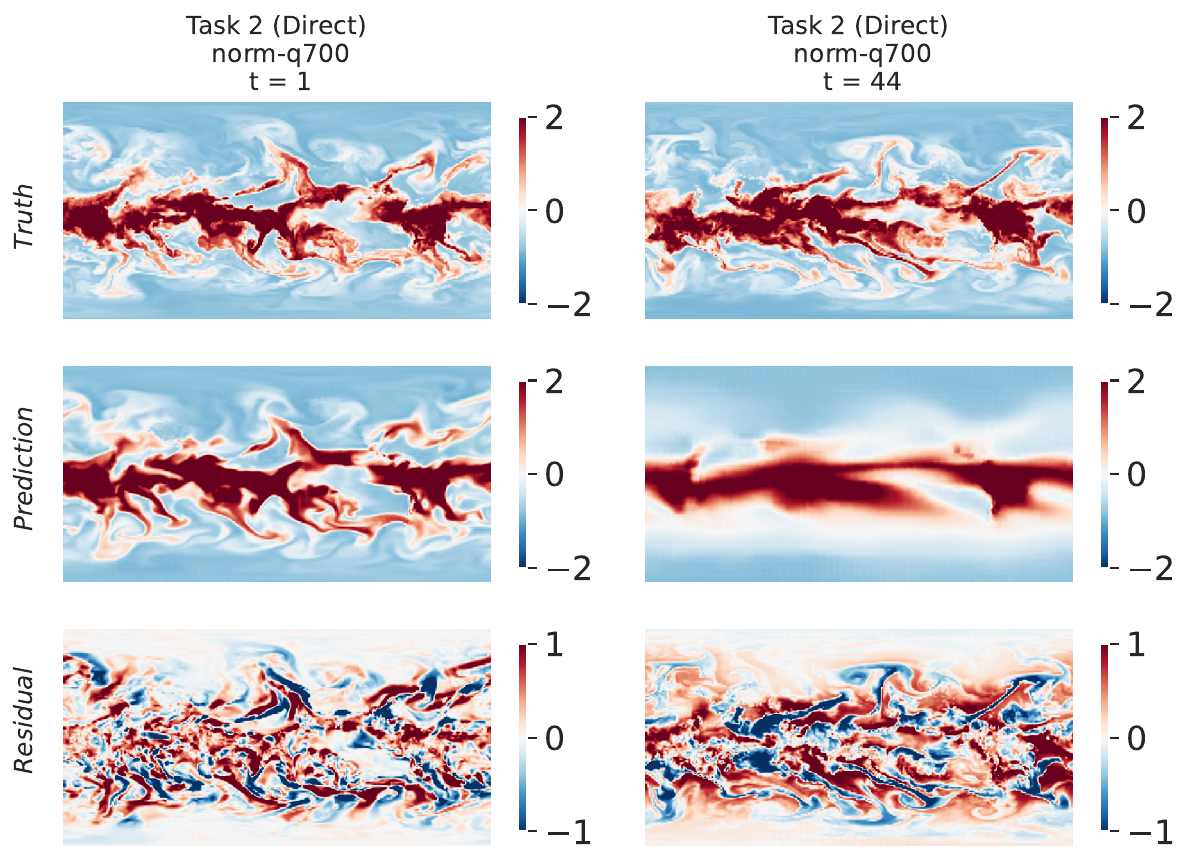}
        \caption{Task 2}
    \end{subfigure}
    
\caption{Normalized q@700-hpa qualitative results for ClimaX-direct.}
\label{si-fig:preds_q500_climax_direct}
\end{figure*}

\end{document}